\documentclass{article}

\PassOptionsToPackage{authoryear,round}{natbib}
\usepackage[eandd,final]{neurips_2026}
\setcitestyle{authoryear,round}

\usepackage[utf8]{inputenc}
\usepackage[T1]{fontenc}
\usepackage{graphicx}
\usepackage[normalem]{ulem}
\usepackage{adjustbox}
\usepackage{amsmath}
\usepackage{amssymb}
\usepackage{placeins}
\usepackage{float}
\usepackage{wrapfig}
\usepackage{booktabs}
\usepackage{array}
\usepackage{tabularx}
\usepackage{longtable}
\usepackage{pifont}
\usepackage[hidelinks]{hyperref}
\usepackage{tcolorbox}
\tcbuselibrary{breakable}
\newenvironment{promptblock}[1]{%
  \begin{tcolorbox}[%
    breakable,
    colback=gray!5,
    colframe=black!30,
    coltitle=black,
    fonttitle=\bfseries,
    sharp corners,
    boxrule=0.5pt,
    arc=3pt,
    left=8pt,
    right=8pt,
    top=6pt,
    bottom=6pt,
    title={\textbf{#1}}]%
  \small\raggedright
}{%
  \end{tcolorbox}
}
\newenvironment{prompttext}{%
  \begingroup
  \small
  \parindent0pt
  \raggedright
  \obeylines
  \obeyspaces
}{%
  \par
  \endgroup
}

\newcommand{\appsection}[1]{%
  \FloatBarrier
  \section{#1}%
}
\makeatletter
\newcommand{\captionof}[1]{\def\@captype{#1}\caption}
\makeatother

\title{OmniToM: Benchmarking Theory of Mind in LLMs via Explicit Belief Modeling}

\author{
\textbf{Adam Bawatneh},
\textbf{Sagar Sapkota}, \textbf{Amrit Singh Bedi},\\
\textbf{Santu Karmaker}, and \textbf{Mubarak Shah}\\
University of Central Florida\\
\texttt{\{adam.bawatneh, santu\}@ucf.edu, shah@crcv.ucf.edu}\\
\href{https://bridgeai-lab.github.io/OMNITOM/}{Project page} \quad|\quad
\href{https://github.com/BridgeAI-Lab/OMNITOM}{Code} \quad|\quad
\href{https://huggingface.co/datasets/BridgeAI-Lab/omnitom-benchmark}{Benchmark}
}

\begin{document}

\maketitle

\begin{abstract}
Theory of Mind (ToM), the ability to infer others’ knowledge, intentions, and emotions, is commonly evaluated in large language models (LLMs) using endpoint question answering, where performance is judged solely by the final answer to a social reasoning query. This paradigm obscures whether the model actually constructs the underlying mental-state representations required for robust reasoning, particularly in scenarios involving divergent, evolving, or mistaken beliefs.
In order to address this research gap, we introduce OmniToM, a benchmark that directly evaluates these representations by requiring explicit modeling of belief structures for all relevant actors within a narrative.
These structures are composed of belief propositions: minimal statements of what an actor takes to be true about the world or another actor's mental state, allowing knowledge, intentions, emotions, and false beliefs to be analyzed in a common format. Models are evaluated in two stages: Stage 1: Belief Extraction, which extracts from the story the beliefs relevant to its social dynamics, and Stage 2: Belief Labeling, which assigns each belief a seven-dimensional schema label covering recursive order, truth status, knowledge access, explicitness, content type, mental source, and context. Built from 895 stories from the existing ToMBench story corpus and augmented with 22,343 labeled belief propositions, OmniToM uses a human-validated LLM-assisted annotation pipeline. Across diverse models in zero-shot evaluation, OmniToM reveals an actor-specific belief-tracking bottleneck: current LLMs struggle with the knowledge-access and representational decisions required to transform narrative facts into actors’ beliefs and shared mental states.

\end{abstract}

\section{Introduction}

\begin{figure*}[t]
    \centering
    \includegraphics[width=\linewidth]{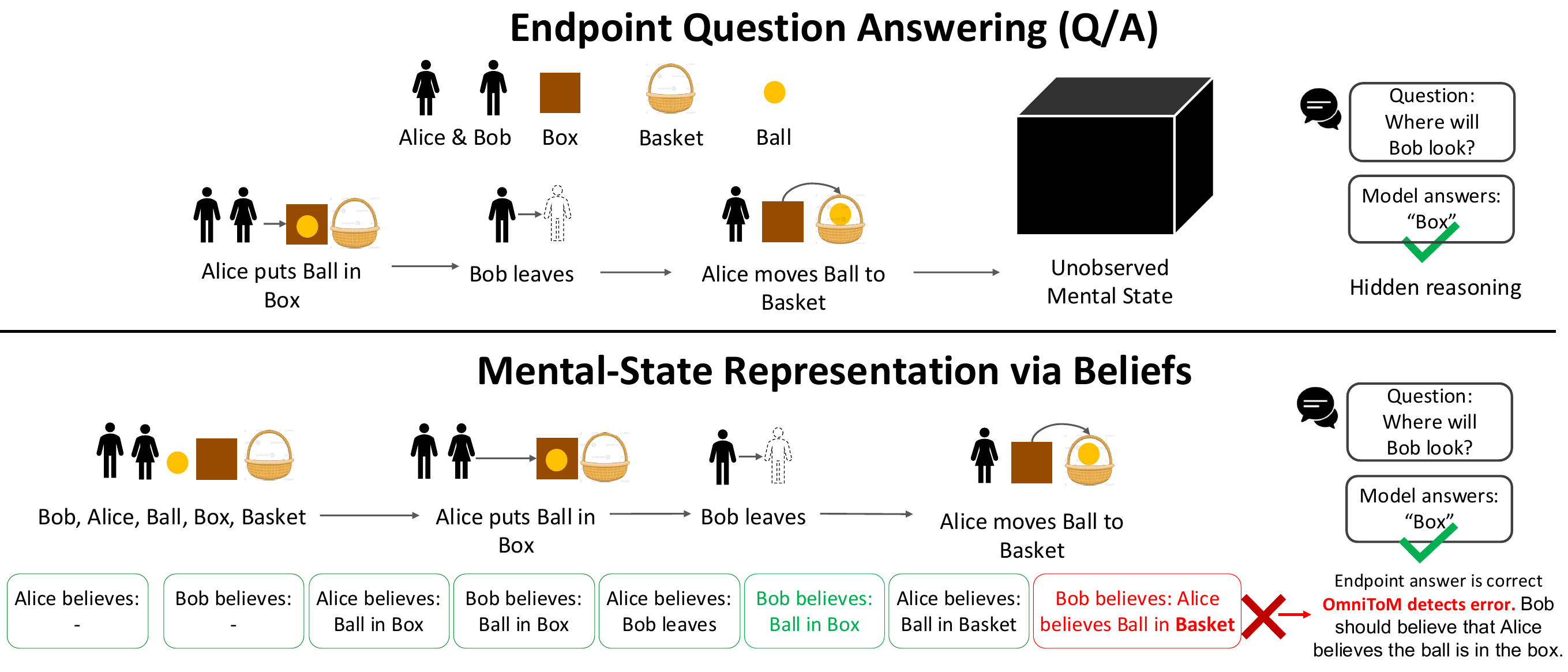}
\caption{Endpoint QA versus explicit belief modeling. Alice places a ball in a box, Bob leaves, and Alice moves it to a basket. Top: Endpoint Question Answering (QA) evaluates only the final answer to \textit{``Where will Bob look?''}; the model correctly answers \textit{``Box''}, while the broader mental-state representation remains unobserved. Bottom: Explicit belief modeling reveals that the model correctly tracks Bob's belief about the ball, but incorrectly represents Bob's belief about Alice's belief. Since Bob did not observe the move, he should still believe that Alice believes the ball is in the box. OmniToM exposes this belief error despite the correct endpoint answer.}
\label{fig:qa_vs_reconstruction}
\end{figure*}

Social reasoning requires tracking how information is distributed across actors, not only what happened in the world. To predict what a person will do, a model must represent what each actor knows, infers, intends, or falsely believes. This capacity is commonly studied as Theory of Mind (ToM), the ability to attribute mental states such as beliefs, intentions, and emotions to oneself and others \citep{premack1978does, beaudoin2020systematic}. For large language model (LLM) evaluation, the key question is therefore not merely whether a model can answer a social-reasoning question, but whether it recovers the supporting multi-actor mental-state representation needed for robust ToM reasoning. Existing LLM ToM benchmarks usually test this ability indirectly through endpoint question answering (QA): given a story or dialogue, the model is scored by whether it produces the correct final answer \citep{le2019revisiting, kim2023fantom, wu2023hitom, xu2024opentom, chen-etal-2024-tombench}. As illustrated in Fig.~\ref{fig:qa_vs_reconstruction}, endpoint correctness leaves the underlying mental-state representation unobserved. A model may answer a false-belief question correctly, while failing to represent the actor-specific mental states that make the answer valid, including what each actor takes to be true, infers, remembers, or attributes to others. Recent work has, therefore, raised concerns that endpoint ToM scores may reflect benchmark artifacts, shortcut strategies, or task scaffolding rather than robust mental-state tracking \citep{le2019revisiting, sclar2023minding, pi2024scalpel}.

Endpoint QA also limits fine-grained analysis. ToM is not a single monolithic ability: it involves interacting aspects of mental-state reasoning, including recursive belief attribution, factual alignment, information access, pragmatic inference, intentions, emotions, and temporal belief change \citep{beaudoin2020systematic,WimmerPerner1983,PernerWimmer1985,Flavell1986,Happe1994,GoodmanStuhlmueller2013}. Existing benchmarks provide valuable coverage of particular ToM settings, including false belief \citep{le2019revisiting}, higher-order reasoning \citep{wu2023hitom}, dialogue-based information asymmetry \citep{kim2023fantom}, perspective taking \citep{xu2024opentom}, and broader multi-task ToM evaluation \citep{chen-etal-2024-tombench}. However, when these abilities are evaluated primarily through endpoint answers or task-specific outputs, it remains difficult to analyze how different aspects of the underlying mental-state representation interact, for example whether a model failed because it missed a relevant actor, updated the wrong belief, misidentified who had access to information, or mislabeled the source or content of a mental state. This leaves a central evaluation gap: existing benchmarks can test whether a model selects the correct endpoint answer, but they do not directly test whether the model recovered the multi-actor mental-state representation that makes the answer socially meaningful.

Therefore, we introduce OmniToM, a benchmark designed to address this research gap through explicit belief-structure modeling. OmniToM operationalizes the supporting mental-state representation as actor-specific belief propositions: minimal statements of what an actor takes to be true about the world or another actor's mental state. This formulation provides a common format for analyzing knowledge, intentions, emotions, false beliefs, and nested mental states without reducing ToM to a single endpoint answer. OmniToM evaluates this representation in two stages. In \textit{Stage~1: Belief Extraction}, a model extracts from the story the belief propositions relevant to its social dynamics. In \textit{Stage~2: Belief Labeling}, the model labels each belief proposition under a unified seven-dimensional schema grounded in ATOMS (\emph{Abilities in Theory of Mind Space}), a literature-derived taxonomy of ToM abilities~\citep{beaudoin2020systematic}. ATOMS guides the range of mental-state reasoning that OmniToM aims to model. OmniToM operationalizes this coverage through seven belief-level dimensions: recursive belief depth (e.g., \textit{Bob believes Alice thinks \(X\)}) (\textit{Order}); alignment with story reality (\textit{Truth Status}); who can access or share the information (\textit{Knowledge Access}); stated versus inferred content (\textit{Representation}); belief subject matter (\textit{Content Type}); acquisition source (\textit{Mental Source}); and whether any special framing condition applies (\textit{Context}).

OmniToM is built from 895 stories derived from ToMBench, a prior ToM benchmark, and augmented with 22{,}343 labeled belief propositions. Its construction was supported by over 1K person-hours of human annotation effort for benchmark development, before a human-validated LLM-assisted annotation pipeline was finalized and scaled to the full benchmark. Across diverse open- and closed-source models under zero-shot evaluation, Stage~2 belief-labeling accuracy reaches 85.95\%, while Stage~1 extraction \(F_1\) peaks at 61.63\%. More importantly, OmniToM localizes the same bottleneck across both stages: Stage~1 performance drops when story facts must be assigned to actor-specific beliefs, and Stage~2 results point to difficulty with \textit{Knowledge Access}, with some models also struggling on \textit{Representation}. This suggests that current LLMs struggle not simply to parse social stories, but to track which information each actor has, how it is communicated or inferred, and how it becomes part of that actor's mental-state representation.

Our core contributions are threefold:
\begin{itemize}
    \item We introduce OmniToM, a benchmark of 895 ToMBench-derived stories and 22{,}343 labeled belief propositions, developed through over 1K person-hours of human annotation effort to evaluate multi-actor mental-state representations beyond endpoint answers.

    \item We introduce an ATOMS-grounded belief-level schema for fine-grained ToM analysis, translating task-level ability coverage into seven per-proposition dimensions: \textit{Order}, \textit{Truth Status}, \textit{Knowledge Access}, \textit{Representation}, \textit{Content Type}, \textit{Mental Source}, and \textit{Context}.

    \item We evaluate diverse open- and closed-source LLMs and find an actor-specific information-tracking bottleneck: models struggle to determine which story facts each actor knows, shares, or infers, and how those facts become beliefs.
\end{itemize}

\section{Related Work}
\label{sec:related_work}

\begin{table*}[t]
  \setlength{\belowcaptionskip}{4pt}
    \centering
    \small
    \caption{Comparison with representative ToM benchmarks for language models. We compare each benchmark's evaluation format and whether it explicitly evaluates the seven OmniToM schema dimensions as separate metrics.  
    A checkmark (\ding{51}) indicates explicit evaluation; a dash (--) indicates that the dimension is absent or only implicit.}
    \label{tab:benchmark_comparison}
    \begin{adjustbox}{max width=\textwidth,center}
    \begin{tabular}{l l c c c c c c c}
        \toprule
        \textbf{Benchmark} & \textbf{Evaluation Method} & \textbf{Order} & \textbf{Truth} & \textbf{Access} & \textbf{Repr.} & \textbf{Content} & \textbf{Source} & \textbf{Context} \\
        \midrule
        ToMi \citep{le2019revisiting} & Synthetic QA & \ding{51} & \ding{51} & -- & -- & -- & -- & -- \\
        Hi-ToM \citep{wu2023hitom} & Recursive QA & \ding{51} & -- & -- & -- & -- & -- & -- \\
        FANToM \citep{kim2023fantom} & Dialogue QA & -- & \ding{51} & -- & -- & -- & -- & \ding{51} \\
        OpenToM \citep{xu2024opentom} & Perspective QA & -- & \ding{51} & -- & -- & \ding{51} & -- & -- \\
        ToMBench \citep{chen-etal-2024-tombench} & Multi-task QA & \ding{51} & \ding{51} & -- & -- & \ding{51} & -- & \ding{51} \\
        SymbolicToM \citep{sclar2023minding} & Structured QA & \ding{51} & \ding{51} & -- & -- & -- & -- & -- \\
        \midrule
        OmniToM (Ours) & Belief-structure modeling & \ding{51} & \ding{51} & \ding{51} & \ding{51} & \ding{51} & \ding{51} & \ding{51} \\
        \bottomrule
    \end{tabular}
    \end{adjustbox}
\end{table*}

\paragraph{Endpoint QA and mental-state representations.}
LLM benchmarks for Theory of Mind (ToM) predominantly evaluate social reasoning through endpoint question answering (QA): a model reads a story or dialogue and is scored by whether it returns the correct final answer \citep{le2019revisiting, wu2023hitom, kim2023fantom, xu2024opentom, chen-etal-2024-tombench}. As illustrated in Fig.~\ref{fig:qa_vs_reconstruction}, endpoint QA can leave the supporting mental-state representation unobserved: a model may answer correctly without tracking what each actor takes to be true, remembers, infers, intends, feels, or attributes to others. Recent work has begun to make such intermediate reasoning more explicit. SymbolicToM introduces a multi-character belief tracker, but uses it as a scaffold for improving downstream QA rather than as the primary benchmark target \citep{sclar2023minding}. Perceptions-to-Beliefs evaluates a narrower precursor pathway, asking whether models can infer what characters perceive and convert those perceptions into beliefs using perception annotations added to ToMi and FANToM \citep{jung2024perceptions}. This shows that models may identify perceptual access while still struggling to infer the beliefs that follow from it. OmniToM targets a broader representation: the full set of actor-specific belief propositions relevant to a story's social dynamics, including beliefs grounded in perception, memory, testimony, inference, imagination, and higher-order attribution.

\paragraph{Ability spaces and schema-guided analysis.}
Evaluating mental-state representations requires analyzing which aspects of those representations succeed or fail. ATOMS (Abilities in Theory of Mind Space) organizes ToM measures into task-level ability categories and sub-abilities, including beliefs, knowledge, intentions, desires, emotions, percepts, and non-literal communication~\citep{beaudoin2020systematic}. This provides a principled coverage scaffold, but not a direct label inventory for individual belief propositions. OmniToM adapts this ability-space perspective into an ATOMS-grounded belief-level schema that labels each proposition by recursive depth (\textit{Order}), factual alignment (\textit{Truth Status}), information sharing (\textit{Knowledge Access}), explicitness (\textit{Representation}), subject matter (\textit{Content Type}), acquisition source (\textit{Mental Source}), and framing (\textit{Context}). Prior benchmarks instantiate important subsets of this space: ToMi focuses on false-belief QA, Hi-ToM on higher-order reasoning, FANToM on information asymmetry, OpenToM on physical and psychological states, and ToMBench on multi-task ToM QA~\citep{le2019revisiting,wu2023hitom,kim2023fantom,xu2024opentom,chen-etal-2024-tombench}. Table~\ref{tab:benchmark_comparison} compares these benchmarks by evaluation format and by whether they explicitly score OmniToM's seven dimensions, highlighting OmniToM's shift from endpoint evaluation to structured evaluation of the belief representation that supports those answers.

\paragraph{LLM-assisted benchmark construction and evaluation.}
Constructing dense belief-structure benchmarks is costly: each story can require many actor-specific propositions, and each proposition must be labeled along multiple schema dimensions. Recent work increasingly uses LLMs to support scalable data annotation and synthesis when fully manual labeling is impractical \citep{tan-etal-2024-large}. OmniToM uses LLMs in this spirit, but constrains their role through task-conditioned prompting and human validation. In particular, we use TELeR, a prompt taxonomy for benchmarking complex tasks, to specify task-conditioned extraction, labeling, and evaluation prompts \citep{karmaker-santu-feng-2023-teler}, then validate and select the resulting pipeline on a human-annotated subset before scaling to the full benchmark. Because OmniToM also uses LLM-based semantic evaluation for open-ended belief extraction, it follows prior LLM-as-a-judge work \citep{zheng2023judging} while adopting the human-checking emphasis highlighted by JudgeBench \citep{tan2025judgebench}. Thus, LLMs support benchmark construction and evaluation, but the final pipeline is task-conditioned, human-validated, and selected through agreement checks before use.

\section{Benchmark Formulation}
\label{sec:framework}

\begin{figure}[t]
\centering
\includegraphics[width=\columnwidth]{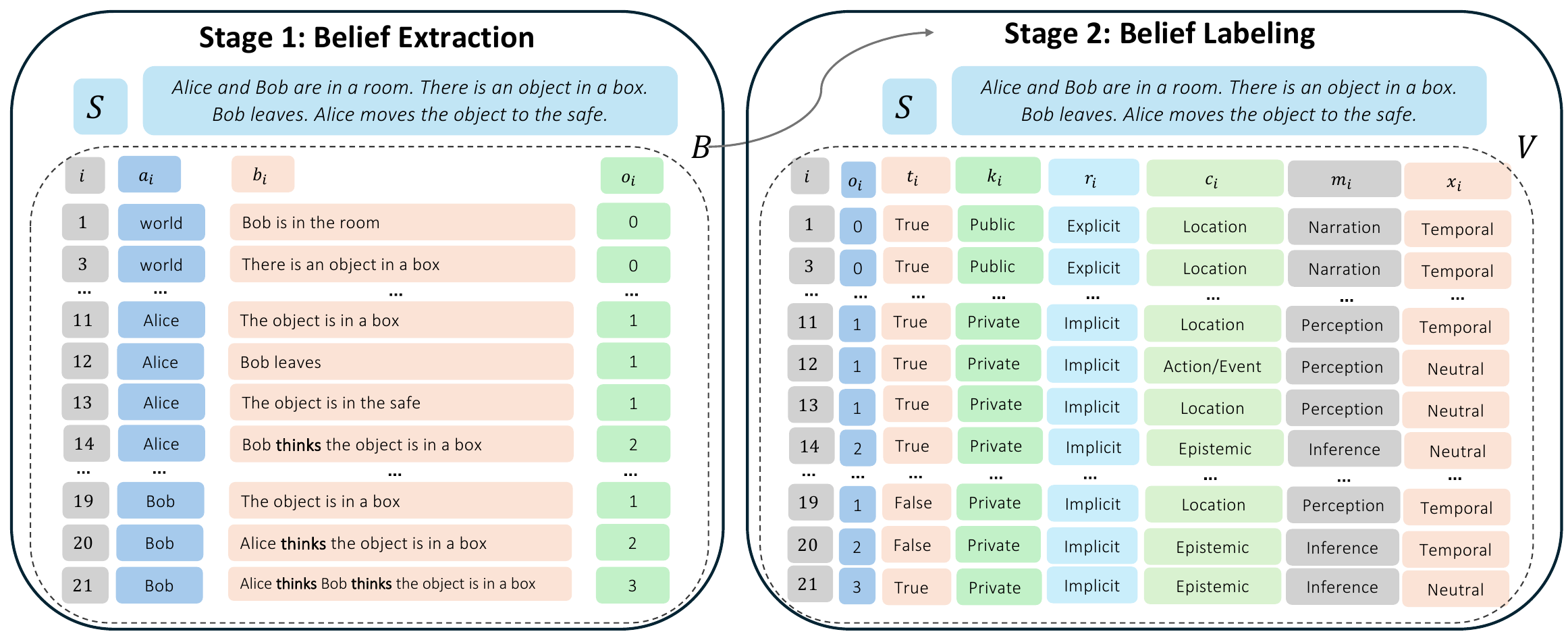}
    \caption{OmniToM two-stage functional workflow. Stage~1 (Belief Extraction) maps the story \(\mathcal{S}\) to extracted world-fact and actor-belief propositions, \(f_{ext}(\mathcal{S}) \rightarrow \mathbf{B}\), where \(\mathbf{B}\) contains actor \(a_i\), belief proposition \(b_i\), and order \(o_i\) tuples and uses the special actor \textit{world} for narrated facts. Stage~2 (Belief Labeling) takes the story and extracted propositions, \(f_{label}(\mathcal{S}, \mathbf{B}_{ab}) \rightarrow \mathbf{V}\), and outputs one seven-dimensional schema vector per proposition, \(\mathbf{s}_i=(o_i,t_i,k_i,r_i,c_i,m_i,x_i)\), corresponding to \textit{Order} (\(o_i\)), \textit{Truth Status} (\(t_i\)), \textit{Knowledge Access} (\(k_i\)), \textit{Representation} (\(r_i\)), \textit{Content Type} (\(c_i\)), \textit{Mental Source} (\(m_i\)), and \textit{Context} (\(x_i\)). The visualization shows condensed snapshots from a canonical false-belief example; worked examples appear in App.~\ref{app:category-examples}.}
\label{fig:omnitom-pipeline}
\end{figure}

\paragraph{Formal Task Definition.}
OmniToM formulates benchmark evaluation as \emph{explicit belief-structure modeling}: given a story, a model extracts a structured representation composed of narrated world facts and belief propositions held by all relevant actors, then labels each extracted proposition under a shared schema. Concretely (Fig.~\ref{fig:omnitom-pipeline}), Stage~1 maps the story \(\mathcal{S}\) to a belief structure, \(f_{ext}(\mathcal{S}) \rightarrow \mathbf{B}\), and Stage~2 labels the extracted belief propositions according to the schema, \(f_{label}(\mathcal{S}, \mathbf{B}_{ab}) \rightarrow \mathbf{V}\). Thus, the benchmark consists of two linked evaluation targets: belief-structure extraction followed by belief labeling. Both stages condition on the complete story, preserving multi-actor dependencies, information asymmetries, and belief changes across the narrative.

\paragraph{Stage~1: Belief Extraction.}
Under \(f_{ext}(\mathcal{S}) \rightarrow \mathbf{B}\), the extracted belief structure is \(\mathbf{B}=\{(a_i,b_i,o_i)\}_{i=1}^{n}\), a structured set of propositions: narrated world facts and actor-specific belief propositions, each with an order label. Here, \(a_i\) denotes either the actor (character or group) holding the belief or the special actor \textit{world}, which marks narrated facts not attributed to any single character's internal state. The proposition \(b_i\) denotes the minimal content being represented, and \(o_i \in \{0,1,2,3\}\) denotes recursive belief order, with \(0\) reserved for world-level facts. Following classic work on false belief and recursive belief attribution~\citep{WimmerPerner1983,PernerWimmer1985}, Stage~1 extracts the structure recursively: it first identifies narrated world facts, then the relevant actors, then each actor's first-order beliefs about the world, and finally higher-order beliefs about other actors' beliefs. This formulation focuses on the structured belief dependencies that determine each actor's interpretation of the story. It captures belief inaccuracies, knowledge asymmetries, and complex story-based mental-state reasoning expressed in text~\citep{Happe1994}. Temporal progression is implicit in the ordering of \(\mathbf{B}\), while changes in belief state are further labeled through the schema introduced in Stage~2.

\paragraph{Stage~2: Belief Labeling.}
Given the story and extracted belief propositions, Stage~2 labels each proposition with a seven-dimensional schema vector, \(f_{label}(\mathcal{S}, \mathbf{B}_{ab}) \rightarrow \mathbf{V}\), where \(\mathbf{s}_i=(o_i,t_i,k_i,r_i,c_i,m_i,x_i)\). The schema is grounded in ATOMS (\emph{Abilities in Theory of Mind Space})~\citep{beaudoin2020systematic}, which organizes ToM measures by the abilities they test. OmniToM adapts this task-level ability space into a belief-level schema for fine-grained analysis: each extracted proposition is labeled by the representational properties needed to model the underlying mental state, rather than only by the endpoint task category. For example, a false-belief task is not represented by a single ``false belief'' label; it is decomposed into belief-level properties such as who holds the belief, whether it conflicts with reality, who had access to the relevant information, whether the belief is explicit or inferred, and whether the belief is temporally outdated. Intuitively, the schema asks seven questions about each proposition: how deeply nested it is, whether it is true, who could know it, whether it is explicit or inferred, what it is about, how it was acquired, and what special framing applies. Additional schema-label details are provided in App.~\ref{app:atoms-coverage}. The schema consists of seven dimensions:
\begin{itemize}
    \item \textbf{Order} (\(o_i\); \(\{0,1,2,3\}\)) captures the recursive depth of belief attribution: a first-order belief is \textit{``Bob believes $X$,''} whereas a second-order belief, such as \textit{``Bob believes Alice thinks $X$''}, reasons about one actor's belief about another's belief. Beyond order~3, annotations on the human validation subset found insufficient reliable narrative evidence for further nesting.
    \item \textbf{Truth Status} (\(t_i\); \{True, False, Unknown\}) separates belief attribution from factual correctness, which is central to false-belief and appearance--reality phenomena.
    \item \textbf{Knowledge Access} (\(k_i\); \{Private, Shared, Public\}) encodes how information is distributed across actors, enabling explicit analysis of ignorance, asymmetry, and deception.
    \item \textbf{Representation} (\(r_i\); \{Explicit, Implicit\}) distinguishes directly stated beliefs from pragmatically inferred beliefs, helping isolate failures in pragmatic inference.
    \item \textbf{Content Type} (\(c_i\); \{Location, Contents/Physical State, Identity/Relation, Epistemic, Desire/Intention, Emotion, Trait/Value, Action/Event\}) identifies what is believed, allowing errors to be localized by semantic type (e.g., location tracking versus intention inference).
    \item \textbf{Mental Source} (\(m_i\); \{Narration, Perception, Memory, Testimony, Inference, Imagination, Unknown\}) tracks how a belief is acquired, supporting source-sensitive analysis.
    \item \textbf{Context} (\(x_i\); \{Deceptive, Temporal, Counterfactual, Neutral\}) captures story framing conditions that modulate interpretation and belief updating.
\end{itemize}

Together, these dimensions provide broad cognitive coverage while making model behavior auditable by reasoning depth, information access, content, source, and context~\citep{Flavell1986,BaronCohen1999FauxPas,Leslie1987,chen-etal-2024-tombench}. Stage~1 extracts the belief structure, while Stage~2 labels each proposition along these dimensions for finer-grained ToM analysis.


\section{Benchmark Construction}
\label{sec:dataset}
\label{sec:method}

\begin{figure}[t]
  \centering
  \includegraphics[width=0.98\linewidth]{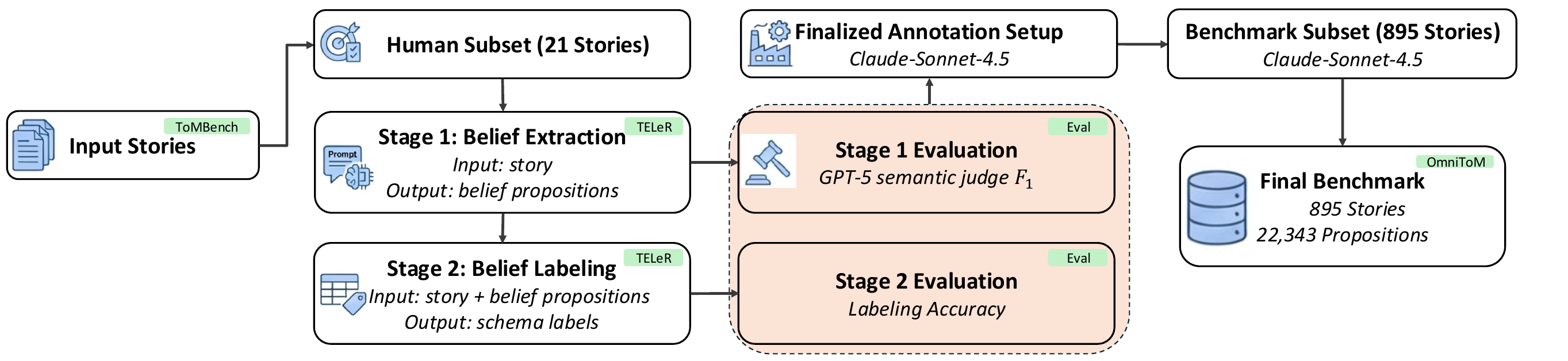}
  \caption{OmniToM human-validated benchmark-construction pipeline. Stories from seven ToMBench categories are split into a 21-story human-annotated validation subset and an 895-story benchmark subset. The validation subset provides human-annotated gold structures for the two-stage setup: Stage~1 scores belief extraction with GPT-5 semantic-judge \(F_1\), and Stage~2 scores belief labeling with labeling accuracy. These validation scores are used to select Claude-Sonnet-4.5 as the benchmark annotation model, which is then applied to the benchmark subset to produce the final OmniToM benchmark with 895 stories and 22{,}343 labeled belief propositions.}
  \label{fig:annotation-pipeline}
\end{figure}

\paragraph{Source Data and Validation Pipeline.}
\begin{wrapfigure}{r}{0.5\columnwidth}
  \vspace{-1.5em}
  \centering
  \includegraphics[width=0.98\linewidth]{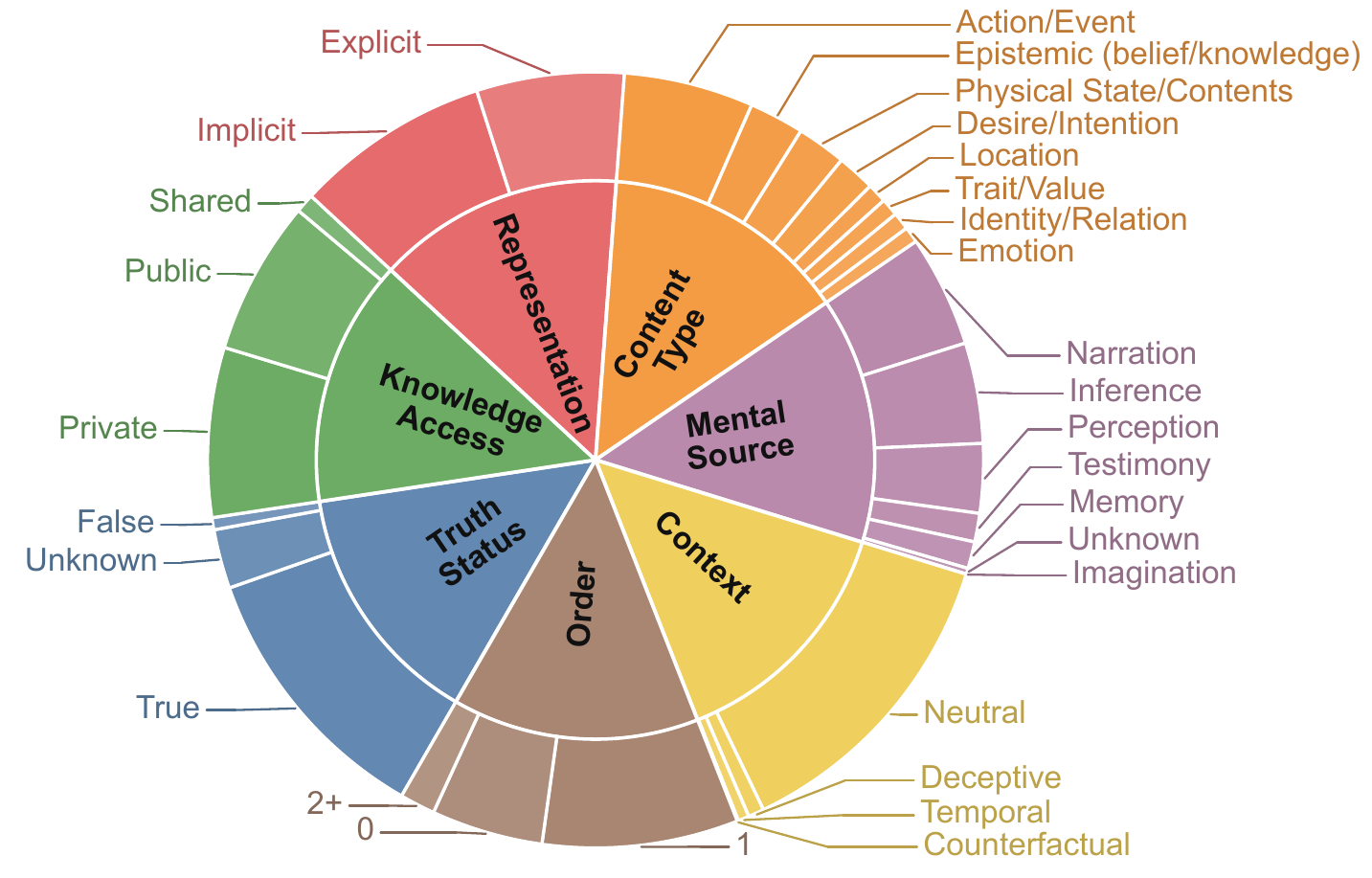}
  \caption{OmniToM label distribution statistics. Label frequencies over 156,401 labels.}
  \label{fig:dataset-stats}
  \vspace{-1.0em}
\end{wrapfigure}
OmniToM sources story text from ToMBench~\citep{chen-etal-2024-tombench}, a multi-task Theory-of-Mind benchmark of short social-reasoning stories. To support explicit belief-structure modeling from text alone, we retain seven high-volume categories with sufficiently self-contained mental-state evidence: Ambiguous Story Task, False Belief Task, Faux-pas Recognition Test, Hinting Task Test, Persuasion Story Task, Scalar Implicature Test, and Strange Story Task. These categories account for 916 source stories; we reserve 21 for human validation and model selection (three per category) and use the remaining 895 as the benchmark split. Constructing OmniToM requires identifying actor-specific belief propositions and assigning seven schema labels per proposition, making fully manual annotation costly and cognitively demanding at scale. We therefore use a human-validated, LLM-assisted pipeline that preserves human judgments while enabling benchmark-scale annotation. The human-annotated validation structures were developed through a multi-month, multi-team effort exceeding 1K person-hours across schema refinement, belief extraction, schema labeling, and validation.

\paragraph{Task-Conditioned LLM Annotations.}
We use TELeR~\citep{karmaker-santu-feng-2023-teler}, a taxonomy for systematic prompt design, to specify structured prompts for extraction, labeling, and evaluation; the taxonomy and our prompt-selection results are detailed in App.~\ref{app:prompt-engineering-teler}. Benchmark construction uses TELeR Level~4 prompts, which achieved the strongest validation performance in our prompt-selection sweep. These prompts combine a task directive, explicit sub-task decomposition, output criteria, and in-context examples, providing the structure needed for reliable table-form annotations. Stage~1 extraction uses category-conditioned prompts because the relevant belief dynamics differ across story types. For example, false-belief stories emphasize outdated beliefs and information access, hinting and strange stories emphasize pragmatic inference, and persuasion stories emphasize goals and influence. Stage~2 uses schema-invariant prompts because the seven-dimensional labeling schema is shared across all categories and is intended to provide a unified belief-level representation, so the same dimensions can be compared across stories with different kinds of social reasoning while keeping each label's meaning consistent.

\paragraph{Validation and Model Selection.}
We evaluate candidate benchmark annotation models on the human-annotated validation subset using the metrics defined in Sec.~\ref{sec:evaluation}. Stage~1 \(F_1\) requires semantic alignment between predicted and gold belief propositions. We therefore select the semantic judge by comparing candidate judges against human alignment decisions on 126 human-evaluated prediction tables. GPT-5 achieves the highest agreement with human judgments, with 87.76\% agreement and Cohen's \(\kappa=0.7572\)~\citep{cohen1960coefficient}, and is used as the independent Stage~1 judge. With this judge fixed, we evaluate candidate annotation models using Stage~1 \(F_1\) and Stage~2 labeling accuracy. Claude-Sonnet-4.5 achieves the strongest validation performance, with 72.88 Stage~1 \(F_1\) and 94.42\% Stage~2 accuracy, and is selected to construct the final benchmark. Full candidate comparisons and agreement statistics are reported in App.~\ref{app:construction-validation}.

\paragraph{Final Benchmark Statistics.}
Applying this finalized pipeline to the 895-story benchmark split yields 22{,}343 labeled belief propositions and 156{,}401 schema labels; Fig.~\ref{fig:dataset-stats} summarizes their distribution across the seven schema dimensions. Belief-order distribution is Order~0: 32.6\%, Order~1: 57.1\%, Order~2: 9.8\%, and Order~3: 0.5\%. Additional benchmark statistics are provided in
App.~\ref{app:construction-validation}.

\section{Evaluation Protocol}
\label{sec:evaluation}

\begin{table*}[t]
  \setlength{\belowcaptionskip}{4pt}
\centering
\caption{Example of Stage~1 semantic alignment for belief extraction. \texttt{MatchCount} indicates how each predicted belief aligns with the gold belief set, and vice versa: 0 denotes no semantic match, 1 denotes a one-to-one match, and 2 denotes a compound match. The \(mc_R=2\) example shows a gold belief whose content is covered by two predicted beliefs.}
\label{tab:judge-matching}
\setlength{\tabcolsep}{2pt}
\renewcommand{\arraystretch}{1.08}
\begin{minipage}{0.99\textwidth}
\footnotesize
\begin{tabular*}{\linewidth}{@{\extracolsep{\fill}}p{0.98\linewidth}@{}}
\toprule
\textit{Story: Xiao Hong wants to change to a bigger office, but that office is occupied by her colleague Xiao Li.} \\
\bottomrule
\end{tabular*}
\end{minipage}
\vspace{0.45em}
\begin{minipage}[t]{0.492\textwidth}
\raggedright
{\centering\small\textbf{Predicted Table \(\hat{\mathbf{B}}\)}\par}
\vspace{0.10em}
\scriptsize
\begin{tabular*}{\linewidth}{@{\extracolsep{\fill}}p{0.84\linewidth}r@{}}
\toprule
\textbf{Belief Proposition} & \(\mathbf{mc_P}\) \\
\midrule
\multicolumn{2}{@{}l}{\textit{Actor: world}} \\
Xiao Hong wants to change to a bigger office & 1 \\
The bigger office is occupied by Xiao Li & 1 \\
\midrule
\multicolumn{2}{@{}l}{\textit{Actor: Xiao Hong}} \\
Xiao Hong wants to change to a bigger office & 1 \\
The bigger office is occupied by Xiao Li & 1 \\
Xiao Hong needs to persuade Xiao Li to vacate the office & 1 \\
Xiao Li might vacate the office if persuaded & 1 \\
\bottomrule
\end{tabular*}
\end{minipage}%
\hfill
\begin{minipage}[t]{0.492\textwidth}
\raggedright
{\centering\small\textbf{Gold Table \(\mathbf{B}^*\)}\par}
\vspace{0.10em}
\scriptsize
\begin{tabular*}{\linewidth}{@{\extracolsep{\fill}}p{0.84\linewidth}r@{}}
\toprule
\textbf{Belief Proposition} & \(\mathbf{mc_R}\) \\
\midrule
\multicolumn{2}{@{}l}{\textit{Actor: world}} \\
Xiao Hong wants to change to a bigger office & 1 \\
The bigger office is occupied by Xiao Li & 1 \\
Xiao Hong and Xiao Li are colleagues & 0 \\
\midrule
\multicolumn{2}{@{}l}{\textit{Actor: Xiao Hong}} \\
Xiao Hong needs the bigger office that Xiao Li occupies & 2 \\
Xiao Hong must persuade Xiao Li to leave the bigger office & 1 \\
Xiao Li may swap offices if Xiao Hong persuades him & 1 \\
\bottomrule
\end{tabular*}
\end{minipage}
\end{table*}

OmniToM uses two evaluation protocols aligned with the two-stage benchmark formulation in Sec.~\ref{sec:framework}. Stage~1 measures \emph{belief-extraction completeness}: whether a model extracts the belief propositions relevant to the story's social dynamics. Stage~2 measures \emph{belief-labeling accuracy}: whether provided belief propositions are assigned the correct seven-dimensional schema labels. In both stages, scores are computed per story and macro-averaged across the corpus.

\paragraph{Stage~1 Evaluation: Belief-Extraction Completeness (\(F_1\)).}
\label{sec:judge}

Given predicted beliefs \(\hat{\mathbf{B}}\) and gold beliefs \(\mathbf{B}^*\), Stage~1 evaluates how well the predicted beliefs recover the gold belief structure. Exact string matching is too strict because equivalent beliefs can differ in wording or granularity: a compound predicted belief may cover multiple atomic gold beliefs, or several predicted beliefs may together express the content of one gold belief. We therefore use \texttt{MatchCount} to align predicted beliefs to gold beliefs. For a predicted belief \(b \in \hat{\mathbf{B}}\), \(mc_P(b)\) denotes how many gold beliefs it semantically matches for the same actor:
\[
mc_P(b)=
\begin{cases}
0, & \text{if } b \text{ matches no gold belief,}\\
1, & \text{if } b \text{ matches one gold belief,}\\
2 \text{ or } 3, & \text{if } b \text{ is compound and matches two or three distinct gold beliefs.}
\end{cases}
\]
One-to-many matches are used conservatively: counts of 2 or 3 are assigned only for compound beliefs that map to multiple distinct gold beliefs, and human validation found no cases requiring counts above 3 (App.~\ref{app:annotation-reliability}). We apply the same procedure in reverse to compute \(mc_R(b)\) for each gold belief \(b \in \mathbf{B}^*\). Thus, \(mc_P\) measures how predicted beliefs cover the gold set, while \(mc_R\) measures how gold beliefs are recovered by the predictions. As illustrated in Table~\ref{tab:judge-matching}, bidirectional alignment provides the matched belief counts used for precision and recall while preventing duplicate or paraphrased beliefs from receiving repeated credit.

For each story \(s\), let \(M_P^{(s)}\) and \(M_R^{(s)}\) denote the number of matched predicted and gold beliefs, respectively, i.e., beliefs with nonzero \texttt{MatchCount}. Let \(|\hat{\mathbf{B}}^{(s)}|\) and \(|\mathbf{B}^{*(s)}|\) denote the number of predicted and gold beliefs for story \(s\). We compute story-level precision, recall, and \(F_1\), and then macro-average across the \(S\) evaluated stories:
\[
P^{(s)} = \frac{M_P^{(s)}}{|\hat{\mathbf{B}}^{(s)}|}, \qquad
R^{(s)} = \frac{M_R^{(s)}}{|\mathbf{B}^{*(s)}|}, \qquad
F_1^{(s)} = \frac{2P^{(s)}R^{(s)}}{P^{(s)}+R^{(s)}}, \qquad
F_{1,\mathrm{macro}} = \frac{1}{S}\sum_{s=1}^{S} F_1^{(s)}.
\]

\paragraph{Stage~2 Evaluation: Belief-Labeling Accuracy.}

Given predicted schema labels \(\hat{y}^{(s)}_{i,d}\) and gold labels \(y^{*(s)}_{i,d}\), Stage~2 evaluates exact-match accuracy over the gold belief propositions. Here, \(s\) indexes the story, \(i\) indexes a gold belief within that story, \(d\) indexes one of the seven schema dimensions, and \(N^{(s)}_{\mathrm{gold}}\) is the number of gold beliefs in story \(s\). A prediction is counted as correct when \(\hat{y}^{(s)}_{i,d}=y^{*(s)}_{i,d}\). We compute per-dimension accuracy and overall story-level labeling accuracy as:
\[
\mathrm{Acc}^{(s)}_d =
\frac{1}{N^{(s)}_{\mathrm{gold}}}
\sum_{i=1}^{N^{(s)}_{\mathrm{gold}}}
\mathbf{1}\!\left[\hat{y}^{(s)}_{i,d}=y^{*(s)}_{i,d}\right],
\qquad
\mathrm{Acc}^{(s)}_{\mathrm{overall}}=\frac{1}{7}\sum_{d=1}^{7}\mathrm{Acc}^{(s)}_d.
\]
We report both per-dimension and overall Stage~2 accuracy by macro-averaging these story-level scores across stories.

\section{Experiments}
\label{sec:experiments}

\begin{table*}[t]
  \setlength{\belowcaptionskip}{4pt}
\centering
\normalsize
\caption{Main OmniToM benchmark results under zero-shot TELeR Level~3 prompts (\%). Stage~1 reports story-macro belief-extraction \(F_1\), and Endpoint QA reports story-macro accuracy; both are followed by symmetric margins for their 95\% story-bootstrap confidence intervals, rounded to two decimal places. Stage~2 reports story-macro belief-labeling accuracy by schema dimension and overall. GPT-5 is omitted from Stage~1 because it serves as the semantic judge. The best point estimate is bold, and the second-best is underlined.}
\label{tab:benchmark-main}
\resizebox{\textwidth}{!}{%
\begin{tabular}{l*{10}{r}}
\toprule
& \multicolumn{1}{c}{Stage~1} & \multicolumn{1}{c}{Endpoint QA} & \multicolumn{8}{c}{Stage~2: Belief-Labeling Accuracy} \\
\cmidrule(lr){2-2} \cmidrule(lr){3-3} \cmidrule(lr){4-11}
Model & \(F_1\) (95\% CI) & Accuracy (95\% CI) & Order & Status & Access & Repr. & CType & Source & Context & Overall \\
\midrule
\multicolumn{11}{l}{\textit{Closed-source models}} \\
Gemini-2.5 Pro & \(\mathbf{61.63 \pm 0.76}\) & \(\mathbf{80.67 \pm 2.15}\) & 96.23 & \underline{86.50} & 66.94 & \textbf{89.01} & \textbf{86.35} & \textbf{86.70} & 87.07 & \underline{85.54} \\
Gemini-2.5 Flash & \(55.75 \pm 1.10\) & \(\underline{77.30 \pm 2.32}\) & 95.56 & 84.97 & 71.34 & \underline{87.58} & \underline{85.97} & 84.10 & \underline{92.14} & \textbf{85.95} \\
GPT-5 & -- & -- & 95.18 & 82.72 & 66.85 & 83.42 & 79.96 & 83.02 & 88.83 & 82.85 \\
\midrule
\multicolumn{11}{l}{\textit{Open-weight models}} \\
Gemma-3 27B & \(\underline{58.03 \pm 0.98}\) & \(53.95 \pm 2.53\) & \underline{96.56} & 82.44 & 71.57 & 54.33 & 73.50 & 78.72 & 92.07 & 78.46 \\
Mistral-Small 24B & \(56.46 \pm 0.94\) & \(71.08 \pm 2.45\) & 95.13 & 82.22 & \textbf{74.59} & 62.79 & 76.01 & 84.82 & 91.90 & 81.06 \\
Mistral-Large 123B & \(53.53 \pm 0.92\) & \(73.50 \pm 2.42\) & \textbf{97.25} & \textbf{86.53} & \underline{74.14} & 72.87 & 82.83 & \underline{86.32} & \textbf{92.97} & 84.70 \\
Qwen3 32B & \(51.63 \pm 0.94\) & \(69.96 \pm 2.44\) & 96.42 & 82.43 & 73.91 & 62.45 & 71.27 & 76.84 & 90.81 & 79.16 \\
Llama-3.3 70B & \(47.12 \pm 0.99\) & \(71.28 \pm 2.45\) & 92.74 & 83.55 & 67.41 & 72.43 & 72.35 & 76.71 & 91.69 & 79.55 \\
Qwen3 8B & \(42.70 \pm 1.18\) & \(63.31 \pm 2.46\) & 73.38 & 67.17 & 57.94 & 63.77 & 51.43 & 61.49 & 74.62 & 64.26 \\
Llama-3.1 8B & \(37.21 \pm 1.18\) & \(60.79 \pm 2.58\) & 71.90 & 65.59 & 56.13 & 64.40 & 48.63 & 55.18 & 76.81 & 62.66 \\
\bottomrule
\end{tabular}%
}
\end{table*}

\paragraph{Experimental Setup and Baselines.}
We evaluate models under zero-shot TELeR Level~3 prompts~\citep{karmaker-santu-feng-2023-teler}. Level~3 provides a task directive and stepwise sub-tasks, but no in-context examples or category-specific evaluation criteria. This setting gives models the task definition without the stronger construction-time scaffolding used in Level~4 prompts. We benchmark ten models: Gemini-2.5 Flash, Gemini-2.5 Pro~\citep{comanici2025gemini25}, GPT-5~\citep{openai2025gpt5}, Gemma-3 27B~\citep{gemma2025gemma3}, Llama-3.1-8B-Instruct~\citep{grattafiori2024llama3}, Llama-3.3-70B-Instruct~\citep{meta2024llama33}, Mistral-Small-24B-Instruct-2501~\citep{mistralai2025small24b}, Mistral-Large-Instruct-2407~\citep{mistral2024large}, Qwen3-8B, and Qwen3-32B~\citep{qwen2025qwen3}. Open-weight evaluations were run on NVIDIA A100 80GB GPUs with 4-bit quantization where required, using approximately 48 GPU-hours per model. GPT-5 is the fixed semantic judge and is therefore omitted as a Stage~1 model. Stage~1 reports story-macro belief-extraction \(F_1\), and Stage~2 reports story-macro accuracy for each schema dimension and overall. Endpoint QA reports story-macro accuracy across the 895 source stories, averaging question correctness within each story and then averaging equally across stories. Stage~1 and Endpoint QA estimates include 95\% percentile confidence intervals from 10{,}000 source-story bootstrap resamples; endpoint resampling treats each story as a cluster of linked questions.

\begin{figure*}[t]
  \centering
  \begin{minipage}[b]{0.495\textwidth}
    \centering
    \includegraphics[width=0.97\linewidth]{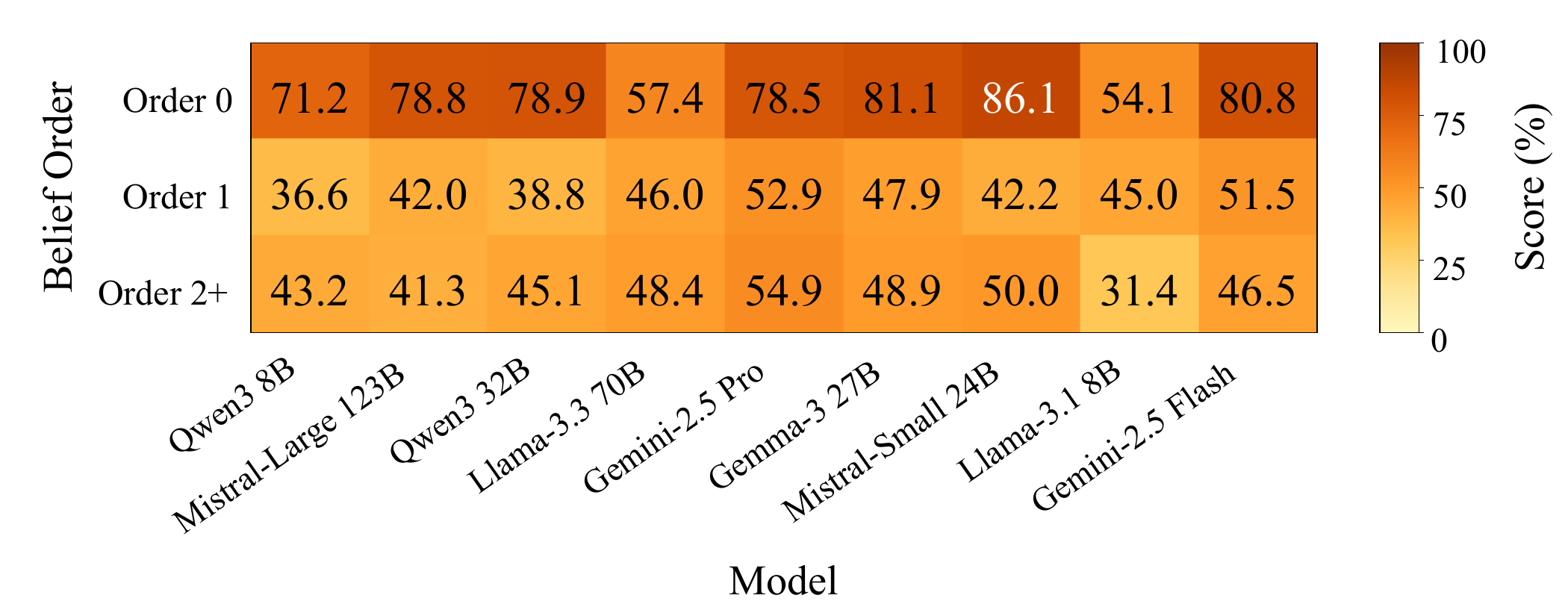}
  \end{minipage}
  \hfill
  \begin{minipage}[b]{0.495\textwidth}
    \centering
    \includegraphics[width=0.97\linewidth]{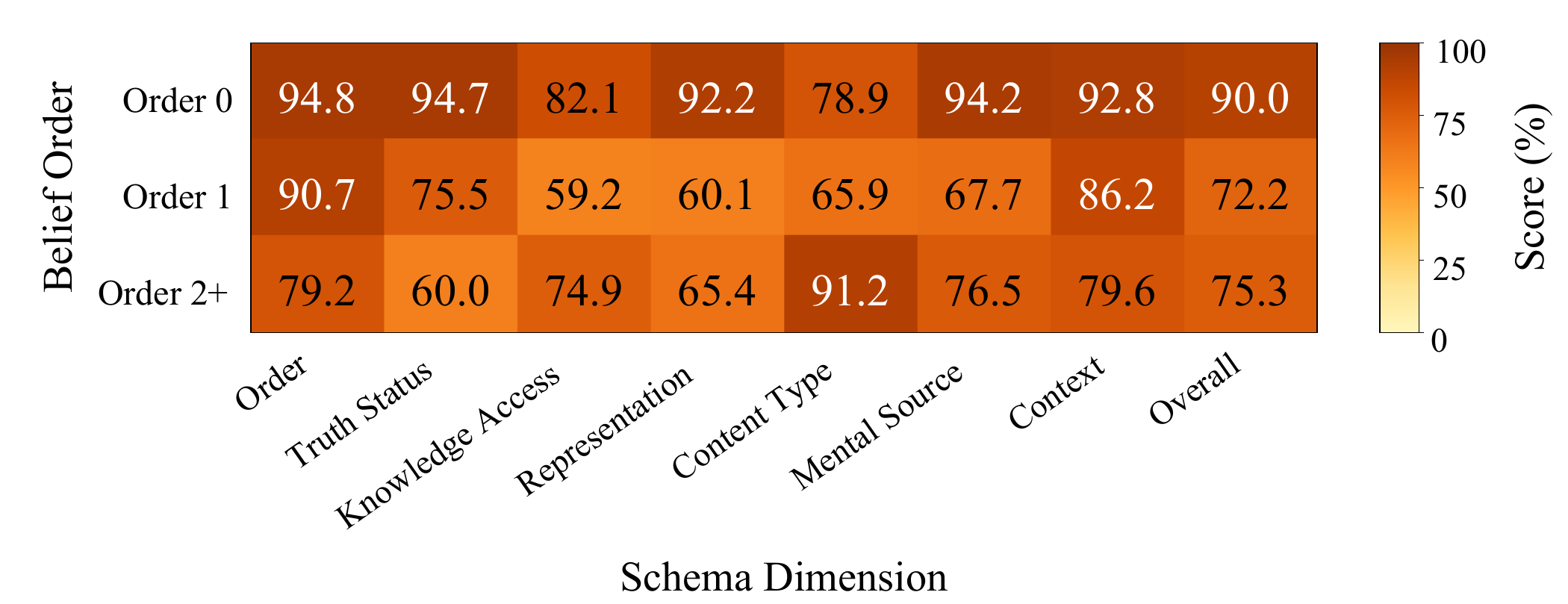}
  \end{minipage}
  \vspace{-0.2em}
  \begin{minipage}[t]{0.495\textwidth}
    \centering
    {\footnotesize (a)}
  \end{minipage}
  \hfill
  \begin{minipage}[t]{0.495\textwidth}
    \centering
    {\footnotesize (b)}
  \end{minipage}
  \caption{Order-wise analysis. (a) Stage~1 extraction \(F_1\) by model and belief-order bucket. (b) Stage~2 labeling accuracy by belief-order bucket and schema dimension.}
  \label{fig:information-tracking-analysis}
\end{figure*}

\paragraph{Main Results Summary.}
Table~\ref{tab:benchmark-main} and Fig.~\ref{fig:information-tracking-analysis} show the same information-tracking bottleneck in both stages. In Stage~1, models more reliably extract narrated facts than beliefs attributed to particular actors. In Stage~2, errors are concentrated in \textit{Knowledge Access} and, for some models, \textit{Representation}, especially for Order~1 beliefs. Together, these results point to difficulty tracking what each actor knows and how that information becomes a belief. Gemini-2.5 Pro has the strongest Stage~1 result at 61.63 story-macro \(F_1\). Gemini-2.5 Flash has the strongest overall Stage~2 accuracy at 85.95\%, followed by Gemini-2.5 Pro at 85.54\%.

\paragraph{Stage~1: Belief Extraction.}
Fig.~\ref{fig:information-tracking-analysis}a shows that models extract narrated Order~0 facts more reliably than actor-specific Order~1+ beliefs. Performance drops when a model must connect a fact in the story to what a particular actor knows or believes. This requires tracking what the actor perceived, missed, remembered, was told, or could infer. Higher-order beliefs add another step: the model must represent what one actor believes about another actor's beliefs. The broad Stage~1 ordering is supported by paired comparisons: Gemini-2.5 Pro is ahead of Gemma-3 27B, and Gemma-3 27B is ahead of Mistral-Small 24B, while the difference between Mistral-Small 24B and Gemini-2.5 Flash remains uncertain. Stage~1 extraction \(F_1\) varies across story families: classic False Belief scenarios, where actors retain outdated beliefs after missing a change, and Scalar Implicature, which involves reasoning over quantifiers or degree terms, are often among the stronger categories. Hinting and Strange Story are often weaker, requiring inference of indirectly communicated goals or interpretation of nonliteral language such as jokes and sarcasm. Additional story-family, belief-order, and paired-comparison results are reported in App.~\ref{app:stage2-category-dimension-analysis}.

\paragraph{Stage~2: Belief Labeling.}
\begin{wrapfigure}{R}{0.5\columnwidth}
  \vspace{-1.5em}
  \centering
  \includegraphics[width=0.98\linewidth]{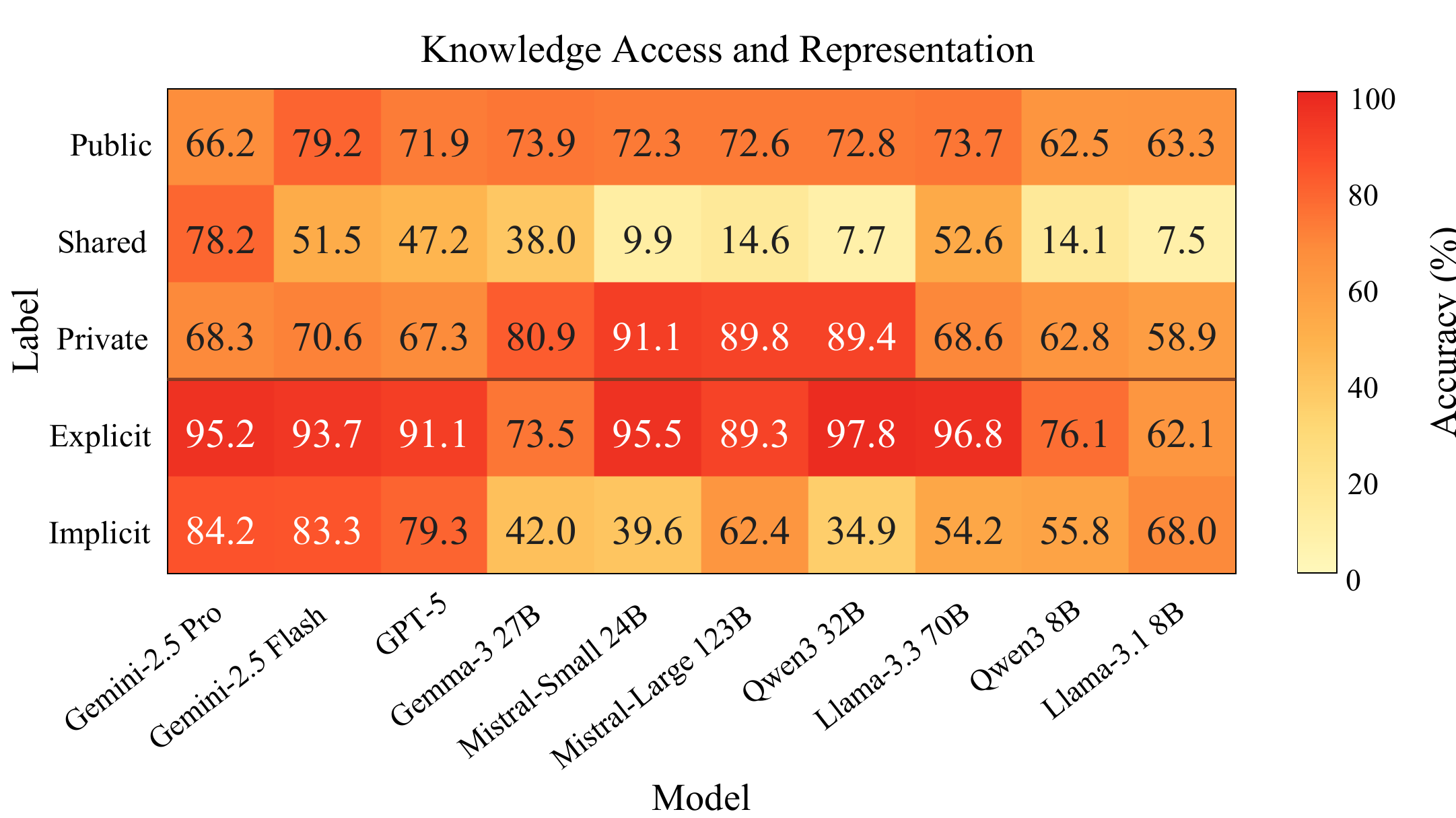}
  \caption{Stage~2 story-macro accuracy by gold \textit{Knowledge Access} and \textit{Representation} label.}
  \label{fig:stage2-knowledge-access-representation}
  \vspace{-0.5em}
\end{wrapfigure}
Stage~2 makes the information-tracking difficulty more specific. As shown in Table~\ref{tab:benchmark-main}, \textit{Knowledge Access} is among the weaker dimensions across models (56.13--74.59\%), while \textit{Representation} is also weak for several models (54.33--89.01\%). Fig.~\ref{fig:information-tracking-analysis}b shows that Order~1 beliefs have the lowest overall labeling accuracy (72.2\%), with especially low accuracy on these dimensions (59.2\% and 60.1\%, respectively). The label-level results in Fig.~\ref{fig:stage2-knowledge-access-representation} sharpen this pattern. Most models score lower on Shared Knowledge Access, and lower on Implicit Representation. This points to difficulty tracking which information actors share and which beliefs they must infer from the story. Furthermore, Order~2+ labeling accuracy is higher overall (75.3\%), so accuracy does not decrease steadily with belief depth. Additionally, some labels are more common than others. For example, 91.81\% of Context labels are Neutral, and a majority-label baseline reaches 91.54\% accuracy on this dimension. We therefore report category and dimension diagnostics, including Stage~2 macro-\(F_1\) across labels and balanced accuracy for imbalanced labels, in App.~\ref{app:stage2-category-dimension-analysis}.

\paragraph{Endpoint QA and Explicit Belief Modeling.}
To see whether models that reconstruct beliefs well also perform well on endpoint QA, we compared Stage~1 extraction with ToMBench's endpoint QA questions linked to the same stories. We find that the rankings differ. For example, Gemma-3 27B ranks second on Stage~1 extraction (58.03 \(F_1\)) but ninth on endpoint QA (53.95\%). This suggests that the two evaluations capture different aspects of performance. In our qualitative analysis, we found cases where models answered all endpoint questions correctly but still made errors in their belief representations, such as attributing knowledge of an event to a character who was absent when it happened. These examples show how a correct endpoint answer can coexist with errors in what characters know or believe. Further comparisons and examples appear in App.~\ref{app:endpoint-qa-comparison} and App.~\ref{app:qualitative-comparison}.

\paragraph{Reference Model Sensitivity.}
To test whether Stage~1 results depend on the model used to generate the reference beliefs, we replaced the Claude-generated references with references generated by Mistral-Large, while keeping model predictions and the GPT-5 judge fixed. In the 21-story annotation-model selection, Mistral-Large was the runner-up to Claude, with Stage~1 \(F_1\) scores of 70.13 and 72.88, respectively (App.~\ref{app:validation-results}). We left both Mistral models out of this comparison because Mistral-Large generated the alternative references, leaving seven models. The alternative reference raised their scores by 1.56--7.12 points. Three models kept their ranks, while two adjacent pairs switched places: Gemma-3 27B and Gemini-2.5 Pro exchanged first and second, and Qwen3 32B and Gemini-2.5 Flash exchanged third and fourth. The broad ordering stayed similar, but using the second-ranked annotation model shifted absolute scores and some close rankings. Detailed results are reported in App.~\ref{app:reference-sensitivity}.

\paragraph{Semantic Judge Sensitivity.}
We also tested whether Stage~1 results depend on the semantic judge by replacing GPT-5 with Llama-3.3 70B, while keeping model predictions and Claude references fixed. In the human validation used to select the judge, GPT-5 had higher agreement with human decisions than Llama-3.3 (87.76\% versus 75.85\%; App.~\ref{app:annotation-reliability}). We left both Llama models out of this comparison to avoid evaluating Llama models with a Llama judge. The Llama judge raised \(F_1\) scores by 14.30--18.78 points. Three models kept their ranks, while Mistral-Small and Mistral-Large switched places. The broad ordering stayed similar, but the judge choice shifted absolute scores and one close pair's ranking. Detailed results are reported in App.~\ref{app:judge-sensitivity}.

\section{Conclusion and Limitations}
\label{sec:conclusion-limitations}

\paragraph{Conclusion.}
We introduced OmniToM, a benchmark that evaluates Theory of Mind through explicit belief-structure modeling rather than endpoint question answering. By requiring models to extract and label multi-actor belief structures with a seven-dimensional schema, OmniToM makes the mental-state representations behind social reasoning directly inspectable. We built the benchmark using a human-validated, LLM-assisted pipeline that makes dense belief annotation scalable. Across 895 stories and 22{,}343 labeled belief propositions, our evaluation of diverse open- and closed-source models identifies an actor-specific information-tracking bottleneck: models often recover story facts but struggle to determine what each actor knows or believes. This difficulty appears in both stages: extraction performance drops when models must recover actor-specific beliefs, and labeling errors concentrate on \textit{Knowledge Access} and \textit{Representation}. Most models are also less accurate on shared knowledge and implicit beliefs, suggesting difficulty tracking what actors share and infer from a story. Sensitivity analyses show that the broad model ordering remains similar across alternative reference beliefs and semantic judges, although absolute scores and close rankings can change. By making the mental-state representations behind social reasoning auditable, OmniToM supports more rigorous evaluation and the development of more reliable social reasoning models.

\paragraph{Limitations.}
OmniToM evaluates text-based ToM reasoning over short, self-contained stories drawn from seven ToMBench categories. It therefore does not cover multimodal or interactive settings, long-horizon information tracking, dense temporal structure, or belief nesting beyond order~3, and it inherits the topical scope of its source stories. The seven-dimensional schema also requires human judgment, although agreement and validation analyses show generally strong reliability. Finally, Stage~1 relies on semantic matching between generated and reference beliefs. Human validation and alternative-judge experiments help characterize this dependence, but absolute extraction scores can still vary with the semantic evaluator. Future work can extend OmniToM to richer settings and larger human-audited belief structures.

\begin{ack}
We thank the student annotators at the University of Central Florida for their contributions to benchmark annotation and validation. We also acknowledge NSF ACCESS for supporting the computing resources used in this work.
\end{ack}

\bibliographystyle{plainnat}
\bibliography{references}

\clearpage
\appendix

\clearpage
\section*{Supplementary Material}

\noindent This supplement provides additional experimental, methodological, and qualitative results supporting the main paper. We first discuss broader impacts. Appendix~\ref{app:additional-results} reports additional experiments and extended benchmark results, including the endpoint-QA comparison, benchmark-reference sensitivity, semantic-judge sensitivity, and detailed Stage~1 and Stage~2 analyses. Appendix~\ref{app:atoms-coverage} defines the OmniToM schema derivation and labeling rules. Appendix~\ref{app:construction-validation} documents source filtering, validation, annotation reliability, semantic-judge validation, and release format. Appendix~\ref{app:prompt-engineering-teler} records the TELeR prompt protocol used for construction, evaluation, and semantic judging. Appendix~\ref{app:qualitative-comparison} provides qualitative comparisons between endpoint QA and the separately elicited Stage~1 belief representation. Appendix~\ref{app:category-examples} provides worked annotation examples for each retained story category.

\section*{Broader Impacts}
\phantomsection
\label{sec:broader-impacts}

OmniToM supports more process-sensitive evaluation of social reasoning by shifting attention from endpoint question answering to explicit belief-structure modeling. This can help researchers analyze where models fail to track information access, communication, and actor-specific beliefs, rather than treating a correct final answer as sufficient evidence of ToM competence. By making intermediate mental-state representations inspectable, OmniToM may support more transparent evaluation of model behavior in social-reasoning settings.

At the same time, OmniToM should not be used to certify real-world social intelligence, interpersonal reliability, or deployed-agent safety. Story-based benchmark scores do not establish that a model can reason appropriately in open-ended human interaction, clinical, educational, legal, or other high-stakes settings. The benchmark is intended as an evaluation tool for controlled text-based scenarios, not as evidence that a deployed system possesses human-like social understanding.

The benchmark also has dual-use considerations. Better tools for evaluating mental-state tracking may support safer human-AI interaction and more transparent model analysis, but they could also be used to optimize systems for persuasion, deception, or strategic modeling of users' beliefs. We therefore frame OmniToM as an evaluation benchmark rather than a training objective, deployment guarantee, or safety certification, and recommend reporting benchmark scores together with the limitations described in Sec.~\ref{sec:conclusion-limitations}.

\appsection{Additional Experimental Results}
\label{app:additional-results}

This appendix reports extended results for the experiments in Section~\ref{sec:experiments}. It examines performance by story family and label, compares Stage~1 belief extraction with endpoint question answering, and tests sensitivity to the reference beliefs and semantic judge. It also reports audits of extraction volume, semantic matches, and output coverage.

Unless stated otherwise, 95\% confidence intervals are percentile intervals from bootstrap resampling at the story level. Paired model comparisons use the same resampled stories for both models. Their bootstrap win share is the percentage of resamples in which the model with the higher point estimate also has the higher score. The intervals describe variation across stories and are not adjusted for multiple comparisons. Other tables retain their experiment-specific intervals.

\subsection{Performance Breakdowns}
\label{app:stage2-category-dimension-analysis}

\subsubsection{Stage~1 and Stage~2 results by story family}

We break down both stages by story family to see whether the overall patterns hold across different kinds of social reasoning. In Stage~1, False Belief and Scalar Implicature are often among the stronger categories, while Hinting and Strange Story are often lower, though the pattern varies by model. These categories involve different demands: False Belief asks models to track an actor who missed a change, Scalar Implicature requires interpreting quantifiers or degree terms, and Hinting and Strange Story require recovering indirect or nonliteral meaning. Table~\ref{tab:stage1-family-results} reports Stage~1 extraction \(F_1\), and Table~\ref{tab:stage2-dim-category-merged} reports Stage~2 accuracy by dimension and story family. Stage~2 results also vary by dimension, with no family consistently easiest or hardest.

\begin{table}[H]
  \setlength{\belowcaptionskip}{4pt}
\centering
\caption{Stage~1 belief-extraction \(F_1\) by ToMBench story family (\%). Columns correspond, from left to right, to Ambiguous Story Task, False Belief Task, Faux-pas Recognition Test, Hinting Task Test, Persuasion Story Task, Scalar Implicature Test, and Strange Story Task.}
\label{tab:stage1-family-results}
\begin{tabular*}{\textwidth}{@{\extracolsep{\fill}}lrrrrrrrr@{}}
\toprule
\textbf{Model} & \textbf{AST} & \textbf{FBT} & \textbf{FPT} & \textbf{HT} & \textbf{PST} & \textbf{SIT} & \textbf{SST} & \textbf{Overall} \\
\midrule
Gemini-2.5 Pro & \textbf{54.33} & \underline{71.55} & \textbf{62.11} & \textbf{55.10} & \textbf{64.84} & 63.74 & \textbf{60.21} & \textbf{61.63} \\
Gemma-3 27B & 48.72 & \textbf{72.39} & 56.05 & 45.46 & 56.72 & \textbf{68.76} & 55.77 & \underline{58.03} \\
Mistral-Small 24B & \underline{52.97} & 54.58 & \underline{59.79} & 48.32 & 56.97 & \underline{66.20} & 53.17 & 56.46 \\
Gemini-2.5 Flash & 42.40 & 56.48 & 57.78 & \underline{50.34} & \underline{58.55} & 62.91 & \underline{56.31} & 55.75 \\
Mistral-Large 123B & 47.75 & 71.28 & 53.66 & 41.78 & 58.53 & 57.38 & 48.35 & 53.53 \\
Qwen3 32B & 46.88 & 57.32 & 53.38 & 41.41 & 57.25 & 56.51 & 48.67 & 51.63 \\
Llama-3.3 70B & 37.51 & 64.07 & 46.33 & 36.27 & 47.23 & 57.70 & 41.58 & 47.12 \\
Qwen3 8B & 38.72 & 49.70 & 43.59 & 37.14 & 47.87 & 47.20 & 37.60 & 42.70 \\
Llama-3.1 8B & 26.34 & 48.29 & 35.80 & 30.85 & 35.01 & 53.52 & 30.08 & 37.21 \\
\bottomrule
\end{tabular*}
\end{table}

\setlength{\LTpre}{6pt}
\setlength{\LTpost}{6pt}
\setlength{\LTleft}{0pt}
\setlength{\LTright}{0pt}
\setlength{\LTcapwidth}{\textwidth}
\begingroup
\setlength{\tabcolsep}{1pt}
\renewcommand{\arraystretch}{1.05}
\begin{longtable}{@{}>{\raggedright\arraybackslash}p{0.263\textwidth}*{7}{>{\raggedleft\arraybackslash}p{0.10\textwidth}}@{}}
\caption{\normalsize Stage~2 belief-labeling accuracy (\%). Consolidated view of all seven schema dimensions across the seven story categories. Category abbreviations: AST (Ambiguous Story Task), FBT (False Belief Task), FPT (Faux-pas Recognition Test), HT (Hinting Task Test), PST (Persuasion Story Task), SIT (Scalar Implicature Test), SST (Strange Story Task).}\label{tab:stage2-dim-category-merged}\\[0.4em]
\textbf{Model} & \textbf{AST} & \textbf{FBT} & \textbf{FPT} & \textbf{HT} & \textbf{PST} & \textbf{SIT} & \textbf{SST} \\
\hline
\endfirsthead
\caption[]{\normalsize Stage~2 belief-labeling accuracy (\%). (continued)}\\[0.4em]
\hline
\textbf{Model} & \textbf{AST} & \textbf{FBT} & \textbf{FPT} & \textbf{HT} & \textbf{PST} & \textbf{SIT} & \textbf{SST} \\
\hline
\endhead
\hline
\endfoot
\hline
\endlastfoot
\multicolumn{8}{@{}l}{\footnotesize\textbf{Order}} \\
Gemini-2.5 Pro & 92.89 & \textbf{100.00} & \underline{95.60} & 92.43 & 95.28 & \textbf{100.00} & 95.94 \\
Gemini-2.5 Flash & 92.05 & 99.77 & 93.83 & 90.37 & 94.94 & 99.32 & \underline{96.46} \\
GPT-5 & \underline{95.72} & 99.90 & 93.70 & 90.67 & 95.71 & 97.65 & 93.84 \\
Gemma-3 27B & 95.04 & \underline{99.91} & 95.50 & \underline{95.23} & \textbf{98.14} & 99.31 & 94.28 \\
Mistral-Small 24B & 93.34 & \textbf{100.00} & 92.34 & 92.48 & 93.56 & 99.46 & 94.38 \\
Mistral-Large 123B & 95.54 & \textbf{100.00} & \textbf{95.72} & \textbf{95.96} & \underline{96.90} & \textbf{100.00} & \textbf{96.58} \\
Qwen3 32B & \textbf{95.88} & \textbf{100.00} & 95.26 & 94.52 & 95.66 & \underline{99.91} & 94.46 \\
Llama-3.3 70B & 88.20 & 99.79 & 91.70 & 89.44 & 89.40 & 96.95 & 92.31 \\
Qwen3 8B & 57.14 & 70.54 & 71.01 & 80.43 & 82.08 & 73.20 & 76.67 \\
Llama-3.1 8B & 64.00 & 59.69 & 77.41 & 70.20 & 75.94 & 81.38 & 69.44 \\
\hline
\multicolumn{8}{@{}l}{\footnotesize\textbf{Truth Status}} \\
Gemini-2.5 Pro & 82.04 & \textbf{93.50} & \underline{87.96} & 76.85 & \textbf{89.58} & \underline{81.13} & \underline{91.54} \\
Gemini-2.5 Flash & 78.16 & \underline{91.73} & \textbf{88.34} & 71.62 & \underline{88.81} & 79.35 & \textbf{91.56} \\
GPT-5 & \textbf{89.30} & 89.81 & 81.41 & 84.11 & 81.98 & 75.25 & 82.39 \\
Gemma-3 27B & 86.45 & 86.49 & 86.33 & 83.74 & 86.00 & 66.32 & 85.68 \\
Mistral-Small 24B & 79.25 & 91.09 & 85.12 & 68.65 & 84.26 & 76.45 & 87.39 \\
Mistral-Large 123B & \underline{86.92} & 84.76 & 87.48 & \textbf{85.91} & 85.80 & \textbf{84.48} & 88.68 \\
Qwen3 32B & 86.00 & 84.79 & 86.15 & \underline{84.19} & 83.00 & 71.42 & 84.16 \\
Llama-3.3 70B & 83.59 & 88.91 & 87.25 & 79.07 & 86.80 & 71.38 & 88.16 \\
Qwen3 8B & 55.09 & 67.94 & 65.03 & 68.36 & 71.89 & 64.71 & 73.04 \\
Llama-3.1 8B & 58.82 & 53.95 & 72.83 & 56.94 & 67.42 & 72.71 & 67.32 \\
\hline
\multicolumn{8}{@{}l}{\footnotesize\textbf{Knowledge Access}} \\
Gemini-2.5 Pro & 68.71 & \textbf{71.60} & \underline{61.16} & 70.92 & 74.94 & 56.72 & 69.81 \\
Gemini-2.5 Flash & 66.32 & \underline{63.09} & \textbf{64.13} & 78.36 & 90.71 & 68.56 & 72.12 \\
GPT-5 & \underline{72.94} & 58.08 & 59.00 & 73.80 & 86.12 & 55.02 & 69.88 \\
Gemma-3 27B & 68.37 & 54.72 & 54.35 & 79.06 & \underline{91.67} & 76.31 & 76.25 \\
Mistral-Small 24B & \textbf{73.43} & 55.76 & 54.19 & \textbf{79.97} & 87.99 & \textbf{77.96} & \textbf{86.57} \\
Mistral-Large 123B & 71.78 & 56.49 & 54.08 & \textbf{79.97} & 89.41 & 76.87 & \underline{85.27} \\
Qwen3 32B & 71.75 & 57.34 & 54.52 & \underline{79.92} & 87.69 & \underline{77.05} & 84.30 \\
Llama-3.3 70B & 65.66 & 54.10 & 53.03 & 74.48 & \textbf{92.53} & 63.57 & 72.02 \\
Qwen3 8B & 45.39 & 39.47 & 43.82 & 72.57 & 77.25 & 53.90 & 69.14 \\
Llama-3.1 8B & 49.43 & 42.96 & 51.65 & 57.19 & 70.51 & 64.99 & 54.68 \\
\hline
\multicolumn{8}{@{}l}{\footnotesize\textbf{Representation}} \\
Gemini-2.5 Pro & \underline{89.69} & \underline{95.55} & \textbf{81.30} & \underline{92.16} & \textbf{88.19} & \textbf{88.45} & \textbf{90.19} \\
Gemini-2.5 Flash & \textbf{90.05} & \textbf{96.34} & 79.42 & \textbf{92.94} & 86.48 & \underline{83.12} & \underline{89.14} \\
GPT-5 & 77.68 & 88.88 & 73.32 & 90.97 & \underline{86.71} & 79.50 & 88.22 \\
Gemma-3 27B & 41.90 & 48.33 & 45.78 & 56.65 & 74.40 & 59.92 & 54.20 \\
Mistral-Small 24B & 44.27 & 85.76 & 75.85 & 68.09 & 56.83 & 59.65 & 54.39 \\
Mistral-Large 123B & 68.39 & 80.30 & 68.78 & 78.88 & 69.14 & 80.74 & 67.30 \\
Qwen3 32B & 59.03 & 51.37 & 75.51 & 71.34 & 55.77 & 57.22 & 63.04 \\
Llama-3.3 70B & 61.14 & 58.94 & \underline{79.80} & 76.84 & 83.42 & 70.67 & 73.05 \\
Qwen3 8B & 48.29 & 51.08 & 56.64 & 76.31 & 72.24 & 64.09 & 71.67 \\
Llama-3.1 8B & 58.02 & 54.74 & 56.42 & 68.22 & 74.62 & 70.11 & 66.54 \\
\hline
\multicolumn{8}{@{}l}{\footnotesize\textbf{Content Type}} \\
Gemini-2.5 Pro & \textbf{85.98} & \underline{91.25} & \textbf{88.34} & 84.33 & 73.29 & \underline{92.25} & \textbf{85.56} \\
Gemini-2.5 Flash & 82.84 & 90.10 & \underline{85.41} & \textbf{85.65} & \underline{75.89} & \textbf{93.00} & \underline{85.55} \\
GPT-5 & 80.99 & 86.53 & 84.40 & 78.73 & 73.04 & 77.64 & 78.90 \\
Gemma-3 27B & 73.25 & 81.17 & 71.60 & 77.43 & 65.31 & 67.42 & 77.80 \\
Mistral-Small 24B & 73.99 & 78.96 & 68.19 & 77.66 & 72.99 & 86.03 & 74.09 \\
Mistral-Large 123B & \underline{83.39} & 86.71 & 76.32 & \underline{84.46} & \textbf{76.35} & 89.61 & 82.44 \\
Qwen3 32B & 75.77 & \textbf{91.76} & 64.84 & 75.14 & 68.84 & 61.41 & 70.56 \\
Llama-3.3 70B & 74.35 & 87.51 & 59.51 & 74.52 & 66.37 & 75.39 & 72.59 \\
Qwen3 8B & 47.18 & 58.46 & 45.29 & 59.50 & 52.64 & 42.15 & 56.77 \\
Llama-3.1 8B & 42.69 & 57.42 & 43.03 & 50.08 & 45.31 & 50.30 & 50.78 \\
\hline
\multicolumn{8}{@{}l}{\footnotesize\textbf{Mental Source}} \\
Gemini-2.5 Pro & 89.32 & \textbf{92.98} & \textbf{83.71} & 89.66 & \textbf{88.81} & \underline{81.18} & 86.26 \\
Gemini-2.5 Flash & 86.93 & 86.69 & 75.22 & 89.38 & 84.74 & 81.04 & \textbf{87.06} \\
GPT-5 & \underline{89.68} & 86.16 & \underline{80.65} & 87.96 & 83.18 & 74.02 & 84.24 \\
Gemma-3 27B & 83.70 & 88.66 & 66.28 & 82.16 & 76.38 & 78.16 & 80.07 \\
Mistral-Small 24B & 89.50 & 89.99 & 75.90 & \underline{90.15} & 84.80 & \textbf{82.06} & 85.78 \\
Mistral-Large 123B & \textbf{91.91} & \underline{91.00} & 78.95 & \textbf{91.98} & \underline{88.66} & 81.06 & \underline{86.62} \\
Qwen3 32B & 85.00 & 84.75 & 62.87 & 83.47 & 69.21 & 75.65 & 80.12 \\
Llama-3.3 70B & 79.16 & 87.10 & 77.05 & 84.10 & 76.96 & 63.27 & 76.78 \\
Qwen3 8B & 52.74 & 63.76 & 48.01 & 74.72 & 74.25 & 53.78 & 67.19 \\
Llama-3.1 8B & 53.82 & 51.15 & 48.74 & 62.83 & 65.27 & 47.13 & 59.70 \\
\hline
\multicolumn{8}{@{}l}{\footnotesize\textbf{Context}} \\
Gemini-2.5 Pro & 91.42 & 79.57 & 92.35 & 90.28 & 97.36 & 82.92 & 81.64 \\
Gemini-2.5 Flash & 94.65 & 86.70 & 97.15 & \textbf{95.27} & 97.54 & 85.79 & \textbf{90.73} \\
GPT-5 & 95.01 & \textbf{90.00} & 91.31 & 91.50 & 95.66 & 83.02 & 83.48 \\
Gemma-3 27B & 94.86 & 88.45 & 98.36 & 94.26 & \textbf{98.81} & 90.09 & 85.40 \\
Mistral-Small 24B & \underline{96.10} & 84.21 & 98.39 & \underline{94.57} & 98.68 & \textbf{90.73} & 85.46 \\
Mistral-Large 123B & \textbf{96.53} & \underline{88.71} & \textbf{98.65} & 94.16 & \underline{98.74} & \underline{90.72} & \underline{87.78} \\
Qwen3 32B & 95.59 & 84.70 & 95.32 & 93.23 & 98.66 & 89.59 & 84.38 \\
Llama-3.3 70B & 94.25 & 84.09 & \underline{98.54} & 94.42 & 98.67 & 90.65 & 85.54 \\
Qwen3 8B & 62.19 & 60.92 & 76.54 & 85.07 & 88.78 & 75.36 & 73.38 \\
Llama-3.1 8B & 72.09 & 59.47 & 86.56 & 80.11 & 87.12 & 82.15 & 70.07 \\
\hline
\end{longtable}
\endgroup

\FloatBarrier

\subsubsection{Stage~1 uncertainty and belief-order results}

Table~\ref{tab:stage1-paired-comparisons} compares neighboring models in the Stage~1 ranking. Table~\ref{tab:stage1-order-results} shows extraction performance by belief order, including the values behind Fig.~\ref{fig:information-tracking-analysis}a.

\begin{table}[H]
  \setlength{\belowcaptionskip}{4pt}
\centering
\caption{Selected paired Stage~1 comparisons among neighboring models. Win share is the fraction of paired bootstrap resamples in which the higher-point-score model remains ahead.}
\label{tab:stage1-paired-comparisons}
\begin{adjustbox}{max width=\textwidth,center}
\begin{tabular}{@{}lrrr@{}}
\toprule
\textbf{Comparison} & \textbf{\(\Delta F_1\)} & \textbf{95\% paired CI} & \textbf{Win share} \\
\midrule
Gemini-2.5 Pro minus Gemma-3 27B & +3.61 & $\pm$ 1.03 & 100.00\% \\
Gemma-3 27B minus Mistral-Small 24B & +1.56 & $\pm$ 1.09 & 99.77\% \\
Mistral-Small 24B minus Gemini-2.5 Flash & +0.71 & $\pm$ 1.20 & 88.16\% \\
Gemini-2.5 Flash minus Mistral-Large 123B & +2.22 & $\pm$ 1.24 & 100.00\% \\
Mistral-Large 123B minus Qwen3 32B & +1.90 & $\pm$ 1.01 & 100.00\% \\
\bottomrule
\end{tabular}
\end{adjustbox}
\end{table}

\begin{table}[H]
  \setlength{\belowcaptionskip}{4pt}
\centering
\caption{Stage~1 belief-extraction \(F_1\) by reference-belief order (\%). Order~2+ pools second- and higher-order beliefs.}
\label{tab:stage1-order-results}
\begin{adjustbox}{max width=\textwidth,center}
\begin{tabular}{@{}lrrr@{}}
\toprule
\textbf{Model} & \textbf{Order 0} & \textbf{Order 1} & \textbf{Order 2+} \\
\midrule
Gemini-2.5 Pro & 78.49 & \textbf{52.90} & \textbf{54.94} \\
Gemma-3 27B & \underline{81.08} & 47.95 & 48.86 \\
Mistral-Small 24B & \textbf{86.14} & 42.25 & \underline{50.00} \\
Gemini-2.5 Flash & 80.75 & \underline{51.49} & 46.49 \\
Mistral-Large 123B & 78.76 & 41.95 & 41.34 \\
Qwen3 32B & 78.94 & 38.75 & 45.10 \\
Llama-3.3 70B & 57.44 & 46.04 & 48.44 \\
Qwen3 8B & 71.23 & 36.60 & 43.21 \\
Llama-3.1 8B & 54.12 & 45.04 & 31.44 \\
\bottomrule
\end{tabular}
\end{adjustbox}
\end{table}

Most adjacent model differences are supported, although the interval for Mistral-Small 24B versus Gemini-2.5 Flash includes zero. Across models, Order~0 facts are generally easier to extract than actor-specific beliefs. Performance drops on Order~1 beliefs, while Order~2+ results vary by model rather than declining steadily with belief depth.

\FloatBarrier

\subsubsection{Stage~2 aggregate and imbalance-aware results}

We report Stage~2 accuracy for each schema dimension and overall, with confidence intervals for each model. The paired comparisons focus on neighboring models with similar overall scores.

\begin{table}[H]
  \setlength{\belowcaptionskip}{4pt}
\centering
\caption{Stage~2 story-macro accuracy and 95\% story-bootstrap intervals, in percent. Overall is the macro-average of the seven dimensions within each story, then averaged across stories. Part 1 of 2.}
\label{tab:stage2-dimension-intervals}
\begin{adjustbox}{max width=\textwidth,center}
\begin{tabular}{@{}lrrrr@{}}
\toprule
\textbf{Model} & \textbf{Order} & \textbf{Truth} & \textbf{Knowledge Access} & \textbf{Representation} \\
\midrule
Gemini-2.5 Pro & 96.23 $\pm$ 0.40 & \underline{86.50 $\pm$ 0.71} & 66.94 $\pm$ 1.06 & \textbf{89.01 $\pm$ 0.68} \\
Gemini-2.5 Flash & 95.56 $\pm$ 0.61 & 84.97 $\pm$ 0.89 & 71.34 $\pm$ 1.09 & \underline{87.58 $\pm$ 0.83} \\
GPT-5 & 95.18 $\pm$ 0.99 & 82.72 $\pm$ 1.09 & 66.85 $\pm$ 1.38 & 83.42 $\pm$ 1.49 \\
Gemma-3 27B & \underline{96.56 $\pm$ 0.36} & 82.44 $\pm$ 0.97 & 71.57 $\pm$ 1.10 & 54.33 $\pm$ 1.81 \\
Mistral-Small 24B & 95.13 $\pm$ 0.52 & 82.22 $\pm$ 0.92 & \textbf{74.59 $\pm$ 1.13} & 62.79 $\pm$ 1.45 \\
Mistral-Large 123B & \textbf{97.25 $\pm$ 0.33} & \textbf{86.53 $\pm$ 0.83} & \underline{74.14 $\pm$ 1.11} & 72.87 $\pm$ 1.58 \\
Qwen3 32B & 96.42 $\pm$ 0.41 & 82.43 $\pm$ 0.88 & 73.91 $\pm$ 1.12 & 62.45 $\pm$ 1.14 \\
Llama-3.3 70B & 92.74 $\pm$ 0.61 & 83.55 $\pm$ 0.88 & 67.41 $\pm$ 1.18 & 72.43 $\pm$ 0.90 \\
Qwen3 8B & 73.38 $\pm$ 2.25 & 67.17 $\pm$ 2.10 & 57.94 $\pm$ 2.03 & 63.77 $\pm$ 2.08 \\
Llama-3.1 8B & 71.90 $\pm$ 2.12 & 65.59 $\pm$ 1.78 & 56.13 $\pm$ 1.71 & 64.40 $\pm$ 1.95 \\
\bottomrule
\end{tabular}
\end{adjustbox}
\end{table}

\begin{table}[H]
  \setlength{\belowcaptionskip}{4pt}
\centering
\caption{Stage~2 story-macro accuracy and 95\% story-bootstrap intervals, in percent. Part 2 reports the remaining dimensions and the overall score.}
\label{tab:stage2-dimension-intervals-cont}
\begin{adjustbox}{max width=\textwidth,center}
\begin{tabular}{@{}lrrrr@{}}
\toprule
\textbf{Model} & \textbf{Content Type} & \textbf{Mental Source} & \textbf{Context} & \textbf{Overall} \\
\midrule
Gemini-2.5 Pro & \textbf{86.35 $\pm$ 0.78} & \textbf{86.70 $\pm$ 0.63} & 87.07 $\pm$ 0.91 & \underline{85.54 $\pm$ 0.29} \\
Gemini-2.5 Flash & \underline{85.97 $\pm$ 0.87} & 84.10 $\pm$ 0.83 & \underline{92.14 $\pm$ 0.79} & \textbf{85.95 $\pm$ 0.50} \\
GPT-5 & 79.96 $\pm$ 1.25 & 83.02 $\pm$ 1.05 & 88.83 $\pm$ 1.15 & 82.85 $\pm$ 0.90 \\
Gemma-3 27B & 73.50 $\pm$ 1.04 & 78.72 $\pm$ 0.84 & 92.07 $\pm$ 0.73 & 78.46 $\pm$ 0.44 \\
Mistral-Small 24B & 76.01 $\pm$ 1.01 & 84.82 $\pm$ 0.77 & 91.90 $\pm$ 0.77 & 81.06 $\pm$ 0.38 \\
Mistral-Large 123B & 82.83 $\pm$ 0.86 & \underline{86.32 $\pm$ 0.74} & \textbf{92.97 $\pm$ 0.68} & 84.70 $\pm$ 0.38 \\
Qwen3 32B & 71.27 $\pm$ 1.16 & 76.84 $\pm$ 1.04 & 90.81 $\pm$ 0.77 & 79.16 $\pm$ 0.44 \\
Llama-3.3 70B & 72.35 $\pm$ 1.04 & 76.71 $\pm$ 0.88 & 91.69 $\pm$ 0.81 & 79.55 $\pm$ 0.44 \\
Qwen3 8B & 51.43 $\pm$ 1.87 & 61.49 $\pm$ 2.03 & 74.62 $\pm$ 2.30 & 64.26 $\pm$ 1.91 \\
Llama-3.1 8B & 48.63 $\pm$ 1.68 & 55.18 $\pm$ 1.82 & 76.81 $\pm$ 1.96 & 62.66 $\pm$ 1.60 \\
\bottomrule
\end{tabular}
\end{adjustbox}
\end{table}

\begin{table}[H]
  \setlength{\belowcaptionskip}{4pt}
\centering
\caption{Paired Stage~2 comparisons for adjacent models with overall point estimates within two percentage points. Difference is higher-point-score model minus the next model in the ranking.}
\label{tab:stage2-close-pairs}
\begin{adjustbox}{max width=\textwidth,center}
\begin{tabular}{@{}lrrr@{}}
\toprule
\textbf{Comparison} & \textbf{Difference} & \textbf{Paired 95\% CI} & \textbf{Win share} \\
\midrule
Gemini-2.5 Flash minus Gemini-2.5 Pro & +0.41 & $\pm$ 0.51 & 93.75\% \\
Gemini-2.5 Pro minus Mistral-Large 123B & +0.84 & $\pm$ 0.45 & 99.99\% \\
Mistral-Large 123B minus GPT-5 & +1.85 & $\pm$ 0.94 & 100.00\% \\
GPT-5 minus Mistral-Small 24B & +1.79 & $\pm$ 0.95 & 99.97\% \\
Mistral-Small 24B minus Llama-3.3 70B & +1.51 & $\pm$ 0.44 & 100.00\% \\
Llama-3.3 70B minus Qwen3 32B & +0.39 & $\pm$ 0.47 & 94.81\% \\
Qwen3 32B minus Gemma-3 27B & +0.71 & $\pm$ 0.51 & 99.73\% \\
Qwen3 8B minus Llama-3.1 8B & +1.60 & $\pm$ 2.42 & 89.87\% \\
\bottomrule
\end{tabular}
\end{adjustbox}
\end{table}

Most neighboring model differences are supported, but the intervals for Gemini-2.5 Flash versus Gemini-2.5 Pro, Llama-3.3 70B versus Qwen3 32B, and Qwen3 8B versus Llama-3.1 8B include zero. Order is generally the strongest dimension, while Knowledge Access and Representation are more challenging. Table~\ref{tab:stage2-imbalance-summary} compares the best model accuracy with majority and prior-random baselines, and also reports macro-\(F_1\) and balanced accuracy.

\begin{table}[H]
  \setlength{\belowcaptionskip}{4pt}
\centering
\caption{Stage~2 imbalance-aware summary (\%). Best model scores are shown for each dimension. Majority and prior-random baselines use the empirical label distribution.}
\label{tab:stage2-imbalance-summary}
\begin{adjustbox}{max width=\textwidth,center}
\begin{tabular}{@{}lrrrrr@{}}
\toprule
\textbf{Dimension} & \textbf{Majority} & \shortstack{\textbf{Prior-}\\\textbf{random}} & \shortstack{\textbf{Best}\\\textbf{accuracy}} & \shortstack{\textbf{Best macro-}\\\textbf{\(F_1\)}} & \shortstack{\textbf{Best balanced}\\\textbf{accuracy}} \\
\midrule
Order & 55.81 & 43.82 & 97.25 & 95.05 & 97.24 \\
Truth Status & 78.65 & 65.45 & 86.53 & 74.62 & 75.76 \\
Knowledge Access & 50.00 & 45.16 & 74.59 & 59.46 & 70.32 \\
Representation & 55.73 & 50.80 & 89.01 & 88.62 & 89.32 \\
Content Type & 34.68 & 20.55 & 86.35 & 83.49 & 84.62 \\
Mental Source & 33.60 & 24.90 & 86.70 & 70.57 & 80.52 \\
Context & 91.54 & 84.37 & 92.97 & 60.91 & 78.96 \\
\bottomrule
\end{tabular}
\end{adjustbox}
\end{table}

Most dimensions remain above the frequency-based baselines. Knowledge Access remains difficult under balanced measures, while Context's high raw accuracy is close to its majority-label baseline, reflecting the prevalence of Neutral labels.

\FloatBarrier

\subsubsection{Stage~2 accuracy by gold label}
\label{app:stage2-dim-category-results}\label{app:stage2-label-breakdown}

The tables break down Stage~2 accuracy by gold label, showing where models succeed or struggle within each dimension.

\begin{table}[!htbp]
  \setlength{\belowcaptionskip}{4pt}
\centering
\caption{Stage~2 story-macro accuracy by gold Order label (\%). Belief counts by label, in column order: 7,276; 12,763; 2,200; 104.}
\label{tab:stage2-label-order}
\begin{adjustbox}{max width=\textwidth,center}
\begin{tabular}{@{}lrrrr@{}}
\toprule
\textbf{Model} & \multicolumn{1}{r}{\textbf{0}} & \multicolumn{1}{r}{\textbf{1}} & \multicolumn{1}{r}{\textbf{2}} & \multicolumn{1}{r}{\textbf{3} } \\
\midrule
Gemini-2.5 Pro & \textbf{100.00} & 94.44 & \textbf{95.87} & \textbf{100.00} \\
Gemini-2.5 Flash & 99.57 & 93.68 & \underline{94.52} & \textbf{100.00} \\
GPT-5 & 97.97 & 94.57 & 90.18 & \textbf{100.00} \\
Gemma-3 27B & \underline{99.90} & \textbf{98.35} & 72.42 & \underline{98.11} \\
Mistral-Small 24B & 99.55 & 93.38 & 89.44 & \underline{98.11} \\
Mistral-Large 123B & 99.67 & \underline{98.20} & 80.70 & \underline{98.11} \\
Qwen3 32B & 99.41 & 97.84 & 76.11 & \underline{98.11} \\
Llama-3.3 70B & 99.50 & 90.30 & 83.59 & \underline{98.11} \\
Qwen3 8B & 78.94 & 74.70 & 49.59 & 43.40 \\
Llama-3.1 8B & 73.86 & 71.58 & 60.68 & 68.87 \\
\bottomrule
\end{tabular}
\end{adjustbox}
\end{table}

\begin{table}[!htbp]
  \setlength{\belowcaptionskip}{4pt}
\centering
\caption{Stage~2 story-macro accuracy by gold Truth Status label (\%). Belief counts by label, in column order: 788; 17,697; 3,858.}
\label{tab:stage2-label-truth-status}
\begin{adjustbox}{max width=\textwidth,center}
\begin{tabular}{@{}lrrr@{}}
\toprule
\textbf{Model} & \multicolumn{1}{r}{\textbf{False}} & \multicolumn{1}{r}{\textbf{True}} & \multicolumn{1}{r}{\textbf{Unknown} } \\
\midrule
Gemini-2.5 Pro & \textbf{85.74} & 93.92 & 58.71 \\
Gemini-2.5 Flash & \underline{79.39} & \underline{94.73} & 45.83 \\
GPT-5 & 61.11 & 85.31 & \textbf{78.32} \\
Gemma-3 27B & 59.15 & 89.22 & \underline{59.86} \\
Mistral-Small 24B & 46.33 & \textbf{97.25} & 22.73 \\
Mistral-Large 123B & 61.27 & 93.81 & 57.87 \\
Qwen3 32B & 52.75 & 90.58 & 53.15 \\
Llama-3.3 70B & 65.06 & 93.07 & 45.19 \\
Qwen3 8B & 29.35 & 77.44 & 30.43 \\
Llama-3.1 8B & 25.02 & 77.25 & 24.39 \\
\bottomrule
\end{tabular}
\end{adjustbox}
\end{table}

\begin{table}[!htbp]
  \setlength{\belowcaptionskip}{4pt}
\centering
\caption{Stage~2 story-macro accuracy by gold Knowledge Access label (\%). Belief counts by label, in column order: 11,046; 10,055; 1,242.}
\label{tab:stage2-label-knowledge-access}
\begin{adjustbox}{max width=\textwidth,center}
\begin{tabular}{@{}lrrr@{}}
\toprule
\textbf{Model} & \multicolumn{1}{r}{\textbf{Private}} & \multicolumn{1}{r}{\textbf{Public}} & \multicolumn{1}{r}{\textbf{Shared} } \\
\midrule
Gemini-2.5 Pro & 68.35 & 66.17 & \textbf{78.23} \\
Gemini-2.5 Flash & 70.58 & \textbf{79.16} & 51.55 \\
GPT-5 & 67.28 & 71.93 & 47.22 \\
Gemma-3 27B & 80.94 & \underline{73.90} & 38.03 \\
Mistral-Small 24B & \textbf{91.06} & 72.28 & 9.94 \\
Mistral-Large 123B & \underline{89.82} & 72.61 & 14.57 \\
Qwen3 32B & 89.39 & 72.79 & 7.74 \\
Llama-3.3 70B & 68.64 & 73.73 & \underline{52.61} \\
Qwen3 8B & 62.82 & 62.52 & 14.11 \\
Llama-3.1 8B & 58.94 & 63.30 & 7.51 \\
\bottomrule
\end{tabular}
\end{adjustbox}
\end{table}

\begin{table}[!htbp]
  \setlength{\belowcaptionskip}{4pt}
\centering
\caption{Stage~2 story-macro accuracy by gold Representation label (\%). Belief counts by label, in column order: 9,611; 12,732.}
\label{tab:stage2-label-representation}
\begin{adjustbox}{max width=\textwidth,center}
\begin{tabular}{@{}lrr@{}}
\toprule
\textbf{Model} & \multicolumn{1}{r}{\textbf{Explicit}} & \multicolumn{1}{r}{\textbf{Implicit} } \\
\midrule
Gemini-2.5 Pro & 95.23 & \textbf{84.16} \\
Gemini-2.5 Flash & 93.72 & \underline{83.32} \\
GPT-5 & 91.06 & 79.29 \\
Gemma-3 27B & 73.48 & 41.96 \\
Mistral-Small 24B & 95.46 & 39.57 \\
Mistral-Large 123B & 89.35 & 62.38 \\
Qwen3 32B & \textbf{97.82} & 34.85 \\
Llama-3.3 70B & \underline{96.81} & 54.16 \\
Qwen3 8B & 76.06 & 55.83 \\
Llama-3.1 8B & 62.13 & 68.04 \\
\bottomrule
\end{tabular}
\end{adjustbox}
\end{table}

\begin{table}[!htbp]
  \setlength{\belowcaptionskip}{4pt}
\centering
\setlength{\tabcolsep}{2.5pt}
\caption{Stage~2 story-macro accuracy by gold Content Type label (\%). Belief counts by label, in column order: 8,457; 3,211; 2,484; 1,048; 3,573; 1,144; 1,270; 1,156.}
\label{tab:stage2-label-content-type}
\begin{tabularx}{\textwidth}{@{}l*{8}{>{\raggedleft\arraybackslash}X}@{}}
\toprule
\textbf{Model} & \multicolumn{1}{r}{{\footnotesize\shortstack{\textbf{Action/}\\\textbf{Event}}}} & \multicolumn{1}{r}{{\footnotesize\shortstack{\textbf{Contents/}\\\textbf{Physical}\\\textbf{State}}}} & \multicolumn{1}{r}{{\footnotesize\shortstack{\textbf{Desire/}\\\textbf{Intention}}}} & \multicolumn{1}{r}{\textbf{Emotion}} & \multicolumn{1}{r}{{\footnotesize\textbf{Epistemic}}} & \multicolumn{1}{r}{{\footnotesize\shortstack{\textbf{Identity}\\\textbf{/ Relation}}}} & \multicolumn{1}{r}{{\footnotesize\textbf{Location}}} & \multicolumn{1}{r}{{\footnotesize\shortstack{\textbf{Trait/}\\\textbf{Value}}} } \\
\midrule
Gemini-2.5 Pro & \textbf{95.30} & 60.63 & \underline{79.22} & \underline{70.18} & 89.46 & 89.30 & 87.77 & \textbf{92.77} \\
Gemini-2.5 Flash & \underline{91.30} & \underline{83.69} & \textbf{81.14} & 65.61 & 89.55 & \textbf{91.88} & 86.10 & 78.44 \\
GPT-5 & 90.24 & 49.26 & 68.61 & \textbf{75.18} & 86.25 & \underline{89.87} & 85.60 & \underline{88.72} \\
Gemma-3 27B & 73.86 & 71.23 & 75.79 & 63.35 & 84.65 & 82.89 & \underline{87.87} & 60.01 \\
Mistral-Small 24B & 80.70 & \textbf{84.77} & 55.59 & 56.66 & \textbf{93.14} & 77.87 & 82.42 & 35.69 \\
Mistral-Large 123B & 89.80 & 82.69 & 78.69 & 69.28 & 91.96 & 80.20 & 81.48 & 32.64 \\
Qwen3 32B & 83.11 & 64.80 & 53.42 & 39.17 & \underline{92.84} & 85.63 & 83.21 & 19.83 \\
Llama-3.3 70B & 73.15 & 73.57 & 74.23 & 54.53 & 85.29 & 88.51 & \textbf{88.43} & 13.66 \\
Qwen3 8B & 57.04 & 44.25 & 56.12 & 28.61 & 61.16 & 72.48 & 68.47 & 14.37 \\
Llama-3.1 8B & 53.01 & 45.59 & 33.94 & 15.31 & 75.77 & 47.25 & 64.33 & 8.65 \\
\bottomrule
\end{tabularx}
\end{table}

\begin{table}[!htbp]
  \setlength{\belowcaptionskip}{4pt}
\centering
\setlength{\tabcolsep}{2.5pt}
\caption{Stage~2 story-macro accuracy by gold Mental Source label (\%). Belief counts by label, in column order: 14; 6,555; 1,750; 7,277; 4,510; 1,859; 378.}
\label{tab:stage2-label-mental-source}
\begin{tabularx}{\textwidth}{@{}l*{7}{>{\raggedleft\arraybackslash}X}@{}}
\toprule
\textbf{Model} & \multicolumn{1}{r}{{\footnotesize\textbf{Imagination}}} & \multicolumn{1}{r}{\textbf{Inference}} & \multicolumn{1}{r}{\textbf{Memory}} & \multicolumn{1}{r}{\textbf{Narration}} & \multicolumn{1}{r}{{\footnotesize\textbf{Perception}}} & \multicolumn{1}{r}{{\footnotesize\textbf{Testimony}}} & \multicolumn{1}{r}{\textbf{Unknown} } \\
\midrule
Gemini-2.5 Pro & \textbf{100.00} & 90.05 & \textbf{74.21} & \textbf{99.98} & 68.45 & \textbf{88.90} & 27.83 \\
Gemini-2.5 Flash & \textbf{100.00} & 86.05 & 61.19 & 99.50 & 71.27 & 66.89 & \textbf{78.77} \\
GPT-5 & \underline{40.00} & 83.22 & 66.84 & 97.73 & 69.27 & \underline{79.22} & \underline{34.67} \\
Gemma-3 27B & 20.00 & 78.89 & 41.61 & \underline{99.97} & \textbf{78.41} & 48.86 & 2.83 \\
Mistral-Small 24B & \underline{40.00} & \underline{91.98} & \underline{66.85} & 98.36 & \underline{78.05} & 58.46 & 2.83 \\
Mistral-Large 123B & \underline{40.00} & \textbf{93.28} & 63.00 & 99.90 & 75.40 & 64.23 & 1.89 \\
Qwen3 32B & 20.00 & 80.03 & 31.71 & 98.88 & 70.98 & 41.98 & 12.03 \\
Llama-3.3 70B & \underline{40.00} & 75.03 & 51.09 & 95.82 & 70.20 & 65.81 & 1.89 \\
Qwen3 8B & 0.00 & 63.61 & 18.50 & 79.39 & 59.63 & 20.98 & 2.59 \\
Llama-3.1 8B & 20.00 & 64.57 & 23.65 & 72.12 & 45.62 & 12.50 & 0.00 \\
\bottomrule
\end{tabularx}
\end{table}

\begin{table}[!htbp]
  \setlength{\belowcaptionskip}{4pt}
\centering
\caption{Stage~2 story-macro accuracy by gold Context label (\%). Belief counts by label, in column order: 74; 1,034; 20,514; 721.}
\label{tab:stage2-label-context}
\begin{adjustbox}{max width=\textwidth,center}
\begin{tabular}{@{}lrrrr@{}}
\toprule
\textbf{Model} & \multicolumn{1}{r}{\textbf{Counterfactual}} & \multicolumn{1}{r}{\textbf{Deceptive}} & \multicolumn{1}{r}{\textbf{Neutral}} & \multicolumn{1}{r}{\textbf{Temporal} } \\
\midrule
Gemini-2.5 Pro & \textbf{37.10} & \textbf{96.19} & 86.55 & \textbf{71.48} \\
Gemini-2.5 Flash & \underline{13.33} & 68.30 & 95.41 & 39.25 \\
GPT-5 & 12.29 & \underline{74.72} & 89.88 & \underline{65.29} \\
Gemma-3 27B & 0.00 & 13.69 & 98.43 & 29.15 \\
Mistral-Small 24B & 7.14 & 15.99 & \textbf{99.54} & 0.50 \\
Mistral-Large 123B & 7.14 & 32.47 & 98.83 & 17.55 \\
Qwen3 32B & 10.48 & 35.67 & 96.26 & 22.14 \\
Llama-3.3 70B & 7.14 & 16.95 & \underline{99.12} & 0.00 \\
Qwen3 8B & 7.14 & 10.84 & 80.67 & 4.95 \\
Llama-3.1 8B & 0.00 & 2.21 & 83.11 & 10.41 \\
\bottomrule
\end{tabular}
\end{adjustbox}
\end{table}

Models are often less accurate on Shared Knowledge Access and Implicit Representation than on other labels within those dimensions. Context accuracy is high for Neutral labels, which are much more common, but lower for Deceptive, Temporal, and Counterfactual labels.

\FloatBarrier

\FloatBarrier

\subsection{Endpoint Question Answering versus OmniToM Stage~1}
\label{app:endpoint-qa-comparison}

We compare Stage~1 belief-extraction \(F_1\) with story-macro accuracy on the English ToMBench endpoint questions linked to the same stories. Following ToMBench, we asked each question with five randomized answer-option orders and scored it by majority vote.

\begin{table}[!htbp]
  \setlength{\belowcaptionskip}{4pt}
\centering
\caption{Endpoint QA accuracy and OmniToM Stage~1 on linked ToMBench stories. Both scores include 95\% story-bootstrap confidence intervals. Ranks are computed separately within each evaluation.}
\label{tab:endpoint-stage1-overall}
\setlength{\tabcolsep}{1.5pt}
\begin{adjustbox}{max width=\textwidth,center}
\begin{tabular}{@{}lrrrr@{}}
\toprule
\textbf{Model} & \textbf{Endpoint accuracy} & \textbf{Endpoint rank} & \textbf{Stage~1 \(F_1\)} & \textbf{Stage~1 rank} \\
\midrule
Gemini-2.5 Pro & \textbf{80.67 $\pm$ 2.15} & 1 & \textbf{61.63 $\pm$ 0.76} & 1 \\
Gemma-3 27B & 53.95 $\pm$ 2.53 & 9 & \underline{58.03 $\pm$ 0.97} & 2 \\
Mistral-Small 24B & 71.08 $\pm$ 2.45 & 5 & 56.46 $\pm$ 0.95 & 3 \\
Gemini-2.5 Flash & \underline{77.30 $\pm$ 2.32} & 2 & 55.75 $\pm$ 1.10 & 4 \\
Mistral-Large 123B & 73.50 $\pm$ 2.42 & 3 & 53.53 $\pm$ 0.91 & 5 \\
Qwen3 32B & 69.96 $\pm$ 2.44 & 6 & 51.63 $\pm$ 0.94 & 6 \\
Llama-3.3 70B & 71.28 $\pm$ 2.45 & 4 & 47.12 $\pm$ 0.98 & 7 \\
Qwen3 8B & 63.31 $\pm$ 2.46 & 7 & 42.70 $\pm$ 1.18 & 8 \\
Llama-3.1 8B & 60.79 $\pm$ 2.58 & 8 & 37.21 $\pm$ 1.18 & 9 \\
\bottomrule
\end{tabular}
\end{adjustbox}
\end{table}

\FloatBarrier

The rankings differ across the two evaluations. Gemma-3 27B ranks second on Stage~1 extraction but ninth on endpoint QA. Mistral-Small 24B shows a different pattern, ranking fifth on endpoint QA despite having lower Stage~1 \(F_1\) than Gemma (56.46 versus 58.03). The qualitative examples in App.~\ref{app:qualitative-comparison} likewise show that correct endpoint answers can coexist with errors elsewhere in a model's belief representation.

Table~\ref{tab:endpoint-qa-by-family} breaks endpoint accuracy down by ToMBench story family.

\begin{table}[!htbp]
  \setlength{\belowcaptionskip}{4pt}
\centering
\caption{Endpoint QA accuracy by ToMBench story family (\%). Within each family, accuracy is first averaged over a story's questions and then equally across stories. Columns correspond, from left to right, to Ambiguous Story Task, False Belief Task, Faux-pas Recognition Test, Hinting Task Test, Persuasion Story Task, Scalar Implicature Test, and Strange Story Task.}
\label{tab:endpoint-qa-by-family}
\begin{tabular*}{\textwidth}{@{\extracolsep{\fill}}lrrrrrrr@{}}
\toprule
\textbf{Model} & \textbf{AST} & \textbf{FBT} & \textbf{FPT} & \textbf{HT} & \textbf{PST} & \textbf{SIT} & \textbf{SST} \\
\midrule
Gemini-2.5 Pro & \textbf{84.18} & \underline{91.41} & \underline{77.00} & \textbf{99.00} & \underline{64.95} & \textbf{62.66} & \textbf{88.41} \\
Gemini-2.5 Flash & 81.63 & \textbf{91.75} & \textbf{79.93} & \underline{89.00} & \textbf{65.98} & 50.32 & \underline{86.39} \\
Gemma-3 27B & 58.67 & 59.79 & 66.67 & 51.00 & 45.36 & 35.06 & 59.74 \\
Mistral-Small 24B & 71.94 & 89.69 & 74.30 & 82.00 & 60.82 & 42.21 & 80.76 \\
Mistral-Large 123B & \underline{82.14} & 85.22 & 75.35 & 84.00 & 63.92 & 46.10 & 82.45 \\
Qwen3 32B & 73.98 & 83.33 & 70.19 & 83.00 & 62.89 & 43.18 & 78.58 \\
Llama-3.3 70B & 75.51 & 88.66 & 73.88 & 77.00 & 60.82 & 47.40 & 79.23 \\
Qwen3 8B & 66.33 & 63.92 & 68.43 & 64.00 & 57.73 & \underline{52.92} & 68.12 \\
Llama-3.1 8B & 66.33 & 65.12 & 67.43 & 59.00 & 54.64 & 41.88 & 69.40 \\
\bottomrule
\end{tabular*}
\end{table}

Gemma-3 27B performs better on False Belief than on Scalar Implicature, while the Gemini models are among the strongest on both. This difference across families helps explain why the model rankings vary between evaluations.

\FloatBarrier

\subsection{Reference Model Sensitivity}
\label{app:evaluation-sensitivity}
\label{app:reference-sensitivity}

The primary Stage~1 benchmark uses belief references generated by Claude-Sonnet-4.5, which was selected using the human-validated subset. To test sensitivity to this choice, we score the same predictions against references generated by Mistral-Large-Instruct-2407 while keeping the GPT-5 judge fixed. The tables show the resulting scores, rankings, and paired score changes.

\begin{table}[H]
  \setlength{\belowcaptionskip}{4pt}
\centering
\caption{Stage~1 sensitivity to the benchmark reference. Both estimates use the fixed GPT-5 semantic judge and story-macro \(F_1\), with 95\% story-bootstrap intervals. Ranks are based on point estimates among the seven models shown, with higher scores ranked first.}
\label{tab:reference-sensitivity-score}
\begin{adjustbox}{max width=\textwidth,center}
\begin{tabular}{@{}lrrrr@{}}
\toprule
\textbf{Model} & \shortstack{\textbf{Claude reference}\\\textbf{(95\% CI)}} & \textbf{Rank} & \shortstack{\textbf{Mistral reference}\\\textbf{(95\% CI)}} & \textbf{Rank} \\
\midrule
Gemini-2.5 Pro & \textbf{61.63 $\pm$ 0.76} & 1 & \underline{63.20 $\pm$ 0.83} & 2 \\
Gemma-3 27B & \underline{58.03 $\pm$ 0.97} & 2 & \textbf{63.62 $\pm$ 0.91} & 1 \\
Gemini-2.5 Flash & 55.75 $\pm$ 1.10 & 3 & 58.01 $\pm$ 0.90 & 4 \\
Qwen3 32B & 51.63 $\pm$ 0.94 & 4 & 58.75 $\pm$ 0.98 & 3 \\
Llama-3.3 70B & 47.12 $\pm$ 0.96 & 5 & 52.60 $\pm$ 1.03 & 5 \\
Qwen3 8B & 42.70 $\pm$ 1.18 & 6 & 49.25 $\pm$ 1.28 & 6 \\
Llama-3.1 8B & 37.21 $\pm$ 1.18 & 7 & 40.76 $\pm$ 1.17 & 7 \\
\bottomrule
\end{tabular}
\end{adjustbox}
\end{table}

\begin{table}[H]
  \setlength{\belowcaptionskip}{4pt}
\centering
\caption{Paired change in Stage~1 story-macro \(F_1\) when replacing the Claude reference with the Mistral reference. The interval resamples stories as pairs.}
\label{tab:reference-sensitivity-change}
\begin{adjustbox}{max width=\textwidth,center}
\begin{tabular}{@{}lrrr@{}}
\toprule
\textbf{Model} & \textbf{Mistral minus Claude} & \textbf{Paired 95\% CI} & \textbf{Win share} \\
\midrule
Gemini-2.5 Pro & +1.56 & $\pm$ 0.76 & 100.00\% \\
Gemini-2.5 Flash & +2.26 & $\pm$ 1.03 & 100.00\% \\
Gemma-3 27B & +5.59 & $\pm$ 0.81 & 100.00\% \\
Qwen3 32B & +7.12 & $\pm$ 0.77 & 100.00\% \\
Llama-3.3 70B & +5.48 & $\pm$ 0.78 & 100.00\% \\
Qwen3 8B & +6.55 & $\pm$ 0.72 & 100.00\% \\
Llama-3.1 8B & +3.55 & $\pm$ 0.71 & 100.00\% \\
\bottomrule
\end{tabular}
\end{adjustbox}
\end{table}

Using the Mistral reference raises \(F_1\) for all seven models by 1.56--7.12 points. Three models retain their ranks, while Gemma-3 27B and Gemini-2.5 Pro, and Qwen3 32B and Gemini-2.5 Flash, switch places. The broad ordering remains similar, but absolute scores and some close rankings depend on the reference. Mistral models are excluded because Mistral-Large generated the alternative reference.

\FloatBarrier

\subsection{Semantic Judge Sensitivity}
\label{app:judge-sensitivity}

Stage~1 uses GPT-5 to align predicted beliefs with the references. To test sensitivity to the judge, we re-score the same predictions and Claude references with Llama-3.3-70B-Instruct, excluding Llama source models from the comparison. In human validation, GPT-5 agrees with human decisions more often than Llama-3.3 70B (87.76\% versus 75.85\%; App.~\ref{app:construction-validation}). The table shows the scores and rankings under each judge.

\begin{table}[H]
  \setlength{\belowcaptionskip}{4pt}
\centering
\caption{Stage~1 sensitivity to the semantic judge. Confidence intervals use 10,000 paired story-bootstrap resamples for Llama minus GPT-5. Ranks are based on point estimates among the five source models shown, with higher scores ranked first.}
\label{tab:judge-sensitivity}
\begin{tabular*}{\textwidth}{@{\extracolsep{\fill}}lrrrrrr@{}}
\toprule
\textbf{Source model} & \textbf{GPT-5 \(F_1\)} & \textbf{Rank} & \textbf{Llama \(F_1\)} & \textbf{Rank} & \textbf{\(\Delta F_1\)} & \textbf{95\% paired CI} \\
\midrule
Gemma-3 27B & \textbf{58.03} & 1 & \textbf{74.82} & 1 & +16.79 & $\pm$ 0.92 \\
Mistral-Small 24B & \underline{56.46} & 2 & 70.93 & 3 & +14.47 & $\pm$ 1.22 \\
Mistral-Large 123B & 53.53 & 3 & \underline{72.25} & 2 & +18.72 & $\pm$ 1.19 \\
Qwen3 32B & 51.63 & 4 & 70.41 & 4 & +18.78 & $\pm$ 1.24 \\
Qwen3 8B & 42.70 & 5 & 57.00 & 5 & +14.30 & $\pm$ 1.58 \\
\bottomrule
\end{tabular*}
\end{table}

The Llama judge raises Stage~1 \(F_1\) by 14.30--18.78 points. Three models retain their ranks, while Mistral-Small and Mistral-Large switch places. The broad ordering remains similar, but absolute scores and one close ranking depend on the judge. GPT-5 remains the primary judge because it agrees more often with human decisions.

\FloatBarrier

\subsection{Output and Coverage Audits}
\label{app:output-coverage-audits}

\subsubsection{Stage~1 Extraction Volume Audit}
\label{app:stage1-volume-audit}

\begin{figure}[!htbp]
  \centering
  \includegraphics[width=0.82\textwidth]{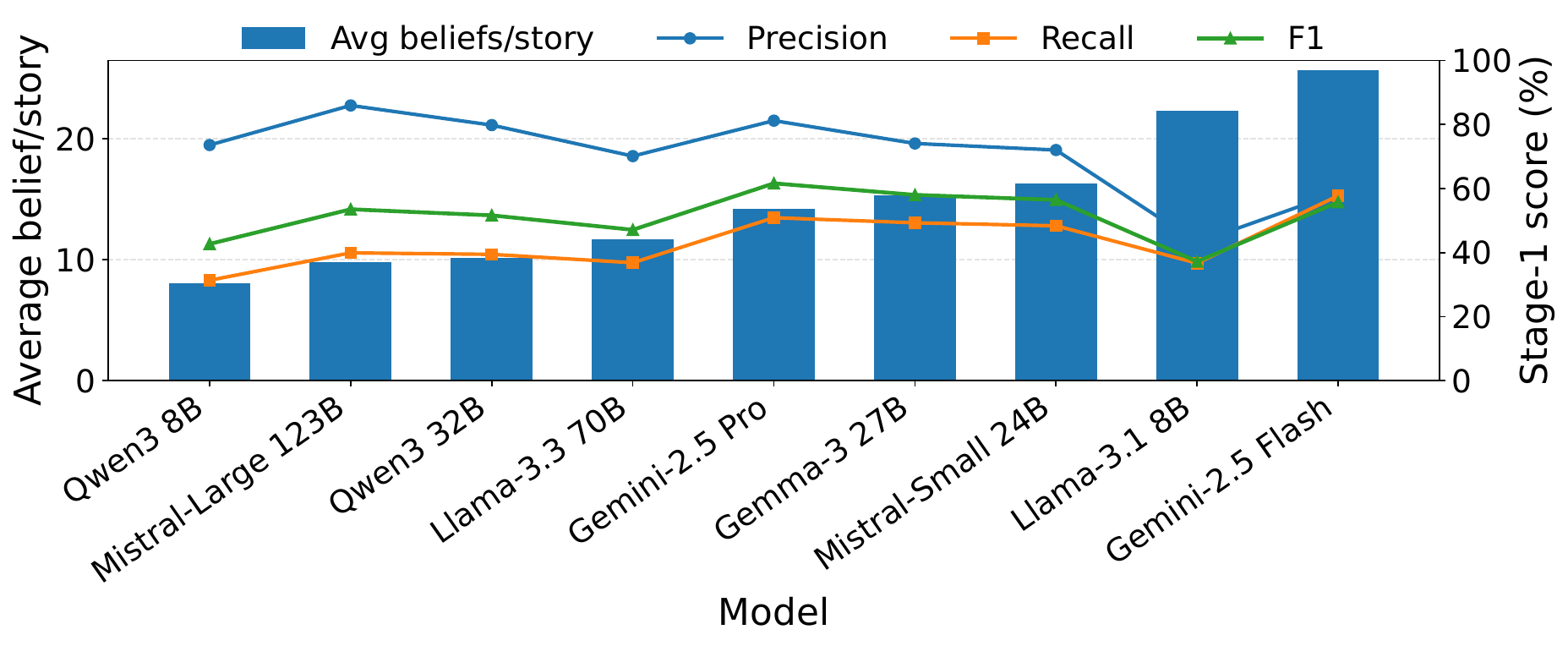}
  \caption{Stage~1 extraction-volume audit. Average generated beliefs per story plotted against Precision, Recall, and \(F_1\).}
  \label{fig:appendix-volume-analysis}
\end{figure}

We examine whether extraction volume helps explain differences in Stage~1 Precision, Recall, and \(F_1\). Figure~\ref{fig:appendix-volume-analysis} shows that models producing more beliefs tend to recover more gold beliefs, but may also add unsupported or redundant propositions. Output volume affects the precision-recall balance, but does not by itself explain actor-specific extraction performance.

\FloatBarrier

\subsubsection{Stage~1 MatchCount Distribution}
\label{app:matchcount-distribution}

Table~\ref{tab:experiment-matchcount-distribution} summarizes how often predicted beliefs receive zero, one, or multiple semantic matches. Most are unmatched or have one match; compound matches are uncommon.

\begin{table}[!htbp]
  \setlength{\belowcaptionskip}{4pt}
\centering
\caption{Predicted-side \texttt{MatchCount} distribution in the full Stage~1 benchmark runs. Counts are computed from model prediction tables evaluated by the fixed GPT-5 semantic judge.}
\label{tab:experiment-matchcount-distribution}
\begin{tabularx}{\textwidth}{@{}l*{6}{>{\raggedleft\arraybackslash}X}@{}}
\toprule
\textbf{Model} & \textbf{MC=0} & \textbf{MC=1} & \textbf{MC=2} & \textbf{MC=3} & \textbf{MC=4} & \textbf{Total} \\
\midrule
Gemini-2.5 Pro & 2{,}521 & 9{,}490 & 583 & 84 & 3 & 12{,}681 \\
Gemini-2.5 Flash & 10{,}934 & 13{,}196 & 104 & 20 & 4 & 24{,}258 \\
Gemma-3 27B & 3{,}486 & 9{,}887 & 306 & 25 & 0 & 13{,}704 \\
Mistral-Small-24B-Instruct-2501 & 4{,}583 & 9{,}574 & 356 & 96 & 4 & 14{,}613 \\
Mistral-Large-Instruct-2407 & 1{,}206 & 6{,}677 & 791 & 108 & 3 & 8{,}785 \\
qwen3-32b & 1{,}930 & 5{,}881 & 1{,}025 & 198 & 15 & 9{,}049 \\
Llama-3.3-70B-Instruct & 3{,}091 & 7{,}043 & 288 & 48 & 1 & 10{,}471 \\
Qwen3-8B & 1{,}660 & 4{,}636 & 638 & 284 & 5 & 7{,}223 \\
Llama-3.1 8B & 13{,}524 & 7{,}429 & 187 & 59 & 2 & 21{,}201 \\
\midrule
\textbf{Total} & 42{,}935 & 73{,}813 & 4{,}278 & 922 & 37 & 121{,}985 \\
\bottomrule
\end{tabularx}
\end{table}

\FloatBarrier

\subsubsection{Unusable-Output Audit}

Table~\ref{tab:output-coverage-audit} summarizes unusable-output rates across models and stages.

\begin{table}[!htbp]
  \setlength{\belowcaptionskip}{4pt}
\centering
\caption{Unusable-output audit across benchmark runs (\%).}
\label{tab:output-coverage-audit}
\begin{adjustbox}{max width=\textwidth,center}
\begin{tabular}{@{}lrr@{}}
\toprule
\textbf{Model} & \textbf{ST1 Unusable} & \textbf{ST2 Unusable} \\
\midrule
Gemini-2.5 Pro & \textbf{0.11\%} & \textbf{0.00\%} \\
Gemini-2.5 Flash & 4.69\% & 0.34\% \\
GPT-5 & -- & 1.79\% \\
Gemma-3 27B & 1.01\% & \textbf{0.00\%} \\
Mistral-Small 24B & 1.01\% & \textbf{0.00\%} \\
Mistral-Large 123B & \underline{0.56\%} & \textbf{0.00\%} \\
Qwen3 32B & 1.34\% & \underline{0.11\%} \\
Llama-3.3 70B & 0.67\% & \textbf{0.00\%} \\
Qwen3 8B & 7.49\% & 14.41\% \\
Llama-3.1 8B & 5.14\% & 3.46\% \\
\bottomrule
\end{tabular}
\end{adjustbox}
\end{table}

Qwen3 8B has the highest unusable-output rates in both stages. Most other models have low rates, and several have complete or near-complete Stage~2 coverage.

\FloatBarrier

\appsection{Schema Reference and Labeling Rules}
\label{app:atoms-coverage}

This appendix explains how OmniToM derives its operational Stage~2 schema from prior task-level ToM ability taxonomies. Each belief proposition receives exactly one label from each of the seven OmniToM dimensions.

\newcommand{\RecordField}[1]{\raggedright\arraybackslash\ttfamily #1}

\subsection{From ATOMS Abilities to OmniToM Belief-Level Dimensions}

ATOMS (\emph{Abilities in Theory of Mind Space}) is a taxonomy of Theory-of-Mind measures derived from a systematic review of ToM tasks and assessment instruments~\citep{beaudoin2020systematic}. The review identifies 220 measures across 830 studies and organizes them into seven broad categories: emotions, desires, intentions, percepts, knowledge, beliefs, and mentalistic understanding of non-literal communication, with 39 finer-grained sub-abilities. We use ATOMS as a coverage scaffold: it identifies the kinds of mental-state abilities a ToM benchmark should cover, but it does not directly define labels for individual belief propositions.

OmniToM adapts this coverage to the belief-proposition level. A task-level ability describes what a story or question tests, such as false belief, hidden emotion, or non-literal communication. A belief-level schema instead describes the representational properties needed inside the story: who holds a belief, whether it is true, who could know it, whether it is explicit or inferred, what it is about, how it was acquired, and what contextual framing affects it.

ToMBench operationalizes this coverage as text-compatible social-reasoning stories with multiple-choice questions, covering eight tasks and 31 abilities~\citep{chen-etal-2024-tombench}. OmniToM uses the retained ToMBench stories as source narratives but shifts evaluation from endpoint answers to explicit belief-structure modeling. Rather than assigning one ability label to an item, OmniToM decomposes each story's representational requirements into labels for relevant world-fact and actor-belief propositions. Fig.~\ref{fig:atoms-to-omnitom-schema} summarizes this derivation. The relationship between task-level abilities and OmniToM dimensions is many-to-many: one ToM ability may require several belief-level properties, and one OmniToM dimension may support multiple abilities.

\begin{center}
    \includegraphics[width=0.98\textwidth]{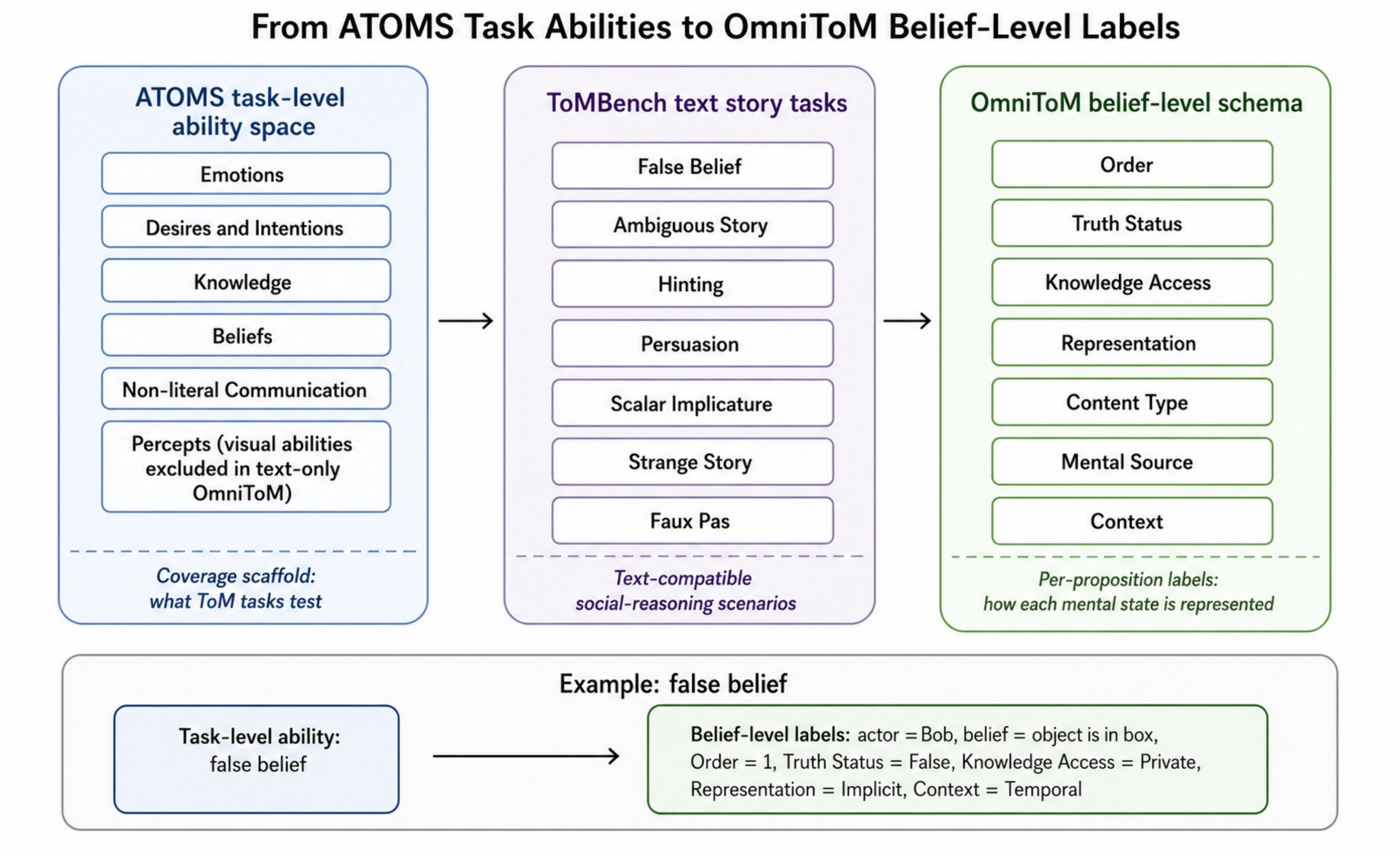}
    \captionof{figure}{ATOMS-to-OmniToM schema derivation. ATOMS provides task-level coverage for ToM abilities, ToMBench instantiates a text-compatible subset as social-reasoning story tasks, and OmniToM decomposes the representational requirements of those tasks into seven belief-level labels assigned to each proposition.}
    \label{fig:atoms-to-omnitom-schema}
\end{center}

False-belief tasks illustrate the mapping. They require tracking that an actor's belief conflicts with narrated reality (\textit{Truth Status}), who had access to the relevant update (\textit{Knowledge Access}), whether the belief is stated or inferred (\textit{Representation}), how the belief was formed (\textit{Mental Source}), and whether it is outdated or deceptive (\textit{Context}). Other task families motivate the remaining coverage: second-order tasks require \textit{Order} and often \textit{Epistemic} content; non-literal communication tasks motivate \textit{Representation} and \textit{Context}; and desire, intention, and emotion tasks motivate the corresponding \textit{Content Type} labels.

Together, these examples clarify the role of ATOMS in OmniToM. The schema does not relabel ATOMS abilities one-to-one; instead, it converts recurring ToM requirements into a unified per-proposition schema for annotating the belief structure of a story. Pure perceptual sub-abilities that require visual stimuli are not directly retained in this text-only benchmark, although perceptual evidence is represented when it functions as a character's \textit{Mental Source}.

\subsection{Schema Dimensions and Labeling Rules}

\paragraph{Order.}
\textit{Label set:} \(\{0,1,2,3\}\). Order captures the recursive depth of the represented belief. Order~0 is reserved for narrator or world-level facts; Order~1 denotes an actor's belief about the world; Order~2 denotes an actor's belief about another actor's belief; and Order~3 denotes a further nested belief attribution. The order cap is based on annotations from the human validation subset: across the retained ToMBench categories, annotators found world facts, first-order beliefs, and occasional second- or third-order beliefs, but no reliably grounded order-4 cases. Putative deeper nestings were treated as underspecified unless directly supported by the narrative.

\paragraph{Truth Status.}
\textit{Label set:} \{\textsc{True}, \textsc{False}, \textsc{Unknown}\}. Truth Status labels whether the proposition is supported, contradicted, or left unresolved by the story evidence. This label is evaluated relative to the narrative, not relative to the believer's confidence. Annotators use \textsc{Unknown} whenever the story does not verify or contradict the proposition; the annotation rule prefers \textsc{Unknown} over speculative completion.

\paragraph{Knowledge Access.}
\textit{Label set:} \{\textsc{Private}, \textsc{Shared}, \textsc{Public}\}. Knowledge Access captures how the proposition is distributed across actors. \textsc{Private} is used for internal or unshared beliefs, even when the underlying event itself is public. \textsc{Shared} is used when the proposition is mutually available within a subset of actors through communication, joint perception, or obvious mutual awareness. \textsc{Public} is reserved for common ground across all relevant actors. This dimension is especially sensitive to whether the proposition is explicitly communicated or only inferred from context.

\paragraph{Representation.}
\textit{Label set:} \{\textsc{Explicit}, \textsc{Implicit}\}. Representation distinguishes directly stated beliefs from beliefs inferred from actions, perception, pragmatic cues, or broader context. \textsc{Explicit} is used when the belief-relevant content is directly narrated, directly spoken, or directly attributed as a mental state. \textsc{Implicit} is used when the annotator must infer the belief from the story's social or pragmatic context.

\paragraph{Content Type.}
\textit{Label set:} \{\textsc{Location}, \textsc{Contents/Physical State}, \textsc{Identity/Relation}, \textsc{Epistemic}, \textsc{Desire/Intention}, \textsc{Emotion}, \textsc{Trait/Value}, \textsc{Action/Event}\}. Content Type identifies what the proposition is about. Location labels where an entity is or was; Contents/Physical State labels what a container holds or the physical condition of an object; Identity/Relation labels actor identities or relationships; Epistemic labels beliefs about knowledge, awareness, or other beliefs; Desire/Intention labels goals, wants, plans, or intended outcomes; Emotion labels affective states; Trait/Value labels preferences, dispositions, or evaluations; and Action/Event labels happenings or actions.

\paragraph{Mental Source.}
\textit{Label set:} \{\textsc{Narration}, \textsc{Perception}, \textsc{Memory}, \textsc{Testimony}, \textsc{Inference}, \textsc{Imagination}, \textsc{Unknown}\}. Mental Source tracks how the belief was acquired. \textsc{Narration} is reserved for world-level facts. \textsc{Perception} is used when an actor directly observes the relevant event or state. \textsc{Memory} is used when the belief depends on a prior perceived state. \textsc{Testimony} is used when the belief comes from another actor's utterance. \textsc{Inference} is used when the belief must be derived from actions, social cues, or context. \textsc{Imagination} covers hypothetical or imagined content, and \textsc{Unknown} is used when the source is not recoverable from the story.

\paragraph{Context.}
\textit{Label set:} \{\textsc{Deceptive}, \textsc{Temporal}, \textsc{Counterfactual}, \textsc{Neutral}\}. Context captures special framing conditions that affect interpretation. \textsc{Temporal} marks stale, recalled, or past-state beliefs, regardless of whether the proposition is ultimately true or false. \textsc{Deceptive} is used for beliefs shaped by lying, concealment, or deliberate misdirection. \textsc{Counterfactual} is used for hypothetical, pretense, or non-actual framing. \textsc{Neutral} is used when none of these special conditions applies.

\appsection{Benchmark Construction and Validation}
\label{app:construction-validation}

This appendix section consolidates the methodological support for benchmark construction: retained source categories, the human-validated annotation pipeline, validation-time model selection, annotation reliability, semantic-judge validation, reasoning-configuration comparisons, and the released benchmark record format.

\subsection{Source Filtering, Retained Categories, and Benchmark Statistics}
\label{app:tombench-category-profiles}
\label{app:benchmark-category-statistics}

OmniToM begins from the 1{,}383-story ToMBench source corpus and retains seven high-volume categories whose stories provide sufficiently self-contained mental-state evidence for explicit belief extraction from story text alone. These categories account for 916 source stories. We exclude categories whose items often require underspecified causal, affective, or world-knowledge completion rather than explicit actor-attributed belief reconstruction. In particular, \textit{Unexpected Outcome} contains many stories, but its items frequently hinge on explaining surprising outcomes from sparse context, making the relevant mental states less consistently recoverable as faithful belief propositions. After holding out 21 human-annotated validation stories, the released benchmark contains 895 stories and 22{,}343 labeled belief propositions. Table~\ref{tab:dataset-category-stats} reports the retained category counts, belief counts, and belief-order distribution.

\begin{table}[!htbp]
  \setlength{\belowcaptionskip}{4pt}
\centering
\caption{OmniToM benchmark category statistics.}
\label{tab:dataset-category-stats}
\begin{tabularx}{\textwidth}{@{}l*{6}{>{\raggedleft\arraybackslash}X}@{}}
\toprule
\textbf{Category} & \textbf{Stories} & \textbf{Beliefs} & \shortstack{\textbf{Beliefs/}\\\textbf{Story}} & \textbf{Order 0} & \textbf{Order 1} & \shortstack{\textbf{Order}\\\textbf{2+}} \\
\midrule
Ambiguous Story Task & 98  & 4,614 & 47.08 & 1,373 & 2,983 & 258 \\
False Belief Task & 97  & 2,168 & 22.35 & 898   & 962   & 308 \\
Faux-pas Recognition Test & 142 & 3,981 & 28.04 & 1,218 & 2,359 & 404 \\
Hinting Task Test & 100 & 2,317 & 23.17 & 620   & 1,382 & 315 \\
Persuasion Story Task & 97  & 1,204 & 12.41 & 483   & 664   & 57  \\
Scalar Implicature Test & 154 & 3,251 & 21.11 & 1,080 & 1,745 & 426 \\
Strange Story Task & 207 & 4,808 & 23.23 & 1,604 & 2,668 & 536 \\
\midrule
Total & 895 & 22,343 & 24.96 & 7,276 & 12,763 & 2,304 \\
\bottomrule
\end{tabularx}
\end{table}

\subsection{Human-Validated Annotation Pipeline}
\label{app:annotation-pipeline-overview}

Benchmark construction must balance semantic rigor with scalable annotation. Exhaustively extracting multi-order belief structures and labeling every belief under the schema introduces substantial cognitive load for human annotators at scale. We therefore adopt a human-validated pipeline that uses human-annotated gold structures to validate the annotation setup and select models, while separating annotation generation from semantic evaluation to reduce self-enhancement bias. Prompt design and the prompt-selection sweep are reported separately in App.~\ref{app:prompt-engineering-teler}.

To support validation and model selection, we reserve a 21-story subset consisting of three stories from each retained category. For Stage~1 extraction, two domain experts independently constructed belief propositions for each story and then reconciled disagreements to produce a human consensus structure. For Stage~2 labeling, we use Mistral-Small 24B~\citep{mistralai2025small24b} only to pre-populate candidate labels under strict schema constraints, reducing annotator burden without delegating final decisions to the model. Three annotators then verified these labels, resolving disagreements via majority vote or expert adjudication. These checks preceded scaling. We did not conduct a separate human audit of the completed benchmark.

\paragraph{Human annotation effort estimate.}
Benchmark development involved a multi-month annotation effort by 11 project annotators, totaling approximately 1.1K person-hours of annotation-related work. This effort covered schema refinement, pilot annotation, belief extraction, schema labeling, semantic-alignment annotation, validation, adjudication, and model selection before the final LLM-assisted pipeline was fixed. The estimate is based on assignment logs and includes annotation work directly supporting the benchmark; it excludes general research time such as paper writing, experiment execution, model analysis, and project meetings. These hours reflect the human work required to define, validate, and adjudicate the annotation process before scaling the finalized pipeline to the 895-story benchmark split.

\FloatBarrier

\subsection{Annotation Model Selection}
\label{app:validation-results}

With the semantic judge fixed, we evaluated candidate annotation models on the 21-story human-annotated validation subset using the same Stage~1 \(F_1\) and Stage~2 accuracy metrics used in the benchmark. The candidate set includes Claude-Sonnet-4.5~\citep{anthropic2025claude}, Mistral-Large 123B~\citep{mistral2024large}, Qwen3 32B~\citep{qwen2025qwen3}, Llama-3.3 70B~\citep{meta2024llama33}, Gemini-2.5 Flash and Gemini-2.5 Pro~\citep{comanici2025gemini25}, Gemma-3 27B~\citep{gemma2025gemma3}, and GPT-5~\citep{openai2025gpt5}.

Table~\ref{tab:validation-21} reports the full comparison. Claude-Sonnet-4.5 obtains the highest Stage~1 \(F_1\), at 72.88, followed by Mistral-Large 123B at 70.13. Claude also obtains the highest Stage~2 accuracy, at 94.42\%, followed by Mistral-Large 123B at 93.76\%. We therefore select Claude-Sonnet-4.5 as the annotation model used to construct the 895-story benchmark split.

\begin{table}[!htbp]
  \setlength{\belowcaptionskip}{4pt}
\centering
\caption{Annotation-model validation on the 21-story human-annotated subset (\%). ST1 reports belief-extraction \(F_1\); ST2 reports belief-labeling accuracy. The two stages are evaluated separately. GPT-5 is not evaluated on Stage~1 because it is the selected semantic judge.}
\label{tab:validation-21}
\begin{adjustbox}{max width=\textwidth,center}
\begin{tabular}{@{}lrr@{}}
\toprule
\textbf{Model} & \textbf{ST1 \(F_1\)} & \textbf{ST2 Acc.} \\
\midrule
Claude-Sonnet-4.5  & \textbf{72.88} & \textbf{94.42} \\
Mistral-Large 123B & \underline{70.13} & \underline{93.76} \\
Qwen3 32B          & 67.82          & 91.14 \\
Llama-3.3 70B      & 67.50          & 93.08 \\
Gemini-2.5 Flash   & 65.33          & 92.97 \\
Gemma-3 27B        & 64.99          & 92.94 \\
Gemini-2.5 Pro     & 64.36          & 91.77 \\
GPT-5              & --             & 92.00 \\
\bottomrule
\end{tabular}
\end{adjustbox}
\end{table}

\paragraph{Stage~2 label-normalization note.}
For Stage~2 labeling evaluation, predicted labels are canonicalized only when they are valid variants of legal schema labels (e.g., trimmed strings, prefixed forms, or known aliases). Predictions that remain noisy or non-label text after normalization are counted as incorrect.

\subsection{Human Annotation Reliability and Semantic-Judge Validation}
\label{app:annotation-reliability}

The reliability analysis covers the human and model validation checks used to support benchmark construction and Stage~1 evaluation. Because the two stages have different output structures, we use agreement measures appropriate to each setting. Stage~1 belief extraction is open-ended, so human extraction agreement is summarized with the existing reviewed-belief overlap measure. Stage~2 assigns categorical labels to fixed belief propositions, allowing us to report both raw agreement and chance-corrected agreement. We use Fleiss' \(\kappa\) for agreement among the three human annotators~\citep{fleiss1971measuring}, and Cohen's \(\kappa\) for pairwise agreement between a model and the adjudicated human reference~\citep{cohen1960coefficient}. We also use Cohen's \(\kappa\) when comparing a semantic judge with binary human alignment decisions.

\paragraph{Stage~1 human extraction.}
For Stage~1 extraction, two domain experts independently constructed belief structures for the human validation stories. Because the beliefs themselves are open-ended propositions rather than labels assigned to a fixed set of items, a categorical chance-corrected statistic such as Cohen's \(\kappa\) is not directly applicable. We therefore retain an overlap-style reconciliation statistic over the reviewed belief set.

Let \(R\) be the set of reviewed Stage~1 belief propositions, and let \(m_i=1\) when belief \(i\) required no reconciliation remark in the \texttt{Belief Remarks} field and \(m_i=0\) otherwise:
\[
\mathrm{Overlap}_{S1}
= \frac{1}{|R|}\sum_{i\in R} m_i .
\]
Of 430 reviewed beliefs, 360 required no reconciliation remark, yielding 83.72\% descriptive overlap. Reconciliation remarks record additions, deletions, and revisions used to construct the final consensus belief structure.

\paragraph{Stage~2 human reliability.}
For Stage~2, three annotators assign categorical labels to the same fixed belief propositions. We report strict all-annotator agreement together with Fleiss' \(\kappa\), which adjusts the observed agreement for agreement expected from the annotators' empirical label frequencies. In general,
\[
\kappa
=
\frac{P_o-P_e}{1-P_e},
\]
where \(P_o\) is the observed agreement and \(P_e\) is the agreement expected under the corresponding label marginals. For the three-annotator setting, \(P_o\) and \(P_e\) are computed using the Fleiss multi-rater formulation.

Table~\ref{tab:stage2-human-reliability} reports both measures for each schema dimension. We additionally report 95\% confidence intervals. Confidence intervals for strict agreement use Wilson intervals, while intervals for Fleiss' \(\kappa\) are estimated by bootstrap resampling over beliefs. The two measures provide complementary information. For example, Context has high raw agreement but a lower chance-corrected \(\kappa\), reflecting the concentration of this dimension in the \textsc{Neutral} label.

\begin{table}[!htbp]
  \setlength{\belowcaptionskip}{4pt}
\centering
\caption{Stage~2 human annotation reliability on the 21-story human-annotated validation subset. Strict agreement requires all three annotators to assign the same label. Values are followed by symmetric margins for 95\% confidence intervals, rounded to the precision shown.}
\label{tab:stage2-human-reliability}
\begin{adjustbox}{max width=\textwidth,center}
\begin{tabular}{@{}lrr@{}}
\toprule
\textbf{Dimension} & \shortstack{\textbf{Strict agreement}\\\textbf{(95\% CI)}} & \shortstack{\textbf{Fleiss' \(\kappa\)}\\\textbf{(95\% CI)}} \\
\midrule
Order & 98.97 $\pm$ 1.11 & 0.9883 $\pm$ 0.0107 \\
Truth Status & 86.67 $\pm$ 3.38 & 0.7178 $\pm$ 0.0722 \\
Knowledge Access & 85.90 $\pm$ 3.46 & 0.8181 $\pm$ 0.0426 \\
Representation & 97.44 $\pm$ 1.63 & 0.9654 $\pm$ 0.0221 \\
Content Type & 88.72 $\pm$ 3.15 & 0.9058 $\pm$ 0.0258 \\
Mental Source & 94.36 $\pm$ 2.32 & 0.9489 $\pm$ 0.0201 \\
Context & 94.62 $\pm$ 2.27 & 0.6334 $\pm$ 0.1325 \\
\bottomrule
\end{tabular}
\end{adjustbox}
\end{table}

Across all seven dimensions, the annotators agree on 2,522 of 2,730 labels, yielding 92.38\% overall strict agreement.

\paragraph{Annotation-model agreement with human Stage~2 labels.}
Because Claude-Sonnet-4.5 is used to scale the final benchmark, we separately measure its agreement with the adjudicated human Stage~2 labels. This analysis is distinct from the model-selection score in Table~\ref{tab:validation-21}: here, the comparison is performed directly between Claude's labels and the adjudicated human labels.

We report exact label accuracy together with Cohen's \(\kappa\). Accuracy measures the fraction of labels that match the human reference directly, while Cohen's \(\kappa\) also accounts for agreement expected from the two label distributions. Confidence intervals are reported for both measures.

As shown in Table~\ref{tab:claude-human-stage2}, agreement varies across the schema dimensions. It is highest for Order and Representation, while Truth Status and Context have lower chance-corrected agreement. Context again shows why the two statistics are useful together: its raw accuracy is 97.17\%, but its Cohen's \(\kappa\) is 0.7718 because the label distribution is highly concentrated.

\begin{table}[!htbp]
  \setlength{\belowcaptionskip}{4pt}
\centering
\caption{Claude-Sonnet-4.5 agreement with adjudicated human Stage~2 labels. Values are followed by symmetric margins for 95\% confidence intervals, rounded to the precision shown.}
\label{tab:claude-human-stage2}
\begin{adjustbox}{max width=\textwidth,center}
\begin{tabular}{@{}lrr@{}}
\toprule
\textbf{Dimension} & \shortstack{\textbf{Accuracy}\\\textbf{(95\% CI)}} & \shortstack{\textbf{Cohen's \(\kappa\)}\\\textbf{(95\% CI)}} \\
\midrule
Order & 98.20 $\pm$ 1.40 & 0.9693 $\pm$ 0.0232 \\
Truth Status & 88.69 $\pm$ 3.16 & 0.6523 $\pm$ 0.0911 \\
Knowledge Access & 89.20 $\pm$ 3.09 & 0.8012 $\pm$ 0.0549 \\
Representation & 95.37 $\pm$ 2.13 & 0.9061 $\pm$ 0.0430 \\
Content Type & 84.58 $\pm$ 3.59 & 0.8126 $\pm$ 0.0425 \\
Mental Source & 91.00 $\pm$ 2.86 & 0.8776 $\pm$ 0.0384 \\
Context & 97.17 $\pm$ 1.70 & 0.7718 $\pm$ 0.1348 \\
\bottomrule
\end{tabular}
\end{adjustbox}
\end{table}

\paragraph{Semantic-Judge Validation.}
Stage~1 benchmark evaluation requires semantic alignment between predicted and gold belief propositions, for which exact string matching is too strict. We therefore validate candidate semantic judges against human alignment decisions under the same \texttt{MatchCount} protocol used for benchmark evaluation. The validation set contains 126 human-evaluated prediction tables, sampled evenly from Qwen3 32B~\citep{qwen2025qwen3}, Gemma-3 27B~\citep{gemma2025gemma3}, and Mistral-Large 123B~\citep{mistral2024large} extraction outputs.

For each validation table \(s\), let \(U^{(s)}\) denote the set of normalized \texttt{(Actor, Belief)} pairs for which both the human annotation and candidate-judge decision are defined. Human--judge agreement for table \(s\) is
\[
\mathrm{Agreement}^{(s)}
= \frac{1}{|U^{(s)}|}\sum_{(a,b)\in U^{(s)}}\mathbf{1}\!\left[y_H^{(s)}(a,b)=y_J^{(s)}(a,b)\right],
\]
where \(y_H^{(s)}(a,b)\in\{0,1\}\) is the human binary semantic-match decision and \(y_J^{(s)}(a,b)\in\{0,1\}\) is the candidate judge decision. A positive decision corresponds to \(\mathrm{MatchCount}(a,b)>0\). We macro-average \(\mathrm{Agreement}^{(s)}\) across the validation tables, giving each table equal weight.

We additionally report Cohen's \(\kappa\) between the human and judge binary decisions. Unlike raw agreement, \(\kappa\) adjusts for agreement expected from the two binary decision distributions. The 95\% confidence interval for table-macro agreement is estimated by resampling validation tables, and the \(\kappa\) interval likewise resamples whole validation tables so that beliefs from the same table remain grouped.

Table~\ref{tab:judge-human-validation} reports the candidate comparison. GPT-5~\citep{openai2025gpt5} obtains the highest table-macro agreement with human decisions, at 87.76\% $\pm$ 2.88, and a Cohen's \(\kappa\) of 0.7572 $\pm$ 0.0624. It is followed by Gemini-2.5 Flash~\citep{comanici2025gemini25} and Claude-Sonnet-4.5~\citep{anthropic2025claude}. Llama-3.3 70B~\citep{meta2024llama33} and DeepSeek-R1-Distill-Qwen 32B~\citep{deepseekai2025deepseekr1} show lower agreement under both measures. Based on this validation, we select GPT-5 as the semantic judge used for Stage~1 benchmark evaluation.

\begin{table}[!htbp]
  \setlength{\belowcaptionskip}{4pt}
\centering
\caption{Agreement between candidate semantic judges and human Stage~1 semantic-alignment decisions. Agreement is macro-averaged across the 126 human-evaluated validation tables. Values are followed by symmetric margins for 95\% confidence intervals, rounded to the precision shown.}
\label{tab:judge-human-validation}
\begin{adjustbox}{max width=\textwidth,center}
\begin{tabular}{@{}lrr@{}}
\toprule
\textbf{Semantic judge} & \shortstack{\textbf{Agreement}\\\textbf{(95\% CI)}} & \shortstack{\textbf{Cohen's \(\kappa\)}\\\textbf{(95\% CI)}} \\
\midrule
GPT-5 & \textbf{87.76 $\pm$ 2.88} & \textbf{0.7572 $\pm$ 0.0624} \\
Gemini-2.5 Flash & \underline{85.80 $\pm$ 3.26} & 0.6869 $\pm$ 0.0653 \\
Claude-Sonnet-4.5 & 84.36 $\pm$ 3.33 & \underline{0.6908 $\pm$ 0.0683} \\
Llama-3.3 70B & 75.85 $\pm$ 3.42 & 0.4442 $\pm$ 0.0557 \\
DeepSeek-R1-Distill-Qwen 32B & 74.06 $\pm$ 4.04 & 0.3754 $\pm$ 0.0834 \\
\bottomrule
\end{tabular}
\end{adjustbox}
\end{table}

We also compare the resulting Stage~1 \(F_1\) scores when semantic alignment is performed by humans and by each candidate judge. This provides a downstream check on how agreement differences translate into the benchmark metric itself. As shown in Table~\ref{tab:judge-human-f1}, GPT-5 has the smallest mean absolute difference from human-scored \(F_1\).

\begin{table}[!htbp]
  \setlength{\belowcaptionskip}{4pt}
\centering
\caption{Human-scored and semantic-judge-scored Stage~1 \(F_1\) on the human validation predictions. All values are percentages.}
\label{tab:judge-human-f1}
\begin{adjustbox}{max width=\textwidth,center}
\begin{tabular}{@{}>{\raggedright\arraybackslash}p{0.38\textwidth}rrr@{}}
\toprule
\textbf{Semantic judge} & \textbf{Human \(F_1\)} & \textbf{Judge \(F_1\)} & \textbf{Mean abs. diff.} \\
\midrule
GPT-5 & 58.21 & 64.51 & \textbf{10.02} \\
Gemini-2.5 Flash & 57.74 & 69.33 & \underline{12.63} \\
Claude-Sonnet-4.5 & 57.74 & 65.71 & 13.79 \\
Llama-3.3 70B & 57.74 & 75.29 & 19.10 \\
DeepSeek-R1-Distill-Qwen 32B & 57.74 & 67.82 & 24.94 \\
\bottomrule
\end{tabular}
\end{adjustbox}
\end{table}

\paragraph{Human MatchCount distribution.}
Finally, we inspect the human semantic-alignment annotations to verify that the one-to-many \texttt{MatchCount} range used by the judge prompt covers the alignments observed in the validation set. Across the 126 human-evaluated prediction tables, the merged human alignment set contains 4{,}715 beliefs. The \texttt{MatchCount} distribution is \texttt{MC=0}: 42.46\% (2{,}002 beliefs), \texttt{MC=1}: 56.22\% (2{,}651), \texttt{MC=2}: 1.29\% (61), and \texttt{MC=3}: 0.02\% (1). No human alignment requires a count above 3, supporting the range used for compound belief matches.

\FloatBarrier

\subsection{Reasoning-Configuration Comparisons}
\label{app:reasoning-configurations}

We compared reasoning configurations on the 21-story human-annotated validation
subset to determine whether additional thinking or reasoning effort improved the
configurations used for benchmark construction. Stage~1 reports belief-extraction
$F_1$, Stage~2 reports overall label accuracy, and the final column reports
Stage~1 semantic-judge agreement with human alignment decisions.

\begin{table}[H]
  \setlength{\belowcaptionskip}{4pt}
\centering
\caption{Reasoning-configuration comparisons on the 21-story human-annotated validation subset (\%). Stage~1 and Stage~2 use Level~4 prompts with two few-shot examples. GPT-5 is not evaluated as a Stage~1 annotation model because it serves as the semantic judge. Judge agreement is table-macro agreement with human Stage~1 semantic-alignment decisions.}
\label{tab:reasoning-configuration-comparison}
\begin{adjustbox}{max width=\textwidth,center}
\begin{tabular}{@{}>{\raggedright\arraybackslash}p{0.38\textwidth}rrr@{}}
\toprule
\textbf{Model and setting} & \textbf{Stage~1 $F_1$} & \textbf{Stage~2 accuracy} & \textbf{Judge agreement} \\
\midrule
Gemini-2.5 Flash, no-think & \textbf{65.33} & \textbf{92.97} & 85.80 \\
Gemini-2.5 Flash, think & \underline{64.59} & \underline{92.92} & 85.44 \\
GPT-5, low reasoning & -- & 92.00 & \underline{87.76} \\
GPT-5, medium reasoning & -- & \textbf{92.46} & \textbf{87.89} \\
\bottomrule
\end{tabular}
\end{adjustbox}
\end{table}

As shown in Table~\ref{tab:reasoning-configuration-comparison}, thinking or higher reasoning effort does not produce a substantial or consistent improvement for Gemini Flash or GPT-5 across the validation metrics. We therefore retain the no-thinking Gemini Flash configuration and the low-reasoning GPT-5 configuration for cost and efficiency.

\FloatBarrier

\subsection{Benchmark Record Format}
\label{app:benchmark-record-format}

The released benchmark is serialized as one JSON object per story in \texttt{benchmark\_story\_belief\_labels.jsonl}. Table~\ref{tab:benchmark-record-format} defines the fields included in each released record.

\begin{table}[!htbp]
  \setlength{\belowcaptionskip}{4pt}
\centering
\caption{Released benchmark record format in \texttt{benchmark\_story\_belief\_labels.jsonl}.}
\label{tab:benchmark-record-format}
\begin{tabularx}{\textwidth}{p{0.44\textwidth}p{0.09\textwidth}>{\raggedright\arraybackslash}X}
\toprule
\textbf{Field} & \textbf{Type} & \textbf{Description} \\
\midrule
\RecordField{story\_id} & integer & Unique story identifier. \\
\RecordField{story\_category} & string & One of the seven retained benchmark categories. \\
\RecordField{story} & string & Raw story text used for both stages. \\
\RecordField{beliefs} & array & List of annotated belief objects for the story. \\
\RecordField{beliefs[].actor} & string & Belief holder; \texttt{world} is reserved for narrator/world facts. \\
\RecordField{beliefs[].belief} & string & Minimal propositional belief statement. \\
\RecordField{beliefs[].labels} & object & Seven-dimensional schema-label object for the belief. \\
\RecordField{beliefs[].labels.order} & string & Recursive order in \(\{0,1,2,3\}\). \\
\RecordField{beliefs[].labels.truth\_status} & string & \texttt{True}, \texttt{False}, or \texttt{Unknown}. \\
\RecordField{beliefs[].labels.knowledge\_access} & string & \texttt{Private}, \texttt{Shared}, or \texttt{Public}. \\
\RecordField{beliefs[].labels.representation} & string & \texttt{Explicit} or \texttt{Implicit}. \\
\RecordField{beliefs[].labels.content\_type} & string & One of \texttt{Location}, \texttt{Contents/Physical State}, \texttt{Identity/Relation}, \texttt{Epistemic}, \texttt{Desire/Intention}, \texttt{Emotion}, \texttt{Trait/Value}, or \texttt{Action/Event}. \\
\RecordField{beliefs[].labels.mental\_source} & string & One of \texttt{Narration}, \texttt{Perception}, \texttt{Memory}, \texttt{Testimony}, \texttt{Inference}, \texttt{Imagination}, or \texttt{Unknown}. \\
\RecordField{beliefs[].labels.context} & string & \texttt{Deceptive}, \texttt{Temporal}, \texttt{Counterfactual}, or \texttt{Neutral}. \\
\bottomrule
\end{tabularx}
\end{table}

\FloatBarrier

\subsection{Licensing and Release Notes}
\label{app:licensing-release}

OmniToM reuses story text from ToMBench~\citep{chen-etal-2024-tombench}. The public ToMBench repository distributes its released materials under the MIT License, and the OmniToM release preserves upstream attribution for the source stories in the release README. The OmniToM annotations, prompt builders, replication code, dataset card, and Croissant metadata with Responsible AI fields are likewise released under the MIT License.

The public code release includes the benchmark JSONL file, prompt-builder scripts, replication runner, README, dataset card, and Croissant/RAI metadata. The benchmark is also available through the linked Hugging Face dataset. Third-party model APIs and hosted inference services used during construction or evaluation are not redistributed as part of the release artifact; users who extend the public runner to such services must follow the corresponding provider terms. Following ToMBench's benchmark-use caution, we recommend using OmniToM for evaluation rather than training to reduce benchmark contamination, although this recommendation is advisory and separate from the data/software license.

\FloatBarrier

\appsection{Prompt Engineering via TELeR Taxonomy}
\label{app:prompt-engineering-teler}
\label{app:system-prompts}

This appendix records the TELeR prompt taxonomy used to specify OmniToM prompt families, followed by the core prompt protocol used in benchmark construction, zero-shot evaluation, and semantic-judge validation. The prompt bodies below print the L3 base prompts and shared L4 augmentation blocks, indicating where category-conditioned instructions and few-shot in-context examples are inserted. Full example bodies are omitted for space; worked annotation examples appear in App.~\ref{app:category-examples}.

\subsection{TELeR Prompt Taxonomy}
\label{app:teler-taxonomy}

TELeR defines a prompt for a complex task as the combination of a \emph{directive} and the associated \emph{data} supplied to the model~\citep{karmaker-santu-feng-2023-teler}. The taxonomy categorizes the directive along four dimensions: \emph{Turn}, distinguishing single-turn from multi-turn prompting; \emph{Expression}, distinguishing question-style from instruction-style directives; \emph{Level of Details}, measuring how much task structure is provided; and \emph{Role}, distinguishing prompts with a defined system role from prompts without one.

The Level-of-Details axis ranges from Level~0, where no directive is given beyond the task data, to Level~6, where a complex directive includes a high-level goal, explicit sub-tasks, evaluation or few-shot guidance, additional retrieved information, and an explicit request for output justification. OmniToM uses system-role defined, single-turn, instruction-style prompts. The directives are task-oriented rather than question-based because benchmark construction and evaluation require structured tables rather than conversational answers.
\begin{center}
    \includegraphics[width=0.98\textwidth]{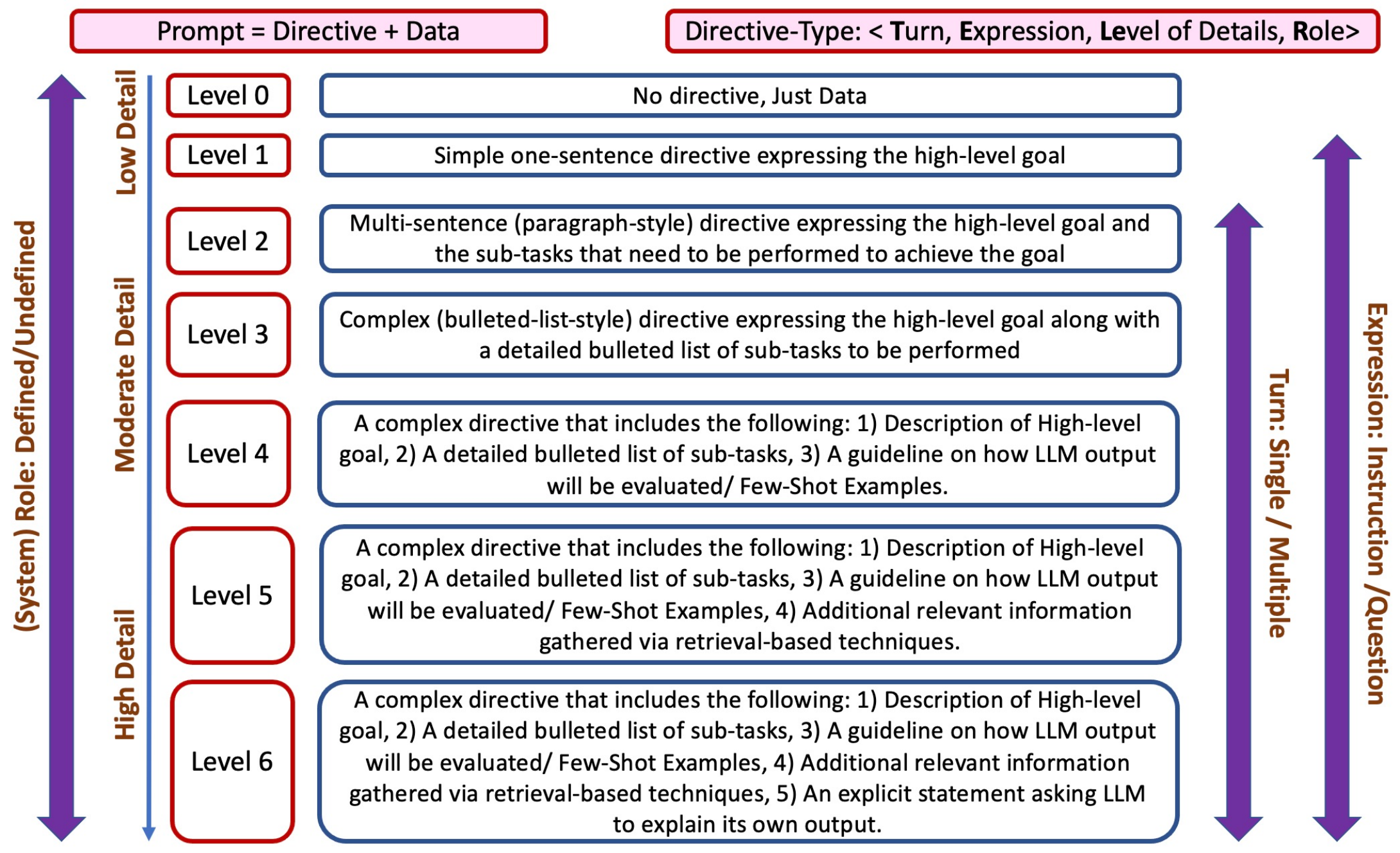}
    \captionof{figure}{TELeR prompt taxonomy from \citet{karmaker-santu-feng-2023-teler}. TELeR categorizes prompts by \textbf{T}urn, \textbf{E}xpression, \textbf{Le}vel of Details, and \textbf{R}ole. Reproduced unchanged under \href{https://creativecommons.org/licenses/by/4.0/}{CC BY 4.0}.}
    \label{fig:teler-prompt-taxonomy}
\end{center}

For benchmark construction, we adopt TELeR Level~4 prompts. Level~4 is the most appropriate point on the taxonomy for OmniToM because belief extraction, belief labeling, and semantic matching all require explicit goals, decomposed subtasks, output/evaluation criteria, and examples of the expected structure. Level~5 would introduce retrieved external information, which is inappropriate because all evidence must come from the story and supplied belief tables. Level~6 would solicit explanations or justifications, which conflicts with the table-only output format needed for reliable parsing and evaluation. For zero-shot benchmark evaluation, we use Level~3 base prompts without few-shot examples, so models are evaluated on task understanding from the directive and subtask decomposition alone.

\subsection{Stage~1 Belief-Extraction Prompt}

\paragraph{Usage in the benchmark.}
Benchmark construction and validation use a TELeR Level~4 extraction prompt with a category-conditioned evaluation block and category-matched few-shot examples. Zero-shot benchmark evaluation uses the Level~3 base prompt without few-shot examples.

\begin{table}[!htbp]
  \setlength{\belowcaptionskip}{4pt}
\centering
\caption{Category-conditioned Stage~1 extraction emphases used in Level~4 construction prompts.}
\label{tab:category-conditioned-extraction}
\begin{adjustbox}{max width=\textwidth,center}
\begin{tabular}{p{0.25\textwidth}p{0.69\textwidth}}
\toprule
\textbf{Category} & \textbf{Extraction emphasis} \\
\midrule
Ambiguous Story Task & Preserve alternative plausible interpretations supported by limited or ambiguous cues, including higher-order beliefs about others' interpretations when present. \\
False Belief Task & Capture the divergence between reality and an uninformed actor's outdated belief, including informed updates and higher-order beliefs about who knows the hidden change. \\
Faux-pas Recognition Test & Represent beliefs about whether an utterance or action is socially inappropriate, who recognizes the faux pas, and any inferred social consequences. \\
Hinting Task Test & Capture the speaker's indirectly communicated practical goal and the listener's inferred belief about the intended request or need. \\
Persuasion Story Task & Capture the persuader's intended outcome, beliefs about how the target can be influenced, and any implied target-side belief change. \\
Scalar Implicature Test & Preserve scalar reasoning over quantifiers or degree terms and any belief updates induced by counting, clarification, or disambiguation. \\
Strange Story Task & Preserve both literal and intended meaning when pragmatic interpretation is required, together with beliefs explaining why the behavior appears socially or logically odd. \\
\bottomrule
\end{tabular}
\end{adjustbox}
\end{table}

\begin{promptblock}{Stage~1 Extraction: Level~3 Zero-Shot Base Prompt}
\begin{prompttext}
You are a Theory of Mind expert whose task is to extract multi-order actor beliefs from the narrative and output a table with columns Actor, Belief, and Order by performing the following steps. A belief is a minimal proposition expressing what an actor takes to be true.
1. Identify narrated events and states that the story presents as facts, and record them as world-level beliefs attributed to the special actor 'world' (order 0).
2. Identify all actors, including characters or groups, who appear in the narrative and are capable of holding beliefs.
3. For each actor, extract beliefs about the narrated events or states of the world, and record them as first-order beliefs (order 1).
4. For each actor, extract beliefs about other actors’ beliefs, applying this notion recursively for nested beliefs, and record them as higher-order beliefs (order 2 or higher).
\end{prompttext}
\end{promptblock}

\begin{promptblock}{Stage~1 Extraction: Shared Level~4 Content Block}
\begin{prompttext}
A good output should satisfy the following:
- Include only beliefs grounded in the narrative; do not invent actors, events, or beliefs.
- Express each belief as a single, atomic, declarative proposition; split compound statements into separate rows.
- Use third-person language only and resolve pronouns to exact actor names as they appear in the narrative.
\end{prompttext}
\end{promptblock}

\begin{promptblock}{Stage~1 Extraction: Category-Conditioned Level~4 Insertion}
\begin{prompttext}
A category-conditioned instruction block is inserted here to name the story category and highlight the relevant extraction emphasis.
Categories: Ambiguous Story Task; False Belief Task; Faux-pas Recognition Test; Hinting Task Test; Persuasion Story Task; Scalar Implicature Test; Strange Story Task.
\end{prompttext}
\end{promptblock}

\begin{promptblock}{Stage~1 Extraction: Level~4 Formatting Block}
\begin{prompttext}
A correctly formatted output should satisfy the following:
- Present the result as a pipe-separated table using the '|' character.
- The first row must be exactly: Actor | Belief | Order
- Order rows with world-level beliefs first when present, then beliefs grouped by actor; within each actor group, order beliefs chronologically by the story’s event sequence.
Few-shot: Category-matched in-context examples are then provided to illustrate the expected extraction format; see App.~\ref{app:category-examples} for worked examples.
\end{prompttext}
\end{promptblock}

\subsection{Stage~2 Belief-Labeling Prompt}

\paragraph{Usage in the benchmark.}
Benchmark construction and validation use a TELeR Level~4 belief-labeling prompt with category-matched labeled few-shot tables. Zero-shot benchmark evaluation uses the Level~3 base prompt without few-shot examples.

\begin{promptblock}{Stage~2 Belief Labeling: Level~3 Zero-Shot Base Prompt}
\begin{prompttext}
You are a Theory of Mind expert whose task is to label a table of actor beliefs, given a narrative, by assigning a label from each of the following closed sets—Order (0/1/2/3), Truth-Status (True/False/Unknown), Knowledge-Access (Private/Shared/Public), Representation (Explicit/Implicit), Content Type (Location, Contents/Physical State, Identity/Relation, Epistemic, Desire/Intention, Emotion, Trait/Value, Action/Event), Mental-Source (Narration, Perception, Memory, Testimony, Inference, Imagination, Unknown), and Context (Deceptive, Temporal, Counterfactual, Neutral)—and outputting only a table with columns Actor and Belief, followed by one column for each labeling set.

In this context, a belief is a minimal proposition expressing what an actor takes to be true about the world or about another actor’s mental state. Label each belief in the provided table by assigning values for the following dimensions, using the narrative as evidence:

1. Determine the Order of the belief, which captures the depth of belief reasoning:
   - Order 0: Narrator- or world-level facts that anchor the story’s ground truth and are not held by any actor.
   - Order 1: First-order beliefs (A believes p).
   - Order 2: Second-order beliefs (A believes B believes p).
   - Order 3: Higher-order recursive beliefs (A believes B believes C believes p).

2. Determine the Truth-Status of the belief relative to the narrative:
   - True if the belief is verified or entailed by the narration.
   - False if the belief is contradicted by the narration.
   - Unknown if the narrative does not provide sufficient evidence.

3. Determine the Knowledge-Access of the belief by assessing who could realistically know it in the story world:
   - Private if the belief is held internally without evidence others know it.
   - Shared if it is mutually known within a subgroup through explicit acknowledgment or obvious mutual awareness.
   - Public if it is common ground across all actors (announced, jointly witnessed, or mutually known to be mutually known).

4. Determine the Representation of the belief:
   - Explicit if the belief is directly stated, spoken, or narrated as a mental state.
   - Implicit if the belief must be inferred from actions, perception, or context.

5. Determine the Content Type by identifying what the proposition is about:
   - Use Action/Event for happenings; Desire/Intention for plans or goals.
   - Use Location when the proposition concerns where an entity is or was, even if it involves a container
   - Use Contents / Physical State only when the belief concerns what a container holds or an object’s condition.
   - Use Epistemic for beliefs about beliefs, knowledge, attention, or awareness.

6. Determine the Mental-Source of the belief, indicating how it was acquired:
   - Narration (Order 0 only), Perception, Memory, Testimony, Inference, Imagination, or Unknown.

7. Determine the Context of the belief:
   - Deceptive if shaped by lying, omission, or misdirection.
   - Temporal if the belief is outdated or reflects recall of a prior true state.
     - Temporal + False indicates an outdated false belief.
     - Temporal + True indicates accurate recall of a past fact.
   - Counterfactual if the belief occurs in a hypothetical or pretense frame.
   - Neutral if none apply.
\end{prompttext}
\end{promptblock}

\begin{promptblock}{Stage~2 Belief Labeling: Level~4 Addition Block}
\begin{prompttext}
A good output should satisfy the following:
- Assign labels only using evidence grounded in the narrative; do not introduce new interpretations, entities, beliefs, or label values.
- Assign exactly one valid label from each closed set to every belief line.
- Use only the predefined label sets for Order, Truth-Status, Knowledge-Access, Representation, Content Type, Mental-Source, and Context.
- If evidence is insufficient or ambiguous, select Unknown rather than speculate.
- Do not include explanations, notes, justifications, or commentary beyond the required labels.

A correctly formatted output should satisfy the following:
- Present the result as a pipe-separated table using the '|' character.
- Each subsequent row must correspond to one belief from the input belief table.
- Preserve the original order of beliefs from the input table.
- Do not include free-text explanations, narrative descriptions, or additional columns.
Few-shot: Category-matched in-context examples are then provided to illustrate the expected label choices and table format; see App.~\ref{app:category-examples} for worked examples.
\end{prompttext}
\end{promptblock}

\subsection{Semantic Judge Prompt}
\label{app:judge-validation-prompts}

\paragraph{Usage in the benchmark.}
The semantic judge is not used to generate benchmark annotations. It is used for Stage~1 semantic alignment during validation and evaluation. For zero-shot evaluation, the selected GPT-5 judge uses the TELeR Level~4 family with appended few-shot alignment examples. Candidate judge models are compared on the 21-story human-annotated validation subset across TELeR Levels~1--4, and the final evaluation protocol fixes the Level~4 prompt family reproduced below.

\begin{promptblock}{Semantic Judge: Level~3 Zero-Shot Base Prompt}
\begin{prompttext}
You are a Theory of Mind evaluation expert whose task is to semantically match rows between two belief tables (Prediction, Ground Truth) extracted from the same short Story Narrative and output only the two tables, explicitly labeled “Prediction Table” and “Ground Truth Table,” with an added MatchCount column indicating how many distinct semantically equivalent rows exist in the other table for the same Actor. In this context, a Belief is a minimal statement of what an actor takes to be true about the world (facts/events) or about other actors’ mental states, expressed in natural language. Perform the task by following these steps:
1) If a Story Narrative is provided, use it only to resolve ambiguity (pronouns, aliases, implicit entities) and paraphrase meaning; if no narrative is provided, ignore narrative context entirely. In all cases, do not add rows and do not introduce new beliefs that are not present in either table.
2) Treat Actors as distinct mental agents and normalize only cosmetic variants of the same Actor name (case/spacing/punctuation and clear shortenings); never merge different Ground Truth Actors.
3) Handle the special actor 'world' first: treat 'world' as the key for narrated facts and events, and align world-level beliefs conservatively, typically one-to-one, allowing only minor normalization differences.
4) Restrict candidate matches to the same Actor group after normalization; if the Actor does not match, the row cannot match regardless of belief similarity.
5) Default to one-to-one with bookkeeping: if (and only if) there exists a clear semantically equivalent belief for the same Actor, assign the row its single best match among currently-unmatched target rows; otherwise assign no match (MatchCount = 0). If multiple rows compete for the same target row, keep only the closest semantic match and force the others to choose different unmatched targets or become 0.
6) Allow one-to-many only for compound rows: if a row clearly contains multiple independent beliefs, you may align it to 2–3 different rows in the other table within the same Actor group, but only if each aligned target row captures a distinct part of the compound meaning.
7) Ensure symmetry: after completing matches for Prediction rows, also compute MatchCount for every Ground Truth row using the same alignment decisions.
\end{prompttext}
\end{promptblock}

\begin{promptblock}{Semantic Judge: Level~4 Addition Block}
\begin{prompttext}
A good output should satisfy the following:
- Only compare beliefs inside the same Actor group; Ground Truth Actors are unique and must not be merged—if Actor differs, it is NOT a match even if belief text is identical.
- The special actor 'world' represents narrated facts and events (not a character in the story); world-level beliefs should generally align one-to-one across tables with only minor normalization differences.
- Match only beliefs that are semantically equivalent; do not invent or force alignments and do not introduce beliefs that are not present in either table.
- Use conservative alignment: when multiple rows are 'close', prefer the best single match and leave other rows unmatched rather than double-counting.
- Prefer one-to-one matches; allow one-to-many (MatchCount 2–3) only for genuinely compound beliefs.
- Do not match across different Actors, even if text looks similar.
- Output exactly two tables labeled Prediction and Ground Truth (in that order), with CSV header Actor,Belief,MatchCount and no extra text.

A correctly formatted output should satisfy the following:
- Output exactly two tables labeled Prediction and Ground Truth (in that order).
- Each table must be comma-separated with header: Actor,Belief,MatchCount
- Preserve original row order and add exactly one MatchCount column.
- Do not include explanations or any text outside the tables.
Few-shot: In-context alignment examples are then provided to illustrate the matching process.
\end{prompttext}
\end{promptblock}

\subsection{Prompt-Selection Results}

Table~\ref{tab:prompt-level-comparison} summarizes the prompt configurations evaluated during benchmark construction and semantic-judge validation. For Stage~1 and Stage~2, Levels~1--3 use zero few-shot examples and Level~4 uses the selected \texttt{L4\_fs2} configuration with two few-shot examples. For semantic-judge validation, Levels~1--3 use zero examples and Level~4 uses the selected \texttt{L4\_fs3} configuration with three few-shot examples.

\begin{table}[!htbp]
  \setlength{\belowcaptionskip}{4pt}
\centering
\caption{Prompt-selection results on the 21-story human-annotated validation subset (\%). Stage~1 and Stage~2 use two few-shot examples at Level~4. Semantic-judge rows use zero few-shot examples at Levels~1--3 and three few-shot examples at Level~4.}
\label{tab:prompt-level-comparison}
\begin{adjustbox}{max width=\textwidth,center}
\begin{tabular}{@{}lrrrr@{}}
\toprule
\textbf{Model} & \textbf{L1} & \textbf{L2} & \textbf{L3} & \textbf{L4} \\
\midrule
\multicolumn{5}{@{}l}{\textbf{Stage~1: Belief Extraction (\(F_1\))}} \\
Llama-3.3 70B
& 36.18 & 36.97 & \underline{45.47} & \textbf{67.50} \\
Mistral-Large 123B
& 37.59 & 44.17 & \underline{53.14} & \textbf{70.13} \\
Gemma-3 27B
& 29.42 & \underline{53.72} & 51.67 & \textbf{64.99} \\
Qwen3 32B
& 32.24 & 46.32 & \underline{49.45} & \textbf{67.82} \\
\midrule
\multicolumn{5}{@{}l}{\textbf{Semantic Judge: Human--Judge Agreement}} \\
Llama-3.3 70B
& 74.31 & 77.07 & \underline{77.19} & \textbf{79.16} \\
GPT-5
& 85.13 & \underline{87.62} & 87.54 & \textbf{87.76} \\
\midrule
\multicolumn{5}{@{}l}{\textbf{Stage~2: Belief Labeling (Overall Accuracy)}} \\
Llama-3.3 70B
& 69.18 & 78.71 & \underline{81.60} & \textbf{93.08} \\
Mistral-Large 123B
& 77.96 & 85.22 & \underline{89.49} & \textbf{93.76} \\
Gemma-3 27B
& 79.79 & \underline{83.60} & 81.47 & \textbf{92.94} \\
Qwen3 32B
& 75.05 & 81.10 & \underline{81.34} & \textbf{91.14} \\
\bottomrule
\end{tabular}
\end{adjustbox}
\end{table}

The results show that the selected Level~4 configuration performs best overall in the validation sweep. Llama-3.3 reaches its highest agreement at Level~4, and GPT-5 reaches 87.76 with the selected Level~4 \texttt{L4\_fs3} configuration, slightly above its Level~2 and Level~3 results. For Stage~1 and Stage~2, the selected Level~4 configuration also performs best for all four annotation models.

\FloatBarrier

\appsection{Qualitative Comparison with Endpoint Question Answering}
\label{app:qualitative-comparison}

This appendix compares ToMBench endpoint question answering with the belief representations elicited separately by OmniToM. Appendix~\ref{app:qualitative-correct} shows cases where the model answers all linked endpoint questions correctly despite errors elsewhere in its belief representation. Appendix~\ref{app:qualitative-incorrect} shows incorrect endpoint answers together with related belief-tracking errors that make the model's representation directly inspectable. Because the two outputs are elicited separately, the belief representation should not be interpreted as a hidden chain of thought or as establishing what caused an endpoint answer. Green marks information captured by at least one correct linked endpoint answer. It does not certify every green proposition as globally correct. In the belief representation, muted red marks a belief-tracking error that is contradicted by the story, inconsistent with the model's other beliefs, or unsupported by the available information. In endpoint QA, muted red marks an incorrect selected answer. Stories, questions, answer choices, and belief propositions reproduce the saved outputs verbatim.

\definecolor{QualitativeError}{RGB}{196,61,61}
\definecolor{QualitativeCorrect}{RGB}{42,142,80}
\definecolor{QualitativeQA}{RGB}{255,248,222}
\definecolor{QualitativeQABorder}{RGB}{198,165,71}
\definecolor{QualitativeBelief}{RGB}{225,241,229}
\definecolor{QualitativeBeliefBorder}{RGB}{38,105,66}

\newtcolorbox{qualitativeexample}[1]{%
  breakable,
  enforce breakable,
  pad at break=4pt,
  enlargepage flexible=\baselineskip,
  toprule at break=0pt,
  bottomrule at break=0pt,
  colback=black!5,
  colframe=black!48,
  colbacktitle=black!13,
  coltitle=black,
  fonttitle=\small\bfseries,
  title={#1},
  boxrule=0.8pt,
  arc=2pt,
  left=8pt,
  right=8pt,
  top=7pt,
  bottom=8pt,
  toptitle=3pt,
  bottomtitle=3pt,
  before skip=10pt,
  after skip=10pt
}

\newtcolorbox{qualitativeqa}{%
  colback=QualitativeQA,
  colframe=QualitativeQABorder,
  boxrule=0.7pt,
  arc=1.5pt,
  left=7pt,
  right=7pt,
  top=5pt,
  bottom=5pt,
  before skip=8pt,
  after skip=8pt
}

\newtcolorbox{qualitativebeliefs}{%
  colback=QualitativeBelief,
  colframe=QualitativeBeliefBorder,
  boxrule=0.7pt,
  arc=1.5pt,
  left=7pt,
  right=7pt,
  top=5pt,
  bottom=5pt,
  before skip=8pt,
  after skip=7pt
}

\newcommand{\qualitativelabel}[1]{\par\smallskip\noindent\textbf{#1}\par\vspace{2pt}}
\newcommand{\choiceplain}[1]{{\footnotesize\strut #1\strut}}
\newcommand{\qualcheck}{\hspace{0.25em}\ding{51}}
\newcommand{\qualcross}{\hspace{0.25em}\ding{55}}
\newcommand{\choicegold}[1]{{\footnotesize\textcolor{QualitativeCorrect!85!black}{\textbf{\strut #1\qualcheck\strut}}}}
\newcommand{\choicewrong}[1]{{\footnotesize\textcolor{QualitativeError}{\textbf{\strut #1\qualcross\strut}}}}
\newcommand{\choiceinlineplain}[1]{\footnotesize\strut #1\strut}
\newcommand{\choiceinlinegold}[1]{\footnotesize\textcolor{QualitativeCorrect!85!black}{\textbf{\strut #1\qualcheck\strut}}}
\newcommand{\choiceinlinewrong}[1]{\footnotesize\textcolor{QualitativeError}{\textbf{\strut #1\qualcross\strut}}}
\newcommand{\beliefheader}{\noindent\begin{minipage}[t]{0.20\linewidth}\footnotesize\itshape Actor\end{minipage}\hspace{0.02\linewidth}\begin{minipage}[t]{0.77\linewidth}\footnotesize\itshape Belief proposition\end{minipage}\par\vspace{2pt}\hrule\vspace{3pt}}
\newcommand{\beliefrow}[2]{\noindent\begin{minipage}[t]{0.20\linewidth}\footnotesize\raggedright #1\end{minipage}\hspace{0.02\linewidth}\begin{minipage}[t]{0.77\linewidth}\footnotesize\raggedright #2\end{minipage}\par\vspace{1pt}}
\newcommand{\beliefrowerror}[2]{\noindent\begin{minipage}[t]{0.20\linewidth}\footnotesize\raggedright #1\end{minipage}\hspace{0.02\linewidth}\begin{minipage}[t]{0.77\linewidth}\footnotesize\raggedright\textcolor{QualitativeError}{\textbf{#2}}\end{minipage}\par\vspace{1pt}}
\newcommand{\beliefrowsupport}[2]{\noindent\begin{minipage}[t]{0.20\linewidth}\footnotesize\raggedright #1\end{minipage}\hspace{0.02\linewidth}\begin{minipage}[t]{0.77\linewidth}\footnotesize\raggedright\textcolor{QualitativeCorrect!85!black}{\textbf{#2}}\end{minipage}\par\vspace{1pt}}

\subsection{Correct Endpoint Answers with Belief-Tracking Errors}
\label{app:qualitative-correct}

These examples show that correct endpoint answers can coexist with errors elsewhere in the separately elicited multi-actor belief representation. Green highlights the information that is reflected in the correct endpoint answers, while muted red highlights belief-tracking errors that those questions do not reveal.

\begin{qualitativeexample}{False Belief Task \quad Llama-3.3 70B}
\small
\qualitativelabel{Story}
Xiao Gang and Zhang Wei are strolling on the terrace, they see a box and a drawer, they find a shirt in the box, Zhang Wei leaves the terrace, Xiao Gang moves the shirt to the drawer.

\begin{qualitativeqa}
\textbf{ToMBench Question Answering}\par\smallskip
\footnotesize Where is the shirt now?\par\smallskip
\choiceinlineplain{A. Cabinet}\quad \choiceinlinegold{B. Drawer}\quad \choiceinlineplain{C. Suitcase}\quad \choiceinlineplain{D. Box}\par\medskip
\footnotesize Where is the shirt before?\par\smallskip
\choiceinlineplain{A. Drawer}\quad \choiceinlineplain{B. Handbag}\quad \choiceinlinegold{C. Box}\quad \choiceinlineplain{D. Basket}\par\medskip
\footnotesize After Zhang Wei returns to the terrace, where does Xiao Gang look for the shirt?\par\smallskip
\choiceinlinegold{A. Drawer}\quad \choiceinlineplain{B. Storage locker}\quad \choiceinlineplain{C. Backpack}\quad \choiceinlineplain{D. Box}\par\medskip
\footnotesize After Zhang Wei returns to the terrace, where does Zhang Wei look for the shirt?\par\smallskip
\choiceinlineplain{A. Briefcase}\quad \choiceinlineplain{B. Cabinet}\quad \choiceinlinegold{C. Box}\quad \choiceinlineplain{D. Drawer}\par\medskip
\footnotesize After Zhang Wei returns to the terrace, where does Xiao Gang think Zhang Wei looks for the shirt?\par\smallskip
\choiceinlineplain{A. Locker}\quad \choiceinlinegold{B. Box}\quad \choiceinlineplain{C. Drawer}\quad \choiceinlineplain{D. Backpack}\par\medskip
\footnotesize After Zhang Wei returns to the terrace, where does Zhang Wei think Xiao Gang looks for the shirt?\par\smallskip
\choiceinlineplain{A. Drawer}\quad \choiceinlineplain{B. Case}\quad \choiceinlineplain{C. Basket}\quad \choiceinlinegold{D. Box}
\end{qualitativeqa}

\begin{qualitativebeliefs}
\textbf{OmniToM Belief Representation}\par\smallskip
\beliefheader
\beliefrow{world}{there is a box on the terrace}
\beliefrow{world}{there is a drawer on the terrace}
\beliefrowsupport{world}{there is a shirt in the box initially}
\beliefrowsupport{world}{Zhang Wei leaves the terrace}
\beliefrowsupport{world}{Xiao Gang moves the shirt to the drawer}
\beliefrow{Xiao Gang}{there is a box on the terrace}
\beliefrow{Xiao Gang}{there is a drawer on the terrace}
\beliefrow{Xiao Gang}{there is a shirt in the box initially}
\beliefrow{Xiao Gang}{Zhang Wei leaves the terrace}
\beliefrowsupport{Xiao Gang}{he moves the shirt to the drawer}
\beliefrow{Zhang Wei}{there is a box on the terrace}
\beliefrow{Zhang Wei}{there is a drawer on the terrace}
\beliefrowsupport{Zhang Wei}{there is a shirt in the box initially}
\beliefrow{Xiao Gang}{Zhang Wei sees a box on the terrace}
\beliefrow{Xiao Gang}{Zhang Wei sees a drawer on the terrace}
\beliefrowsupport{Xiao Gang}{Zhang Wei sees a shirt in the box initially}
\beliefrow{Zhang Wei}{Xiao Gang sees a box on the terrace}
\beliefrow{Zhang Wei}{Xiao Gang sees a drawer on the terrace}
\beliefrowsupport{Zhang Wei}{Xiao Gang sees a shirt in the box initially}
\beliefrowerror{Xiao Gang}{Zhang Wei knows Xiao Gang moves the shirt to the drawer}
\beliefrow{Zhang Wei}{Xiao Gang knows Zhang Wei leaves the terrace}
\end{qualitativebeliefs}

Xiao Gang moves the shirt only after Zhang Wei has left, so Zhang Wei has no access to the transfer. Nevertheless, the extracted representation attributes knowledge of the move to Zhang Wei. All six linked endpoint questions are answered correctly, including questions about where each actor will search and what each actor expects the other to do. The endpoint scores therefore miss a concrete error in the broader representation of who has access to the transfer.
\end{qualitativeexample}

\begin{qualitativeexample}{Hinting Task Test \quad Mistral-Small 24B}
\small
\qualitativelabel{Story}
Jessica and Manny are playing with train models. Jessica holds a blue train, Manny holds a red train. Jessica says to Manny, "I do not like this train."

\begin{qualitativeqa}
\textbf{ToMBench Question Answering}\par\smallskip
What does Jessica really want to say when she says this?\par\smallskip
\begin{tabularx}{\linewidth}{@{}>{\raggedright\arraybackslash}X>{\raggedright\arraybackslash}X@{}}
\choicegold{A. Jessica wants to swap trains with Manny.} & \choiceplain{B. Jessica indicates she does not like playing with train models.}\\[4pt]
\choiceplain{C. Jessica wants to tell Manny she does not like the color blue.} & \choiceplain{D. Jessica is complaining to Manny about her train.}
\end{tabularx}
\end{qualitativeqa}

\begin{qualitativebeliefs}
\textbf{OmniToM Belief Representation}\par\smallskip
\beliefheader
\beliefrow{world}{Jessica and Manny are playing with train models}
\beliefrow{world}{Jessica holds a blue train}
\beliefrow{world}{Manny holds a red train}
\beliefrow{world}{Jessica says to Manny, "I do not like this train."}
\beliefrowsupport{Jessica}{Jessica holds a blue train}
\beliefrowsupport{Manny}{Manny holds a red train}
\beliefrowsupport{Jessica}{Jessica does not like the blue train}
\beliefrowsupport{Manny}{Jessica does not like the blue train}
\beliefrow{Jessica}{Manny holds a red train}
\beliefrow{Manny}{Jessica holds a blue train}
\beliefrowerror{Jessica}{Manny does not know that Jessica does not like the blue train}
\beliefrowerror{Manny}{Jessica does not know that Manny does not like the red train}
\end{qualitativebeliefs}

The endpoint question correctly identifies Jessica's implied goal of swapping trains. The belief representation also captures that Manny receives Jessica's statement that she dislikes the blue train. However, it then represents Jessica as believing that Manny does not know this, creating an inconsistent account of whether the communicated information was received. The representation also introduces the unsupported proposition that Jessica does not know Manny dislikes the red train, a preference never established in the story. These errors occur outside the narrow pragmatic decision tested by the endpoint question.
\end{qualitativeexample}

\begin{qualitativeexample}{Strange Story Task \quad Gemini-2.5 Pro}
\small
\qualitativelabel{Story}
Today, Bohan goes to Xiuya's house for the first time. He goes to her house for coffee and looks forward to seeing the dog that Xiuya often mentions. Bohan really likes dogs. When Bohan arrives at Xiuya's house, Xiuya runs to open the door, and her dog jumps up to welcome Bohan. Xiuya's dog is very big, almost as big as Bohan! When Bohan sees Xiuya's huge dog, he says, "Xiuya, you don't have a dog at all. You have an elephant!"

\begin{qualitativeqa}
\textbf{ToMBench Question Answering}\par\smallskip
Why does Bohan say this?\par\smallskip
\begin{tabularx}{\linewidth}{@{}>{\raggedright\arraybackslash}X>{\raggedright\arraybackslash}X@{}}
\choiceplain{A. Bohan really misunderstands that Xiuya keeps an elephant.} & \choiceplain{B. Bohan lies to make Xiuya laugh.}\\[4pt]
\choiceplain{C. Bohan jokes because he also wants to keep a dog.} & \choicegold{D. Bohan jokes to make Xiuya laugh.}
\end{tabularx}
\end{qualitativeqa}

\begin{qualitativebeliefs}
\textbf{OmniToM Belief Representation}\par\smallskip
\beliefheader
\beliefrow{world}{Bohan went to Xiuya's house for the first time}
\beliefrowsupport{world}{Xiuya has a dog}
\beliefrowsupport{world}{Xiuya's dog is very big}
\beliefrowsupport{world}{Bohan likes dogs}
\beliefrow{world}{Bohan said to Xiuya that she has an elephant and not a dog}
\beliefrowsupport{Bohan}{Xiuya has a dog}
\beliefrowsupport{Bohan}{Xiuya's dog is huge}
\beliefrowerror{Bohan}{Xiuya has an elephant}
\beliefrowerror{Bohan}{Xiuya does not have a dog}
\beliefrow{Xiuya}{She has a dog}
\beliefrow{Bohan}{Xiuya believes she has a dog}
\beliefrowerror{Xiuya}{Bohan believes she has an elephant}
\beliefrowerror{Xiuya}{Bohan believes she does not have a dog}
\end{qualitativebeliefs}

The endpoint question correctly recognizes Bohan's statement as a joke. The belief representation also captures the underlying situation that Xiuya owns a very large dog. However, it then treats the joke itself as a literal belief, representing Bohan as believing that Xiuya has an elephant and no dog. The same literalized interpretation propagates into Xiuya's higher-order representation of Bohan. The correct pragmatic endpoint answer therefore hides several errors in how the nonliteral utterance is represented as actor beliefs.
\end{qualitativeexample}

\begin{qualitativeexample}{Strange Story Task \quad Mistral-Large 123B}
\small
\qualitativelabel{Story}
On the first day at the new job, the supervisor praises their company's new printer as the best in the industry, and Linda looks forward to using the "super easy to use" printer advertised. But when Linda tries to use it, a series of error messages appear, and she spends an hour and a manual to set up the printer. When the machine prints out the document, she breathes a sigh of relief and says, "It really is a super easy to use printer, it is easy to use."

\begin{qualitativeqa}
\textbf{ToMBench Question Answering}\par\smallskip
\footnotesize Is what Linda says true?\par\smallskip
\choiceinlineplain{A. Yes}\quad \choiceinlinegold{B. No}\par\medskip
\footnotesize Why does Linda say this?\par\smallskip
\begin{tabularx}{\linewidth}{@{}>{\raggedright\arraybackslash}X>{\raggedright\arraybackslash}X@{}}
\choicegold{A. Linda wants to mock that the printer is not very easy to use.} & \choiceplain{B. Linda wants to agree with the supervisor that the printer is very good to use.}\\[4pt]
\choiceplain{C. Linda wants to recommend the printer to her colleagues.} & \choiceplain{D. Linda thinks the printer is easy to use.}
\end{tabularx}
\end{qualitativeqa}

\begin{qualitativebeliefs}
\textbf{OmniToM Belief Representation}\par\smallskip
\beliefheader
\beliefrow{world}{The supervisor praises the company's new printer as the best in the industry}
\beliefrow{world}{Linda looks forward to using the "super easy to use" printer}
\beliefrowsupport{world}{A series of error messages appear when Linda tries to use the printer}
\beliefrowsupport{world}{Linda spends an hour and uses a manual to set up the printer}
\beliefrow{world}{The printer successfully prints out the document}
\beliefrow{supervisor}{The company's new printer is the best in the industry}
\beliefrowsupport{Linda}{The printer is "super easy to use" (before trying to use it)}
\beliefrowerror{Linda}{The printer is "super easy to use" (after successfully printing)}
\beliefrow{Linda}{The supervisor believes the printer is the best in the industry}
\end{qualitativebeliefs}

The endpoint answers correctly recognize both that Linda's statement is false and that she is mocking the printer. The belief representation also correctly preserves her initial expectation that the printer would be easy to use before she tries it. After an hour of errors and setup difficulty, however, the representation fails to revise that belief and again attributes ``super easy to use'' to Linda after printing. The endpoint therefore captures the sarcasm while the broader belief trace still contains an incorrect post-event belief state.
\end{qualitativeexample}

\subsection{Incorrect Endpoint Answers with Related Belief-Tracking Errors}
\label{app:qualitative-incorrect}

These examples pair incorrect endpoint answers with related errors in the separately elicited belief representation. Green again marks correctly captured endpoint information, while muted red marks either incorrect endpoint selections or belief-tracking errors. Because the two outputs were elicited separately, the belief representation is used to localize related reasoning failures rather than to claim a causal explanation of the endpoint answer.

\begin{qualitativeexample}{False Belief Task \quad Llama-3.1 8B}
\small
\qualitativelabel{Story}
Xiao Bei and Xiao Ming are wandering in the parking lot, they see a box and a briefcase, they find corn in the box, Xiao Ming leaves the parking lot, Xiao Bei moves the corn to the briefcase.

\begin{qualitativeqa}
\textbf{ToMBench Question Answering}\par\smallskip
\footnotesize Where is the corn now?\par\smallskip
\choiceinlinegold{A. Briefcase}\quad \choiceinlineplain{B. Basket}\quad \choiceinlineplain{C. Cabinet}\quad \choiceinlineplain{D. Box}\par\medskip
\footnotesize Where is the corn before?\par\smallskip
\choiceinlinegold{A. Box}\quad \choiceinlineplain{B. Basket}\quad \choiceinlineplain{C. Handbag}\quad \choiceinlineplain{D. Briefcase}\par\medskip
\footnotesize After Xiao Ming returns to the parking lot, where does Xiao Bei look for the corn?\par\smallskip
\choiceinlineplain{A. Cabinet}\quad \choiceinlineplain{B. Box}\quad \choiceinlinegold{C. Briefcase}\quad \choiceinlineplain{D. Handbag}\par\medskip
\footnotesize After Xiao Ming returns to the parking lot, where does Xiao Ming look for the corn?\par\smallskip
\choiceinlinewrong{A. Briefcase}\quad \choiceinlineplain{B. Cabinet}\quad \choiceinlineplain{C. Box}\quad \choiceinlineplain{D. Safe}\par\medskip
\footnotesize After Xiao Ming returns to the parking lot, where does Xiao Bei think Xiao Ming looks for the corn?\par\smallskip
\choiceinlineplain{A. Cabinet}\quad \choiceinlinewrong{B. Briefcase}\quad \choiceinlineplain{C. Backpack}\quad \choiceinlineplain{D. Box}\par\medskip
\footnotesize After Xiao Ming returns to the parking lot, where does Xiao Ming think Xiao Bei looks for the corn?\par\smallskip
\choiceinlineplain{A. Handbag}\quad \choiceinlineplain{B. Basket}\quad \choiceinlineplain{C. Box}\quad \choiceinlinewrong{D. Briefcase}
\end{qualitativeqa}

\begin{qualitativebeliefs}
\textbf{OmniToM Belief Representation}\par\smallskip
\beliefheader
\beliefrowsupport{world}{The box contains corn}
\beliefrow{world}{Xiao Ming leaves the parking lot}
\beliefrowsupport{world}{Xiao Bei moves the corn to the briefcase}
\beliefrow{world}{Xiao Bei and Xiao Ming are wandering in the parking lot}
\beliefrow{world}{There is a box and a briefcase in the parking lot}
\beliefrow{Xiao Bei}{The box contains corn}
\beliefrow{Xiao Bei}{Xiao Ming leaves the parking lot}
\beliefrow{Xiao Bei}{Xiao Ming believes the box contains corn}
\beliefrow{Xiao Bei}{Xiao Ming believes Xiao Ming leaves the parking lot}
\beliefrow{Xiao Ming}{There is a box and a briefcase in the parking lot}
\beliefrow{Xiao Ming}{The box contains corn}
\beliefrowerror{Xiao Ming}{Xiao Bei moves the corn to the briefcase}
\beliefrow{Xiao Ming}{Xiao Bei believes the box contains corn}
\beliefrow{Xiao Ming}{Xiao Bei believes Xiao Ming leaves the parking lot}
\end{qualitativebeliefs}

Xiao Ming leaves before Xiao Bei moves the corn, so he should retain the outdated belief that it remains in the box. The endpoint QA instead predicts that he will search in the briefcase. The separately elicited belief representation contains a closely related information-access error: it attributes the unseen transfer itself to Xiao Ming. OmniToM therefore makes the relevant tracking failure inspectable rather than only showing that the final answer is wrong. Because the QA and belief outputs were elicited separately, this correspondence should not be interpreted as a causal trace of how the endpoint answer was produced.
\end{qualitativeexample}

\begin{qualitativeexample}{Strange Story Task \quad Mistral-Small 24B}
\small
\qualitativelabel{Story}
Xiaohua's school has a strict rule that students who do not complete their homework cannot participate in extracurricular activities. Xiaohua enjoys extracurricular activity time. Today, Xiaohua wants to participate in the school's basketball game. Xiaohua's dad asks him, "Do you want to participate in extracurricular activities today?" Xiaohua replies, "No."

\begin{qualitativeqa}
\textbf{ToMBench Question Answering}\par\smallskip
\footnotesize Is what Xiaohua tells his dad true?\par\smallskip
\choiceinlinewrong{A. Yes}\quad \choiceinlineplain{B. No}\par\medskip
\footnotesize Why does Xiaohua clearly want to participate in the basketball game, but says there are no extracurricular activities?\par\smallskip
\begin{tabularx}{\linewidth}{@{}>{\raggedright\arraybackslash}X>{\raggedright\arraybackslash}X@{}}
\choiceplain{A. Because he actually does not like basketball.} & \choiceplain{B. Because he forgets there are extracurricular activities.}\\[4pt]
\choicegold{C. Because he fears that the teacher finds out he plays without doing homework.} & \choiceplain{D. Because he wants to play basketball secretly without his dad knowing.}
\end{tabularx}
\end{qualitativeqa}

\begin{qualitativebeliefs}
\textbf{OmniToM Belief Representation}\par\smallskip
\beliefheader
\beliefrowsupport{world}{Xiaohua's school has a strict rule that students who do not complete their homework cannot participate in extracurricular activities}
\beliefrow{world}{Xiaohua enjoys extracurricular activity time}
\beliefrowsupport{world}{Today, Xiaohua wants to participate in the school's basketball game}
\beliefrow{world}{Xiaohua's dad asks Xiaohua if he wants to participate in extracurricular activities today}
\beliefrow{world}{Xiaohua replies "No" to his dad}
\beliefrowsupport{Xiaohua}{Xiaohua's school has a strict rule that students who do not complete their homework cannot participate in extracurricular activities}
\beliefrow{Xiaohua}{Xiaohua enjoys extracurricular activity time}
\beliefrowsupport{Xiaohua}{Today, Xiaohua wants to participate in the school's basketball game}
\beliefrow{Xiaohua}{Xiaohua's dad wants to know if Xiaohua wants to participate in extracurricular activities today}
\beliefrowerror{Xiaohua}{Xiaohua does not want to participate in extracurricular activities today}
\beliefrow{Xiaohua's dad}{Xiaohua does not want to participate in extracurricular activities today}
\beliefrow{Xiaohua's dad}{Xiaohua's school has a strict rule that students who do not complete their homework cannot participate in extracurricular activities}
\beliefrow{Xiaohua's dad}{Xiaohua enjoys extracurricular activity time}
\beliefrowerror{Xiaohua's dad}{Xiaohua wants to participate in the school's basketball game}
\end{qualitativebeliefs}

The story explicitly states that Xiaohua wants to play basketball, but he tells his dad ``No.'' The endpoint model incorrectly treats that denial as true. The belief representation contains two related tracking errors. First, it records Xiaohua's deceptive denial as Xiaohua's own belief that he does not want to participate, contradicting the desire stated earlier in the story. Second, it attributes Xiaohua's actual desire to his dad even though the story never gives the dad access to that private intention. One error confuses an utterance with the speaker's belief, while the other assigns unsupported knowledge to the listener.
\end{qualitativeexample}

\FloatBarrier

\appsection{Worked Annotation Examples by Category}
\label{app:category-examples}
This appendix provides one worked annotation example for each retained benchmark category, using the final Level~4 demonstration format from benchmark construction. Each table preserves the full story text together with the gold OmniToM beliefs and schema labels, grouped by actor. Compact display labels are used for space: \textit{Identity} abbreviates \textit{Identity/Relation}, \textit{Physical} abbreviates \textit{Contents/Physical State}, \textit{Desire} abbreviates \textit{Desire/Intention}, \textit{Trait} abbreviates \textit{Trait/Value}, and \textit{Action} abbreviates \textit{Action/Event}. App.~\ref{app:atoms-coverage} lists the full closed-set labels.

\setlength{\LTpre}{8pt}
\setlength{\LTpost}{4pt}
\setlength{\LTleft}{0pt}
\setlength{\LTright}{0pt}
\setlength{\LTcapwidth}{\textwidth}

\newcolumntype{L}[1]{>{\raggedright\arraybackslash}p{#1}}
\newcommand{\SchemaIdentityRelation}{Identity}
\newcommand{\SchemaEpistemic}{Epistemic}
\newcommand{\SchemaDesire}{Desire}
\newcommand{\SchemaPhysical}{Physical}
\newcommand{\LabelTrue}{True}
\newcommand{\LabelFalse}{False}
\newcommand{\LabelUnknown}{Unknown}
\newcommand{\LabelPublic}{Public}
\newcommand{\LabelShared}{Shared}
\newcommand{\LabelPrivate}{Private}
\newcommand{\LabelExplicit}{Explicit}
\newcommand{\LabelImplicit}{Implicit}
\newcommand{\LabelLocation}{Location}
\newcommand{\LabelAction}{Action}
\newcommand{\LabelTrait}{Trait}
\newcommand{\LabelNarration}{Narration}
\newcommand{\LabelPerception}{Perception}
\newcommand{\LabelMemory}{Memory}
\newcommand{\LabelTestimony}{Testimony}
\newcommand{\LabelInference}{Inference}
\newcommand{\LabelNeutral}{Neutral}
\newcommand{\LabelTemporal}{Temporal}
\newcommand{\ExampleStoryRow}[1]{%
\multicolumn{8}{@{}p{\dimexpr\textwidth\relax}@{}}{\scriptsize Story: #1}\\}
\newcommand{\ExampleSectionRow}[1]{%
\multicolumn{8}{c}{\scriptsize\bfseries #1} \\}
\newcommand{\ExampleTableHeader}[3]{%
\setlength{\tabcolsep}{1.5pt}%
\renewcommand{\arraystretch}{1.08}%
\begin{longtable}{@{}>{\scriptsize\raggedright\arraybackslash}p{\dimexpr\textwidth-6.53cm-14\tabcolsep\relax}>{\scriptsize\raggedleft\arraybackslash}p{0.42cm}>{\scriptsize\raggedright\arraybackslash}p{1.12cm}>{\scriptsize\raggedright\arraybackslash}p{0.95cm}>{\scriptsize\raggedright\arraybackslash}p{0.90cm}>{\scriptsize\raggedright\arraybackslash}p{1.05cm}>{\scriptsize\raggedright\arraybackslash}p{1.08cm}>{\scriptsize\raggedright\arraybackslash}p{1.01cm}@{}}
\caption{\normalsize #1}\label{#2}\\
\toprule
\ExampleStoryRow{#3}
\midrule
\textbf{Belief Proposition} & \textbf{Ord} & \textbf{Truth} & \textbf{Access} & \textbf{Repr.} & \textbf{Type} & \textbf{Source} & \textbf{Context} \\
\midrule
\endfirsthead
\caption[]{\normalsize #1 (continued)}\\
\toprule
\textbf{Belief Proposition} & \textbf{Ord} & \textbf{Truth} & \textbf{Access} & \textbf{Repr.} & \textbf{Type} & \textbf{Source} & \textbf{Context} \\
\midrule
\endhead
\endfoot
\bottomrule
\endlastfoot
}

\ExampleTableHeader{Ambiguous Story Task. Ambiguous nonverbal coordination around a shared plan.}{tab:example-ast}{Mingfeng, Xiaoyu, and Xiaolin are good friends. This afternoon, Mingfeng watches Xiaoyu and her newly adopted pet dog play under the shade of the trees in the community. Suddenly, he winks at Xiaolin, who is next to him, and looks at the nearby pet supply store. Xiaolin looks at Xiaoyu, then responds to Mingfeng's gaze. After that, she stands up and walks towards the pet supply store.}
\ExampleSectionRow{Narrated Facts (world)}
\midrule
Mingfeng, Xiaoyu, and Xiaolin are good friends & 0 & \LabelTrue{} & \LabelPublic{} & \LabelExplicit{} & \SchemaIdentityRelation & \LabelNarration{} & \LabelNeutral{} \\
Mingfeng watches Xiaoyu and her newly adopted pet dog play under the shade of the trees in the community & 0 & \LabelTrue{} & \LabelPublic{} & \LabelExplicit{} & \LabelAction{} & \LabelNarration{} & \LabelNeutral{} \\
Mingfeng winks at Xiaolin & 0 & \LabelTrue{} & \LabelPublic{} & \LabelExplicit{} & \LabelAction{} & \LabelNarration{} & \LabelNeutral{} \\
Mingfeng looks at the nearby pet supply store & 0 & \LabelTrue{} & \LabelPublic{} & \LabelExplicit{} & \LabelAction{} & \LabelNarration{} & \LabelNeutral{} \\
Xiaolin looks at Xiaoyu & 0 & \LabelTrue{} & \LabelPublic{} & \LabelExplicit{} & \LabelAction{} & \LabelNarration{} & \LabelNeutral{} \\
Xiaolin responds to Mingfeng's gaze & 0 & \LabelTrue{} & \LabelPublic{} & \LabelExplicit{} & \LabelAction{} & \LabelNarration{} & \LabelNeutral{} \\
Xiaolin stands up and walks towards the pet supply store & 0 & \LabelTrue{} & \LabelPublic{} & \LabelExplicit{} & \LabelAction{} & \LabelNarration{} & \LabelNeutral{} \\
\midrule
\ExampleSectionRow{Actor Beliefs (Mingfeng)}
\midrule
Xiaoyu has a newly adopted pet dog & 1 & \LabelTrue{} & \LabelPublic{} & \LabelImplicit{} & \SchemaIdentityRelation & \LabelPerception{} & \LabelNeutral{} \\
Xiaoyu is playing with her pet dog under the shade of the trees & 1 & \LabelTrue{} & \LabelPublic{} & \LabelImplicit{} & \LabelAction{} & \LabelPerception{} & \LabelNeutral{} \\
The pet supply store is nearby & 1 & \LabelTrue{} & \LabelPrivate{} & \LabelImplicit{} & \LabelLocation{} & \LabelPerception{} & \LabelNeutral{} \\
Xiaolin is next to Mingfeng & 1 & \LabelTrue{} & \LabelPublic{} & \LabelImplicit{} & \LabelLocation{} & \LabelPerception{} & \LabelNeutral{} \\
Mingfeng's wink will signal Xiaolin about going to the pet supply store & 1 & \LabelUnknown{} & \LabelPrivate{} & \LabelImplicit{} & \SchemaEpistemic & \LabelInference{} & \LabelNeutral{} \\
Xiaolin will understand the wink as a suggestion to go to the pet supply store & 2 & \LabelUnknown{} & \LabelPrivate{} & \LabelImplicit{} & \SchemaEpistemic & \LabelInference{} & \LabelNeutral{} \\
\midrule
\ExampleSectionRow{Actor Beliefs (Xiaolin)}
\midrule
Mingfeng, Xiaoyu, and Xiaolin are good friends & 1 & \LabelTrue{} & \LabelPrivate{} & \LabelImplicit{} & \SchemaIdentityRelation & \LabelMemory{} & \LabelNeutral{} \\
Xiaoyu has a newly adopted pet dog & 1 & \LabelTrue{} & \LabelPublic{} & \LabelImplicit{} & \SchemaIdentityRelation & \LabelPerception{} & \LabelNeutral{} \\
Xiaoyu is playing with her pet dog under the shade of the trees & 1 & \LabelTrue{} & \LabelPublic{} & \LabelImplicit{} & \LabelAction{} & \LabelPerception{} & \LabelNeutral{} \\
The pet supply store is nearby & 1 & \LabelTrue{} & \LabelShared{} & \LabelExplicit{} & \LabelLocation{} & \LabelPerception{} & \LabelNeutral{} \\
Mingfeng winks at Xiaolin & 1 & \LabelTrue{} & \LabelShared{} & \LabelExplicit{} & \LabelAction{} & \LabelPerception{} & \LabelNeutral{} \\
Mingfeng looks at the pet supply store & 1 & \LabelTrue{} & \LabelShared{} & \LabelImplicit{} & \LabelAction{} & \LabelPerception{} & \LabelNeutral{} \\
Mingfeng wants to go to the pet supply store to buy pet toys & 1 & \LabelUnknown{} & \LabelPrivate{} & \LabelImplicit{} & \SchemaDesire & \LabelInference{} & \LabelNeutral{} \\
Mingfeng is signaling Xiaolin to go to the pet supply store & 1 & \LabelTrue{} & \LabelPrivate{} & \LabelImplicit{} & \SchemaEpistemic & \LabelInference{} & \LabelNeutral{} \\
Going to the pet supply store will be helpful for Xiaoyu or her dog & 1 & \LabelUnknown{} & \LabelPrivate{} & \LabelImplicit{} & \SchemaDesire & \LabelInference{} & \LabelNeutral{} \\
\midrule
\ExampleSectionRow{Actor Beliefs (Xiaoyu)}
\midrule
Mingfeng, Xiaoyu, and Xiaolin are good friends & 1 & \LabelTrue{} & \LabelPublic{} & \LabelImplicit{} & \SchemaIdentityRelation & \LabelMemory{} & \LabelNeutral{} \\
Mingfeng is watching Xiaoyu and the dog & 1 & \LabelTrue{} & \LabelPrivate{} & \LabelImplicit{} & \LabelAction{} & \LabelPerception{} & \LabelNeutral{} \\
Xiaolin is looking at Xiaoyu & 1 & \LabelTrue{} & \LabelPrivate{} & \LabelImplicit{} & \LabelAction{} & \LabelPerception{} & \LabelNeutral{} \\
Mingfeng winked at Xiaolin & 1 & \LabelTrue{} & \LabelPrivate{} & \LabelImplicit{} & \LabelAction{} & \LabelPerception{} & \LabelNeutral{} \\
Mingfeng looked at the pet supply store & 1 & \LabelTrue{} & \LabelPrivate{} & \LabelImplicit{} & \LabelAction{} & \LabelPerception{} & \LabelNeutral{} \\
Xiaolin is walking towards the pet supply store & 1 & \LabelTrue{} & \LabelPrivate{} & \LabelImplicit{} & \LabelAction{} & \LabelPerception{} & \LabelNeutral{} \\
Mingfeng and Xiaolin are doing something confusing & 1 & \LabelUnknown{} & \LabelPrivate{} & \LabelImplicit{} & \LabelAction{} & \LabelPerception{} & \LabelNeutral{} \\
\end{longtable}

\ExampleTableHeader{False Belief Task. Hidden transfer and outdated belief.}{tab:example-fbt}{Alice and Bob are in a room. There is an object in a box. Bob leaves. Alice moves the object to the safe.}
\ExampleSectionRow{Narrated Facts (world)}
\midrule
Alice is in the room & 0 & \LabelTrue{} & \LabelPublic{} & \LabelExplicit{} & \LabelLocation{} & \LabelNarration{} & \LabelNeutral{} \\
Bob is in the room & 0 & \LabelTrue{} & \LabelPublic{} & \LabelExplicit{} & \LabelLocation{} & \LabelNarration{} & \LabelTemporal{} \\
There is an object in a box & 0 & \LabelTrue{} & \LabelPublic{} & \LabelExplicit{} & \LabelLocation{} & \LabelNarration{} & \LabelTemporal{} \\
There is a safe in the room & 0 & \LabelTrue{} & \LabelPublic{} & \LabelExplicit{} & \LabelLocation{} & \LabelNarration{} & \LabelNeutral{} \\
Bob leaves the room & 0 & \LabelTrue{} & \LabelPublic{} & \LabelExplicit{} & \LabelAction{} & \LabelNarration{} & \LabelNeutral{} \\
Alice moves the object to the safe & 0 & \LabelTrue{} & \LabelPrivate{} & \LabelExplicit{} & \LabelAction{} & \LabelNarration{} & \LabelNeutral{} \\
The object is in the safe & 0 & \LabelTrue{} & \LabelPrivate{} & \LabelExplicit{} & \LabelLocation{} & \LabelNarration{} & \LabelNeutral{} \\
\midrule
\ExampleSectionRow{Actor Beliefs (Alice)}
\midrule
Bob is in the room & 1 & \LabelTrue{} & \LabelPrivate{} & \LabelImplicit{} & \LabelLocation{} & \LabelPerception{} & \LabelTemporal{} \\
There is a box in the room & 1 & \LabelTrue{} & \LabelPublic{} & \LabelImplicit{} & \LabelLocation{} & \LabelPerception{} & \LabelNeutral{} \\
There is a safe in the room & 1 & \LabelTrue{} & \LabelPublic{} & \LabelImplicit{} & \LabelLocation{} & \LabelPerception{} & \LabelNeutral{} \\
The object is in the box & 1 & \LabelTrue{} & \LabelPublic{} & \LabelImplicit{} & \LabelLocation{} & \LabelPerception{} & \LabelTemporal{} \\
Bob left the room & 1 & \LabelTrue{} & \LabelPrivate{} & \LabelImplicit{} & \LabelAction{} & \LabelPerception{} & \LabelNeutral{} \\
The object is in the safe & 1 & \LabelTrue{} & \LabelPrivate{} & \LabelImplicit{} & \LabelLocation{} & \LabelPerception{} & \LabelNeutral{} \\
Bob thinks the object is in the box & 2 & \LabelTrue{} & \LabelPrivate{} & \LabelImplicit{} & \SchemaEpistemic & \LabelInference{} & \LabelNeutral{} \\
Bob thinks Alice thinks the object is in the box & 3 & \LabelTrue{} & \LabelPrivate{} & \LabelImplicit{} & \SchemaEpistemic & \LabelInference{} & \LabelNeutral{} \\
\midrule
\ExampleSectionRow{Actor Beliefs (Bob)}
\midrule
Alice is in the room & 1 & \LabelTrue{} & \LabelPrivate{} & \LabelImplicit{} & \LabelLocation{} & \LabelPerception{} & \LabelNeutral{} \\
There is a box in the room & 1 & \LabelTrue{} & \LabelPublic{} & \LabelImplicit{} & \LabelLocation{} & \LabelPerception{} & \LabelNeutral{} \\
There is a safe in the room & 1 & \LabelTrue{} & \LabelPublic{} & \LabelImplicit{} & \LabelLocation{} & \LabelPerception{} & \LabelNeutral{} \\
The object is in the box & 1 & \LabelFalse{} & \LabelPrivate{} & \LabelImplicit{} & \LabelLocation{} & \LabelPerception{} & \LabelTemporal{} \\
Alice thinks the object is in the box & 2 & \LabelFalse{} & \LabelPrivate{} & \LabelImplicit{} & \SchemaEpistemic & \LabelInference{} & \LabelTemporal{} \\
Alice thinks Bob thinks the object is in the box & 3 & \LabelTrue{} & \LabelPrivate{} & \LabelImplicit{} & \SchemaEpistemic & \LabelInference{} & \LabelNeutral{} \\
\end{longtable}

\ExampleTableHeader{Faux-pas Recognition Test. Social expectations around a promise to attend a game.}{tab:example-fpt}{On Saturday morning, Xiao Wang and Xiao Zhao meet at the school gate. Xiao Zhao says: "I have a basketball game this afternoon, will you come to watch?" Xiao Wang remembers Xiao Zhao's game, and replies: "Of course, I will definitely go to support you." Xiao Zhao happily says: "Great, thank you!"}
\ExampleSectionRow{Narrated Facts (world)}
\midrule
Xiao Wang and Xiao Zhao meet at the school gate on Saturday morning & 0 & \LabelTrue{} & \LabelPublic{} & \LabelExplicit{} & \LabelAction{} & \LabelNarration{} & \LabelNeutral{} \\
Xiao Zhao says, "I have a basketball game this afternoon, will you come to watch?" & 0 & \LabelTrue{} & \LabelPublic{} & \LabelExplicit{} & \LabelAction{} & \LabelNarration{} & \LabelNeutral{} \\
Xiao Wang remembers Xiao Zhao's game & 0 & \LabelTrue{} & \LabelPublic{} & \LabelExplicit{} & \LabelAction{} & \LabelNarration{} & \LabelNeutral{} \\
Xiao Wang replies, "Of course, I will definitely go to support you." & 0 & \LabelTrue{} & \LabelPublic{} & \LabelExplicit{} & \LabelAction{} & \LabelNarration{} & \LabelNeutral{} \\
Xiao Zhao says, "Great, thank you!" & 0 & \LabelTrue{} & \LabelPublic{} & \LabelExplicit{} & \LabelAction{} & \LabelNarration{} & \LabelNeutral{} \\
\midrule
\ExampleSectionRow{Actor Beliefs (Xiao Zhao)}
\midrule
Xiao Zhao has a basketball game this afternoon & 1 & \LabelTrue{} & \LabelPublic{} & \LabelExplicit{} & \LabelAction{} & \LabelMemory{} & \LabelNeutral{} \\
Xiao Wang might come to watch the basketball game & 1 & \LabelTrue{} & \LabelPrivate{} & \LabelImplicit{} & \SchemaDesire & \LabelInference{} & \LabelNeutral{} \\
Xiao Wang will come to support Xiao Zhao & 1 & \LabelTrue{} & \LabelPrivate{} & \LabelImplicit{} & \SchemaDesire & \LabelTestimony{} & \LabelNeutral{} \\
Xiao Wang remembers the basketball game & 1 & \LabelTrue{} & \LabelPrivate{} & \LabelImplicit{} & \SchemaEpistemic & \LabelTestimony{} & \LabelNeutral{} \\
Xiao Wang wants to support Xiao Zhao & 1 & \LabelUnknown{} & \LabelPrivate{} & \LabelImplicit{} & \SchemaDesire & \LabelInference{} & \LabelNeutral{} \\
\midrule
\ExampleSectionRow{Actor Beliefs (Xiao Wang)}
\midrule
Xiao Zhao has a basketball game this afternoon & 1 & \LabelTrue{} & \LabelPublic{} & \LabelExplicit{} & \LabelAction{} & \LabelMemory{} & \LabelNeutral{} \\
Xiao Wang will go to support Xiao Zhao & 1 & \LabelTrue{} & \LabelPublic{} & \LabelExplicit{} & \SchemaDesire & \LabelMemory{} & \LabelNeutral{} \\
Xiao Zhao thinks Xiao Wang remembers the basketball game & 2 & \LabelUnknown{} & \LabelPrivate{} & \LabelImplicit{} & \SchemaEpistemic & \LabelInference{} & \LabelNeutral{} \\
Xiao Zhao thinks Xiao Wang will come to watch & 2 & \LabelTrue{} & \LabelPrivate{} & \LabelImplicit{} & \SchemaEpistemic & \LabelInference{} & \LabelNeutral{} \\
\end{longtable}

\ExampleTableHeader{Hinting Task Test. Indirect birthday hint about wanting a dog.}{tab:example-ht}{Rebecca's birthday is coming soon. She says to her father, "I like animals, especially dogs."}
\ExampleSectionRow{Narrated Facts (world)}
\midrule
Rebecca's birthday is coming soon & 0 & \LabelTrue{} & \LabelPublic{} & \LabelExplicit{} & \LabelAction{} & \LabelNarration{} & \LabelNeutral{} \\
Rebecca says to Rebecca's father, "I like animals, especially dogs." & 0 & \LabelTrue{} & \LabelPublic{} & \LabelExplicit{} & \LabelAction{} & \LabelNarration{} & \LabelNeutral{} \\
\midrule
\ExampleSectionRow{Actor Beliefs (Rebecca)}
\midrule
Rebecca's birthday is coming soon & 1 & \LabelTrue{} & \LabelPrivate{} & \LabelImplicit{} & \LabelAction{} & \LabelMemory{} & \LabelNeutral{} \\
A dog would be a good birthday gift & 1 & \LabelUnknown{} & \LabelPrivate{} & \LabelImplicit{} & \SchemaDesire & \LabelInference{} & \LabelNeutral{} \\
Rebecca's father can buy Rebecca a dog & 1 & \LabelUnknown{} & \LabelPrivate{} & \LabelImplicit{} & \SchemaDesire & \LabelInference{} & \LabelNeutral{} \\
Rebecca's father will understand that Rebecca wants a dog as a birthday gift & 1 & \LabelUnknown{} & \LabelPrivate{} & \LabelImplicit{} & \SchemaEpistemic & \LabelInference{} & \LabelNeutral{} \\
Rebecca's father thinks Rebecca likes dogs & 2 & \LabelUnknown{} & \LabelPrivate{} & \LabelImplicit{} & \SchemaEpistemic & \LabelInference{} & \LabelNeutral{} \\
\midrule
\ExampleSectionRow{Actor Beliefs (Rebecca's father)}
\midrule
Rebecca likes animals & 1 & \LabelTrue{} & \LabelPublic{} & \LabelExplicit{} & \LabelTrait{} & \LabelTestimony{} & \LabelNeutral{} \\
Rebecca especially likes dogs & 1 & \LabelTrue{} & \LabelPublic{} & \LabelExplicit{} & \LabelTrait{} & \LabelTestimony{} & \LabelNeutral{} \\
Rebecca implies Rebecca wants a dog as a birthday gift & 1 & \LabelUnknown{} & \LabelPrivate{} & \LabelImplicit{} & \SchemaDesire & \LabelInference{} & \LabelNeutral{} \\
Rebecca thinks Rebecca's father should get Rebecca a dog as a good birthday gift & 2 & \LabelUnknown{} & \LabelPrivate{} & \LabelImplicit{} & \SchemaEpistemic & \LabelInference{} & \LabelNeutral{} \\
\end{longtable}

\ExampleTableHeader{Persuasion Story Task. A larger-office goal requiring influence.}{tab:example-pst}{Xiao Hong wants to move to a bigger office, but that office is occupied by her colleague Xiao Li.}
\ExampleSectionRow{Narrated Facts (world)}
\midrule
Xiao Hong wants to move to a bigger office & 0 & \LabelTrue{} & \LabelPublic{} & \LabelExplicit{} & \SchemaDesire & \LabelNarration{} & \LabelNeutral{} \\
The bigger office is occupied by Xiao Li & 0 & \LabelTrue{} & \LabelPublic{} & \LabelExplicit{} & \LabelLocation{} & \LabelNarration{} & \LabelNeutral{} \\
Xiao Hong and Xiao Li are colleagues & 0 & \LabelTrue{} & \LabelPublic{} & \LabelExplicit{} & \SchemaIdentityRelation & \LabelNarration{} & \LabelNeutral{} \\
\midrule
\ExampleSectionRow{Actor Beliefs (Xiao Hong)}
\midrule
Xiao Hong needs the bigger office that Xiao Li occupies & 1 & \LabelTrue{} & \LabelPrivate{} & \LabelImplicit{} & \SchemaDesire & \LabelInference{} & \LabelNeutral{} \\
Xiao Hong must persuade Xiao Li to give up the bigger office & 1 & \LabelTrue{} & \LabelPrivate{} & \LabelImplicit{} & \SchemaDesire & \LabelInference{} & \LabelNeutral{} \\
Xiao Li will agree to exchange offices if Xiao Hong offers convenient conditions & 1 & \LabelUnknown{} & \LabelPrivate{} & \LabelImplicit{} & \SchemaEpistemic & \LabelInference{} & \LabelNeutral{} \\
\end{longtable}

\ExampleTableHeader{Scalar Implicature Test. Approximate quantity reasoning over white chickens.}{tab:example-sit}{On a farm, Farmer Wang keeps 15 chickens, almost a third of which are white. He counts some of them and finds that 4 are white.}
\ExampleSectionRow{Narrated Facts (world)}
\midrule
Farmer Wang is on a farm & 0 & \LabelTrue{} & \LabelPublic{} & \LabelExplicit{} & \LabelLocation{} & \LabelNarration{} & \LabelNeutral{} \\
Farmer Wang keeps 15 chickens & 0 & \LabelTrue{} & \LabelPublic{} & \LabelExplicit{} & \SchemaPhysical & \LabelNarration{} & \LabelNeutral{} \\
Almost a third of the chickens are white & 0 & \LabelTrue{} & \LabelPublic{} & \LabelExplicit{} & \SchemaPhysical & \LabelNarration{} & \LabelNeutral{} \\
Farmer Wang counts some of the chickens & 0 & \LabelTrue{} & \LabelPublic{} & \LabelExplicit{} & \LabelAction{} & \LabelNarration{} & \LabelNeutral{} \\
Farmer Wang finds that 4 are white & 0 & \LabelTrue{} & \LabelPublic{} & \LabelExplicit{} & \SchemaPhysical & \LabelNarration{} & \LabelNeutral{} \\
\midrule
\ExampleSectionRow{Actor Beliefs (Farmer Wang)}
\midrule
There are 15 chickens & 1 & \LabelTrue{} & \LabelPublic{} & \LabelImplicit{} & \SchemaPhysical & \LabelPerception{} & \LabelNeutral{} \\
Before counting, almost a third of the chickens are white & 1 & \LabelTrue{} & \LabelPublic{} & \LabelImplicit{} & \SchemaPhysical & \LabelPerception{} & \LabelNeutral{} \\
Before counting, probably 5 chickens are white & 1 & \LabelUnknown{} & \LabelPrivate{} & \LabelImplicit{} & \SchemaPhysical & \LabelInference{} & \LabelNeutral{} \\
After counting some of the chickens, probably 5 chickens are white in total & 1 & \LabelUnknown{} & \LabelPrivate{} & \LabelImplicit{} & \SchemaPhysical & \LabelInference{} & \LabelNeutral{} \\
\end{longtable}

\ExampleTableHeader{Strange Story Task. Nonliteral reassurance about coughing.}{tab:example-sst}{Emma coughs. Throughout lunchtime, she keeps coughing. Dad says, "Poor Emma, you must have a frog in your throat!"}
\ExampleSectionRow{Narrated Facts (world)}
\midrule
Emma keeps coughing throughout lunchtime & 0 & \LabelTrue{} & \LabelPublic{} & \LabelExplicit{} & \LabelAction{} & \LabelNarration{} & \LabelNeutral{} \\
Dad says, "Poor Emma, you must have a frog in your throat!" & 0 & \LabelTrue{} & \LabelPublic{} & \LabelExplicit{} & \LabelAction{} & \LabelNarration{} & \LabelNeutral{} \\
\midrule
\ExampleSectionRow{Actor Beliefs (Emma)}
\midrule
Dad says Emma has a frog in her throat & 1 & \LabelTrue{} & \LabelPrivate{} & \LabelExplicit{} & \LabelAction{} & \LabelPerception{} & \LabelNeutral{} \\
Dad thinks Emma does not actually have a frog in her throat & 2 & \LabelTrue{} & \LabelPrivate{} & \LabelImplicit{} & \SchemaEpistemic & \LabelInference{} & \LabelNeutral{} \\
Dad thinks Emma's cough sounds like a frog's call & 2 & \LabelTrue{} & \LabelPrivate{} & \LabelImplicit{} & \SchemaEpistemic & \LabelInference{} & \LabelNeutral{} \\
Dad wants to make Emma laugh and feel better & 1 & \LabelUnknown{} & \LabelPrivate{} & \LabelImplicit{} & \SchemaDesire & \LabelInference{} & \LabelNeutral{} \\
Dad thinks the joke will make Emma laugh and feel better & 1 & \LabelUnknown{} & \LabelPrivate{} & \LabelImplicit{} & \SchemaDesire & \LabelInference{} & \LabelNeutral{} \\
\midrule
\ExampleSectionRow{Actor Beliefs (Dad)}
\midrule
Emma keeps coughing during lunchtime & 1 & \LabelTrue{} & \LabelPrivate{} & \LabelImplicit{} & \LabelAction{} & \LabelPerception{} & \LabelNeutral{} \\
Emma does not actually have a frog in her throat & 1 & \LabelTrue{} & \LabelPrivate{} & \LabelImplicit{} & \SchemaPhysical & \LabelInference{} & \LabelNeutral{} \\
Emma's cough sounds like a frog's call & 1 & \LabelUnknown{} & \LabelPrivate{} & \LabelImplicit{} & \SchemaPhysical & \LabelPerception{} & \LabelNeutral{} \\
Saying "Poor Emma, you must have a frog in your throat!" will make Emma laugh and feel better & 1 & \LabelUnknown{} & \LabelPrivate{} & \LabelImplicit{} & \SchemaDesire & \LabelInference{} & \LabelNeutral{} \\
Emma thinks Dad is trying to make Emma laugh and feel better & 2 & \LabelUnknown{} & \LabelPrivate{} & \LabelImplicit{} & \SchemaEpistemic & \LabelInference{} & \LabelNeutral{} \\
\end{longtable}

\FloatBarrier
\clearpage
\onecolumn
\section*{NeurIPS Paper Checklist}

\begin{enumerate}

\item {\bf Claims}
    \item[] Question: Do the main claims made in the abstract and introduction accurately reflect the paper's contributions and scope?
    \item[] Answer: \answerYes{}
    \item[] Justification: The abstract, introduction, Secs.~\ref{sec:framework}--\ref{sec:experiments}, and Sec.~\ref{sec:conclusion-limitations} consistently describe OmniToM as a benchmark for explicit belief modeling, state the two-stage evaluation scope, and identify the actor-specific information-tracking bottleneck observed in the experiments.

\item {\bf Limitations}
    \item[] Question: Does the paper discuss the limitations of the work performed by the authors?
    \item[] Answer: \answerYes{}
    \item[] Justification: Sec.~\ref{sec:conclusion-limitations} states the main limitations, including the text-only setting, inherited source-corpus biases, bounded ToM scope, human-labeled schema subjectivity, and the approximate nature of semantic-judge scoring. App.~\ref{app:annotation-pipeline-overview} also clarifies that pipeline validation preceded scaling and that no separate human audit of the completed benchmark was conducted.

\item {\bf Theory assumptions and proofs}
    \item[] Question: For each theoretical result, does the paper provide the full set of assumptions and a complete and correct proof?
    \item[] Answer: \answerNA{}
    \item[] Justification: The paper does not present theoretical theorems or formal proofs.

\item {\bf Experimental result reproducibility}
    \item[] Question: Does the paper fully disclose all the information needed to reproduce the main experimental results of the paper to the extent that it affects the main claims and/or conclusions of the paper, regardless of whether the code and data are provided or not?
    \item[] Answer: \answerYes{}
    \item[] Justification: Secs.~\ref{sec:dataset}, \ref{sec:evaluation}, and \ref{sec:experiments} specify the retained source categories, benchmark split, judge protocol, model set, prompt regime, and evaluation metrics, including story-macro Endpoint QA accuracy on linked ToMBench questions; App.~\ref{app:construction-validation}, App.~\ref{app:benchmark-record-format}, and App.~\ref{app:prompt-engineering-teler} provide construction, record-format, and prompt details.

\item {\bf Open access to data and code}
    \item[] Question: Does the paper provide open access to the data and code, with sufficient instructions to faithfully reproduce the main experimental results, as described in supplemental material?
    \item[] Answer: \answerYes{}
    \item[] Justification: The public dataset and code releases include the benchmark JSONL file, prompt-builder scripts, replication runner, README instructions, dataset card, and Croissant metadata with Responsible AI fields; App.~\ref{app:benchmark-record-format} and App.~\ref{app:licensing-release} document the released artifact.

\item {\bf Experimental setting/details}
    \item[] Question: Does the paper specify all the training and test details necessary to understand the results?
    \item[] Answer: \answerYes{}
    \item[] Justification: Secs.~\ref{sec:dataset}, \ref{sec:evaluation}, and \ref{sec:experiments} describe the retained source split, prompt regime, semantic-judge protocol, benchmarked models, quantization note, compute setting, Stage~1/Stage~2 metrics, and story-macro Endpoint QA protocol; the supplement provides prompt and schema-format details.

\item {\bf Experiment statistical significance}
    \item[] Question: Does the paper report error bars suitably and correctly defined or other appropriate information about the statistical significance of the experiments?
    \item[] Answer: \answerYes{}
    \item[] Justification: Stage~1 and Endpoint QA scores include 95\% story-bootstrap confidence intervals in Table~\ref{tab:benchmark-main}. App.~\ref{app:additional-results} reports confidence intervals for Stage~2 dimensions and overall, plus selected paired model comparisons. The intervals are percentile intervals from 10{,}000 story-level resamples and are not adjusted for multiple comparisons.

\item {\bf Experiments compute resources}
    \item[] Question: For each experiment, does the paper provide sufficient information on the computer resources needed to reproduce the experiments?
    \item[] Answer: \answerYes{}
    \item[] Justification: Sec.~\ref{sec:experiments} reports the evaluated model set and states that open-weight runs were executed on NVIDIA A100 80GB GPUs with 4-bit quantization where required, requiring approximately 48 GPU-hours per open-weight model.

\item {\bf Code of ethics}
    \item[] Question: Have you read the NeurIPS Code of Ethics and ensured that your research conforms to it?
    \item[] Answer: \answerYes{}
    \item[] Justification: The authors have reviewed the NeurIPS Code of Ethics; the paper is benchmark-focused, Sec.~\ref{sec:conclusion-limitations} states the benchmark's bounded validity, and the \hyperref[sec:broader-impacts]{supplemental Broader Impacts section} discusses misuse risk and the danger of over-interpreting benchmark scores as deployment-safety evidence.

\item {\bf Broader impacts}
    \item[] Question: If appropriate for the scope and focus of your paper, did you discuss potential negative societal impacts of your work?
    \item[] Answer: \answerYes{}
    \item[] Justification: Potential positive and negative societal impacts are discussed in the \hyperref[sec:broader-impacts]{supplemental Broader Impacts section}. We emphasize that OmniToM is an evaluation benchmark, not a deployment guarantee or safety certification, and note potential dual-use risks involving persuasion, deception, or strategic modeling of users' beliefs.

\item {\bf Safeguards}
    \item[] Question: Do you have safeguards in place for responsible release of models with a high risk for misuse?
    \item[] Answer: \answerNA{}
    \item[] Justification: The paper does not release a pretrained model or other high-risk model artifact that would require controlled model-release safeguards.

\item {\bf Licenses}
    \item[] Question: If you are using existing assets, did you cite the creators and respect the license and terms of use?
    \item[] Answer: \answerYes{}
    \item[] Justification: OmniToM cites ToMBench as the source corpus, and App.~\ref{app:licensing-release} states that the source ToMBench repository is MIT-licensed, that the OmniToM release package documents its own MIT license, and that third-party model services are not redistributed.

\item {\bf Assets}
    \item[] Question: If you are releasing new assets, did you document them and provide these details alongside the assets?
    \item[] Answer: \answerYes{}
    \item[] Justification: Secs.~\ref{sec:dataset}--\ref{sec:evaluation} and App.~\ref{app:construction-validation} describe the new benchmark asset; App.~\ref{app:benchmark-record-format} gives the released JSONL schema, and App.~\ref{app:licensing-release} records release conditions, dataset-card documentation, and Croissant/RAI metadata.

\item {\bf Crowdsourcing and research with human subjects}
    \item[] Question: If you used crowdsourcing or conducted research with human subjects, did you include the full text of instructions given to participants and screenshots, if applicable, as well as details about compensation?
    \item[] Answer: \answerNA{}
    \item[] Justification: The benchmark does not use crowdsourcing platforms and does not study annotators as human subjects. Human annotation labor is project-side dataset curation; App.~\ref{app:annotation-pipeline-overview} reports the anonymized annotator count and annotation-effort estimate.

\item {\bf Institutional review board approvals or equivalent for research with human subjects}
    \item[] Question: Did you describe any potential participant risks and obtain Institutional Review Board approvals, or an equivalent approval/review based on the requirements of your institution, if applicable?
    \item[] Answer: \answerNA{}
    \item[] Justification: No human-subjects study is reported, so IRB approval is not applicable.

\item {\bf Declaration of LLM usage}
    \item[] Question: Does the paper describe the usage of LLMs if it is an important, original, or non-standard component of the core methods in this research?
    \item[] Answer: \answerYes{}
    \item[] Justification: LLM usage is central to both construction and evaluation and is described in Secs.~\ref{sec:dataset}, \ref{sec:judge}, and \ref{sec:experiments}, with TELeR prompt details in App.~\ref{app:prompt-engineering-teler}.

\end{enumerate}

\end{document}